\documentclass{article}

\usepackage{geometry, lineno,hyperref,amsfonts,amsfonts,booktabs,graphicx,subfigure,rotating,amsmath,enumerate}
\newgeometry{left=3cm}
\begin{document}

\title{Scene Flow Estimation: A Survey}

\author{Zike Yan\textsuperscript{1,a}, Xuezhi Xiang\textsuperscript{2,a,*}\\ \textsuperscript{a}College of Information and Communication Engineering,\\ Harbin Engineering University, Harbin 150001, China}

\maketitle

\begin{abstract}
This paper is the first to review the scene flow estimation, which analyzes and compares methods, technical challenges, evaluation methodologies and performance of scene flow estimation. Existing algorithms are categorized in terms of scene representation, data source, and calculation scheme, and the pros and cons in each category are compared briefly. The datasets and evaluation protocols are enumerated, and the performance of the most representative methods is presented. A future vision is illustrated with few questions arisen for discussion. This survey presents a general introduction and analysis of scene flow estimation.
\end{abstract}
\renewcommand{\thefootnote}{}
\footnote{\textsuperscript{*} Corresponding author}
\footnote{Email addresses: \href{mailto:yanzike@hrbeu.edu.cn}{yanzike@hrbeu.edu.cn} (Zike Yan\textsuperscript{1}), \href{mailto:xiangxuezhi@hrbeu.edu.cn}{xiangxuezhi@hrbeu.edu.cn} (Xuezhi Xiang\textsuperscript{2})}

%%%%%%%%
%%%%%%%%
%%%%%%%%------------------------------正文部分----------------------------
%%%%%%%%
%%%%%%%%
%%%%%%%%
\section{Introduction}
\label{sec:introduction}
Scene flow is a three-dimensional motion field of the surface in world space, or in other words, it shows the three-dimensional displacement vector of each surface point between two frames. As most computer vision issues are, scene flow estimation is essentially an ill-posed energy minimization problem with three unknowns. Prior knowledge in multiple aspects is required to make the energy function solvable with just a few pairs of images. Hence, it's essential to fully make use of information from the data source and to weigh different prior knowledge for a better performance.

The paper attempts to reveal clues by providing a comprehensive literature survey in this field. Scene flow is first introduced by Vedula in 1999~\cite{01} and has made constant progress over the years. Diverse data sources has emerged thus scene flow estimation don't need to set up the complicated array of cameras. The conventional framework derived from optical flow field~\cite{H-S, L-K} has extended to this three-dimensional motion field estimation task, while diverse ideas and optimization manners has improved the performance noticeably. The widely concerned learning based method has been utilized for scene flow estimation~\cite{Mayer}, which brings fresh blood to this integrated field. Moreover, a few methods have achieved real-time estimation with GPU implementation at the QVGA($320\times240$) resolution~\cite{24,31,40,65}, which insure a promising efficiency. The emergence of these methods stands for the fact that scene flow estimation will be widely utilized and applied to practice soon in the near future. 

The paper is organized as follows. Section~\ref{sec:background} illustrates the relevant issues, challenges and applications of scene flow as a background . Section~\ref{sec:taxonomy} provides classification of scene flow in terms of three major components. Emerging datasets that are publicly available and the diverse evaluation protocols are presented and analyzed in Section~\ref{sec:evaluation}. Section~\ref{sec:discussion} arises few questions to briefly discuss the content mentioned above, and the future vision is provided. Finally, a conclusion is presented in Section~\ref{sec:conclusion}.

\section{Background}
\label{sec:background}
We provide relevant issues, major challenges and applications as the background information for better understanding this field.																										
\subsection{Relevant issues}
\label{subsec:subissue}
Scene flow estimation is an integrated task, which is relevant to multiple issues. Firstly, optical flow is the projection of scene flow onto an image plane, which is the basis of scene flow and has made steady progress over the years. The basic framework and innovations of scene flow estimation mainly derives from optical flow estimation field. Secondly, in a binocular setting, scene flow can be simply acquired by coupling stereo and optical flow, which makes the stereo matching an essential part for scene flow estimation. Most scene flow estimation methods with promising performance are initialized with a robust optical flow method or a stereo matching method. And the innovation in scene flow mostly derived from these two fields. Hence, we provides the changes and trend in the relevant issues as heuristic information.

\subsubsection{Optical flow}
%光流
Optical flow is a two-dimensional motion field. The global variational Horn-Schunck(H-S) method and the local total-least-square(TLS) Lucas-Kanade(L-K) method have led the optical flow field and scene flow field over the years~\cite{H-S,L-K}. Early works was studied and categorized by Barron and Otte with quantitative evaluation models~\cite{Barron,EPE}. Afterwards, Brox implemented the coarse-to-fine strategy to deal with large displacement~\cite{Brox}, while Sun studied the statistics of optical flow methods to find the best way for modeling~\cite{Sun}. Baker proposed a thorough taxonomy of current optical flow methods and introduced the Middlebury dataset for evaluation~\cite{Baker}, and comparisons between error evaluation methodologies, statistics and datasets are presented as well. 

Currently, optical flow estimation has reached to a promising status. A segmentation-based method with the approximate nearest neighbor field to handle large displacement ranks the top of Middlebury dataset in terms of both endpoint error(EPE) and average angular error(AAE) currently~\cite{NNF-local}, where EPE varies from 0.07$px$ to 0.41$px$ in different data and AAE varies from 0.99$^{\circ}$ to 2.39$^{\circ}$. A similar method reached promising results as well~\cite{FMOF}. Moreover, there are a variety of methods which achieve top-tier performance and solve different problems respectively. Rushwan utilized a tensor voting method to preserve discontinuity~\cite{IROF}. Xu introduced a novel extended coarse-to-fine optimization framework for large displacement~\cite{MDP}, while Stoll combines the feature matching method with variational estimation to keep small displacement area from being compromised~\cite{ALD}. He also introduced a multi-frame method utilizing trilateral filter~\cite{TC/T}. To handle non-rigid optical flow, Li proposed a Laplacian mesh energy formula which combines both Laplacian deformation and mesh deformation~\cite{LME}. 

\subsubsection{Stereo matching}
%立体匹配
Stereo matching is essential to scene flow estimation under binocular setting. A stereo algorithm generally consists of four parts:(1) matching cost computation, (2) cost aggregation, (3) estimation and optimization and (4) refinement. It is categorized into local methods and global methods depending on how the cost aggregation and computation are performed. Local methods suffer from the textureless region, while global methods are computationally expensive. A semi-global-matching(SGM) method combines local smoothness and global pixel-wise estimation and leads to a dense matching result at low runtime~\cite{SGM}, which is commonly utilized as the modification. A comprehensive review is presented by Scharstein in 2001~\cite{stereo}.

The upper rank algorithms of Middlebury stereo dataset and KITTI stereo dataset~\cite{stereo,KITTI2012} are mainly occupied by unpublished papers, indicating the rapid development in this field. Learning methods are utilized with promising efficiency and accuracy~\cite{MCCNN,CONTENT-cnn}. Besides, Zhang proposed a mesh-based approach considering the high speed of rendering and ranks the top among the published papers~\cite{meshstereo}, while segmentation-based methods are proven to tackle the textureless problem~\cite{MDP,62}.

\subsection{Challenges}
\label{subsec:problems}
The complex scene and the limited image capture approach post challenges in diverse ways, which are discussed as follows.

\subsubsection{Accuracy and efficiency}
%精度与速度
To achieve better accuracy, sufficient and complicated prior knowledge is obliged, while in terms of efficiency, the data need to be listed down to a reasonable scale and the calculation scheme should be as simple as possible. We cannot only consider the enhancement of efficiency and accuracy, but also value the trade-off between these two. The trade-off is discussed in Section~\ref{sec:discussion}, and the performance is illustrated in Figure~\ref{fig:accuracy-efficiency}. 

\subsubsection{Occlusion}
\begin{figure} [htbp]
\centering    
\subfigure[Occlusion between views] { \label{fig:a}     
\includegraphics[width=0.45\columnwidth]{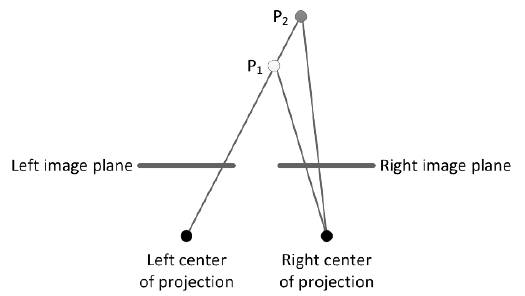}  
}     
\subfigure[Occlusion between frames] { \label{fig:b}     
\includegraphics[width=0.45\columnwidth]{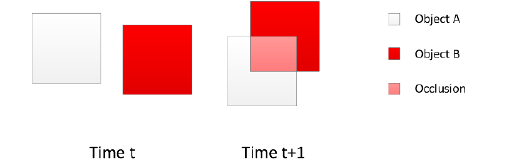}     
}     
\caption{Occlusion conditions}     
\label{fig:occlusion}     
\end{figure}

%遮挡问题
Occlusion is common in a complex scene with multiple moving objects. It occurs between views and frames as Figure~\ref{fig:occlusion} illustrates. It violates the data consistency assumption and may lead to mismatching on account of missing information of the occluded object. Besides, occlusion may perturb the consistency between frames and affect the multi-frames tracking method.

The occlusion between views can be handled well under the multi-view stereopsis with abundant prior knowledge, while temporal constraint may provide robust temporal coherence and prediction to alleviate occlusion between frames.

\subsubsection{Large displacement}
%大位移
Large displacement occurs frequently when an object is moving at a high speed or under a limited frame-rate. Moreover, articulated motion may lead to large displacement as well. This kind of problem is hard to tackle on account that the scene flow algorithms normally assume the constancy and smoothness within a small region, large displacement may make the solution to energy function trapped into a local minimum which leads to enormous errors propagated by iteration procedure. 

Brox implemented the coarse-to-fine method along with a gradient constancy assumption to alleviate the impact caused by large displacement in the optical flow field~\cite{Brox}. Currently, several matching algorithms have been introduced to handle this issue specifically and achieved promising results~\cite{70,71}.

\subsubsection{Varying illumination}
%光线变化
Brightness constancy doesn't obey the illumination-varying environment. However, this issue is common in an outdoor scene, e.g., drifting clouds that block the sunlight, sudden reflection from a window, and lens flares. The night scenario may make it worse when lights start to flash. In the optical flow field, additional assumptions such as gradient constancy and some more complicated constraints have been added to make it more robust to the illumination changes~\cite{Baker}. Schuchert specifically studied range flow estimation under varying illumination~\cite{32}. In his paper, pre-filtering and changes of brightness model improve the accuracy.  Afterwards, Gotardo introduced an albedo consistency assumption for further study~\cite{64}. A relighting procedure was proposed as a key element to handle the multiplexed situation in his paper as well. 

\subsubsection{Insufficient texture}
%低纹理
The lack of texture may make the scene flow estimation still an ill-posed problem, which is a challenge for discovering consistency. It is also a challenge to stereo matching, which may lead to enormous errors in the binocular-based scene flow estimation. The textureless region is still a major contribution to the estimation error.

To overcome this problem, different scene representations have been utilized. For example, Popham introduced a patch-based method~\cite{59}. The motion of each patch doesn't only rely on the texture information, but utilizes the motion from neighbor patches. This makes it more robust for a textureless region. As a solution to the occlusion issue, segmentation-based method is valid because it assumes uniform motion among the small regions to deal with the ambiguousness~\cite{10,25}.

\subsection{Applications}
\label{sec:applications}
%应用！！
Scene flow estimation is a comprehensive problem. Motion information reveals the temporal coherence between two moments. In a long sequence, scene flow can be utilized to get the initial value for the next frame and serve as a constraint in its relevant issue fields. Scene flow can not only profit from its relevant issues, but also facilitate them mutually. Gotardo captured three-dimensional scene flow to provide delicate geometric details~\cite{64}, while Liu utilized scene flow as a soft constraint for stereo matching and a prediction for next frame disparity estimation~\cite{25}. Ghuffar combined local estimation and global regularization in a TLS framework and utilized scene flow for segmentation and trajectory generation~\cite{55}.

Beyond that, scene flow can be a valuable input or mobile robotics and autonomous driving field, which consist of multiple task such as obstacle avoidance and scene understanding. Frank first fused optical flow and stereo by means of Kalman filter for obstacle avoidance~\cite{11}. Alcantarilla combined scene flow estimation with the visual SLAM to enhance the robustness and accuracy~\cite{41}. Herbst got object segmentation with the RGB-D scene flow estimation result~\cite{48}, aiming to achieve autonomous exploration of indoor scenes. Menze utilized scene flow to reason objects by regarding the scene as a set of rigid objects~\cite{67}. Autonomous driving could make use of both the geometric information that represents distance and the scene flow information that represents motion for multiple tasks.

In addition, scene flow can be utilized to serve as a feature as the histogram of optical flow(HOF)~\cite{HOF} or the motion boundary histogram(MBH)~\cite{MBH} feature descriptors for object detection and recognition, e.g., facial expression, gesture, and body motion recognition. It may enrich the information in the descriptor with additional depth dimension and can be applied for motion like rotation or dolly moves that optical flow can't handle. For instance, in 2009, Furukawa recorded the motion model of the facial expression using scene flow estimation\cite{27}.

%%%%%%%%
%%%%%%%%
%%%%%%%%
%%%%%%%%------------------------------Taxonomy部分----------------------------
%%%%%%%%
%%%%%%%%
%%%%%%%%
%%%%%%%%
\section{A taxonomy of scene flow estimation methods}
\label{sec:taxonomy}
\begin{figure}[htbp]
\centerline{\includegraphics[width=0.8\columnwidth]{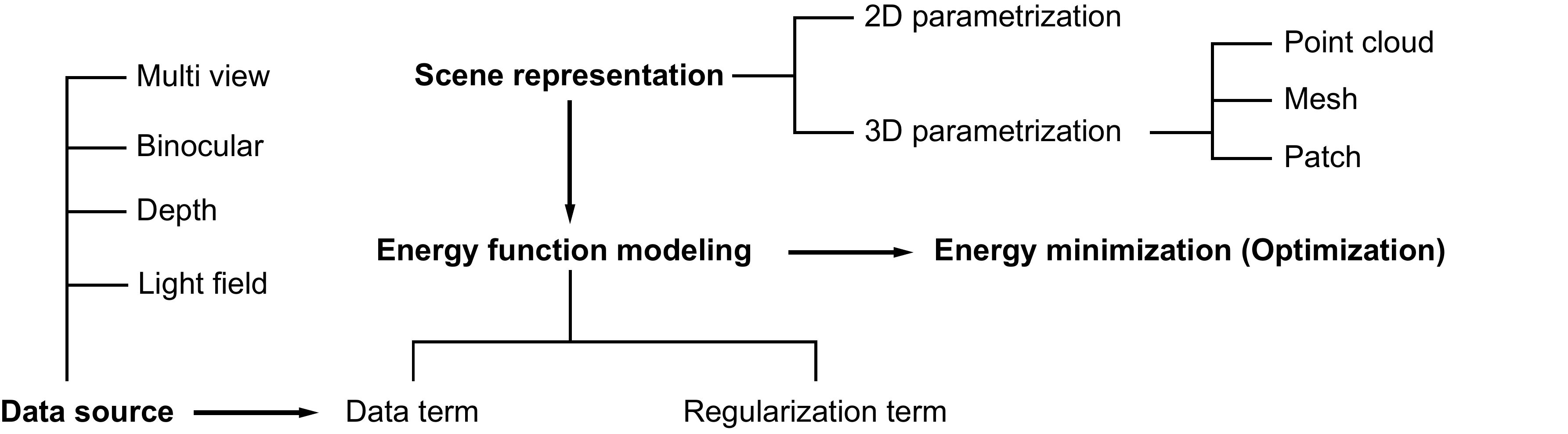}}
\caption{A general taxonomy of scene flow estimation}
\label{fig:overall}
\end{figure}

Scene flow estimation is viewed as an ill-posed problem, which includes three main steps: data acquisition, energy function modeling, energy minimization and optimization. The general taxonomy is depicted in Figure~\ref{fig:overall}. 

The energy function consists of data term and regularization as Equation~\ref{eq:energy} illustrates.
\begin{equation}
\small
\label{eq:energy}
E(\textbf{V})=E_D(\textbf{V})+\alpha E_R(\textbf{V})
\end{equation}

Data terms derive from different data sources assuming brightness constancy(BC) or gradient constancy(GC) as local constraint. While there are three unknown parameters, regularization terms need to be added to regularize the ill-posed problem and provide spatial coherence. Multiple regularization terms may make the  However, miscellaneous regularization terms may lead to redundancy, intractability and over-fitting, and that's why the design and solution of regularization term are key to a method.

Hence, in this section, existing methods are categorized in terms of three fundamental properties that distinguish the major algorithms: \emph{Scene representation} presented the diverse representations for both scene and scene flow. \emph{Data source} describes the major data acquisition manner and the corresponding data term choices. \emph{Calculation scheme} mainly discuss the idea for estimation and optimization manner, including diverse choices of regularization terms and implement .
\subsection{Scene representation}
\label{sub:representation}
%表达方式
Over the years, diverse ways to represent the scene have emerged with different emphasis, which can be broadly categorized into \emph{depth/disparity}, \emph{point cloud}, \emph{mesh} and \emph{patch}.

\subsubsection{Depth/disparity}
\label{subsubsec:2drepresentation}
%视差和深度
The convenient way to represent the scene is to couple color image and depth map as color-D information, where D stands for depth information with RGB-D data or disparity information under a binocular setting. Scene flow under this sort of representation is known as 2.5D scene flow or 2D parameterization of scene flow. It consists of optical flow component which is measured in pixels, and disparity or depth change component which is measured in pixels or $m$. The binocular-based scene flow~\cite{18,19,22,24,31,37,49,68} can be presented as $\textbf{v}=(u,v,\delta d)$, where $(u,v)$ is the 2D optical flow, and the $\delta d$ stands for disparity change. Likewise, in terms of RGB-D scene flow~\cite{48,63,65}, the motion field consists can be presented $\textbf{v}=(u,v,\Delta Z)=(u,v,W)$, where $W$ stands for the depth change.

Particularly, disparity value can be converted into depth value as Equation~\ref{eq:disparitytodepth} illustrates.
\begin{equation}
\small
\label{eq:disparitytodepth}
Z=\frac{f*b}{d}
\end{equation}
where $f$ is the focal length of the camera, and $b$ is the camera baseline value.

\subsubsection{Point cloud}
\label{subsubsec:pointcloud}
\begin{figure}[htbp]
\centering
\subfigure[The projected scene point from a image pixel] { \label{fig:3D-mapping}     
\includegraphics[width=0.4\columnwidth]{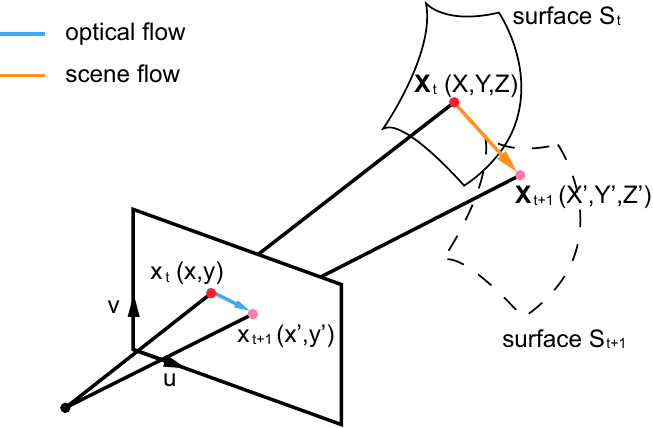}  
}
\hspace{0.1\columnwidth}    
\subfigure[The mesh representation] { \label{fig:mesh}     
\includegraphics[width=0.4\columnwidth]{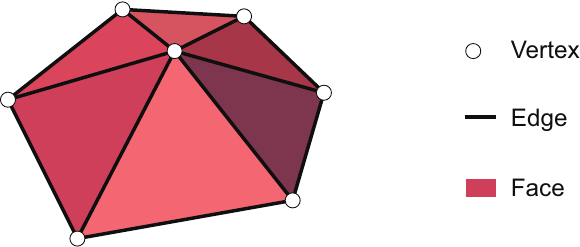}     
}     
\caption{Brief sketches about point cloud representation and mesh representation}     
\label{fig:binocular}     
\end{figure}

To truly present the three-dimensional scene, the image pixel need to be projected into the scene space. The projection is illustrated in Figure~\ref{fig:3D-mapping} and presented in Equation~\ref{eq:3D-mapping}.
\begin{equation}
\small
\label{eq:3D-mapping}
\pi^{-1}(\textbf{x},Z)=(Z\frac{x-c_x}{f_x},Z\frac{y-c_y}{f_y},Z)^T
\end{equation}
where $\textbf{x}$ is the image pixel $\textbf{x}=(x,y)$, $\pi^{-1}(\textbf{x},Z)$ is the projection $\mathbb{R}^2\times \mathbb{R}\to \mathbb{R}^3$ from a pixel value $\textbf{x}$ and a depth value $z$ to a 3D point $\textbf{X}=(X,Y,Z)$. $f_x$ and $f_y$ stands for the camera focal length, and $c_x$ and $c_y$ are the principal points.

Formulation~\ref{eq:3D-mapping} can also be presented as:
\begin{equation}
\footnotesize
\label{eq:mappingmatrix}
Z\begin{bmatrix} x\\y\\1 \end{bmatrix} =\begin{bmatrix} f_x & 0 & c_x \\ 0 & f_y & c_y \\ 0 & 0 & 1 \end{bmatrix} \begin{bmatrix} X\\Y\\Z\end{bmatrix} = A \textbf{X}
\end{equation}
where matrix $A$ is known as the camera projection matrix.

Hence, scene flow under point cloud representation~\cite{01,03,06,10,12,26,28,36,50,53,54,55,60,73} can be presented as $\textbf{V}=(\Delta X,\Delta Y,\Delta Z)=(U,V,W)$, which truly reveal the three-dimensional displacement.

\subsubsection{Mesh}
%mesh
Meshes represent a surface as a set of planar polygons, e.g., triangle, which connected to each other as is shown in Figure~\ref{fig:mesh}. This representation is a efficient way for rendering, and it occupies less memory. Mesh is essentially sort of point cloud representation as the vertex can be viewed as a point in the three-dimensional point cloud. 
Scene flow estimation methods with a mesh representation~\cite{05,07,23,27,29,34,38,44,59} are only under a multi-view setting and the geometry estimation is given simultaneously. The motion of vertice is solved with a point cloud methods, while the rest part are solved by interpolating along the meshes.

\subsubsection{Patch}
%patch
In terms of patch representation, the surface is represented by collections of small planar or sphere patches. Each patch is six-dimensional in terms of three-dimensional patch center position and three-dimensional patch direction. Patch representation is similar to mesh representation with different emphasis, where mesh representation focuses on the precision and deformable property of each vertex, and patch representation values the local consistency in terms of rigidity and motion within a small neighborhood region. 

Early paper viewed patch as a surface element(surfel) under a multi-view setting\cite{04,16}. A few binocular-based scene flow methods utilized patches to fit the surface of the scene~\cite{18,30,51,59,61,69,64} on account that this kind of patch-based methods are common in stereo matching field. In addition, Hornacek uniquely exploited a pair of RGB-D data to seek patch correspondences in the 3D world space and leads to dense body motion field including both translation and rotation~\cite{58}.

\subsection{Data source}
\label{sub:sensor}
%传感器类型
Over the years, scene flow has been estimated under three main kinds of data source: a calibrated multi-camera system, the binocular stereo camera or the RGB-D camera which consists of both RGB color information and depth information. Moreover, the emerging light field has been applied into the scene flow estimation with promising performance~\cite{68}. Data terms under different data sources differs from each other. Hence, in this section, scene flow estimation methods are categorized into these four kinds: \emph{multi-view stereopsis}, \emph{binocular setting}, \emph{RGB-D data} and \emph{light field data}.
\subsubsection{Multi view stereopsis}
%多目
Most of the algorithms in the early 2000s assume a multi-view system, with multiple cameras set in a complex calibrated scene. Multi-view scene flow estimation is usually along with 3D geometry reconstruction simultaneously. Ample data sources and diverse prior knowledge ensure the robustness of the estimation, and occlusion issue can be handled well. However, it is commonly at a high computational cost with an intricate full-view scene to deal with.

Vedula proposed two choices for regularization and distinguished three scenarios in 1999~\cite{01}, which guides the multi-view scene flow estimation till now. In his paper, a multi-view scene flow can be estimated from one optical flow and the known surface geometry. The equation is formulated in Equation~\ref{eq:vedula}.
\begin{equation}
\small
\label{eq:vedula}
\textbf{V}(\textbf{X})=\frac{\partial \textbf{X}}{\partial \textbf{x}}\textbf{v}(\textbf{x})
\end{equation}
where $\textbf{X}=(X,Y,Z)$ is the three-dimensional scene point, $\textbf{x}=(x,y)$ is the two-dimensional image pixel, $\textbf{V}(\textbf{X})$ is the scene flow, $\textbf{v}(\textbf{x})$ is the optical flow, and $\frac{\partial \textbf{X}}{\partial \textbf{x}}$ is the inverse Jacobian which can be estimated from the surface gradient $\nabla S(\textbf{X})$. 

Afterwards, Zhang proposed two systems for estimation~\cite{03,06,10}, where IMS assumed each small patch undergoes 3D affine motion, and EGS used segmentation to keep the boundary. These papers modeled energy function with multiple constraint, which provided a basic estimation process. Similarly, Pons presented a common variational framework with local similarity criteria constraint~\cite{09,20}. Henceforth, different scene representations were introduced to describe the surface~\cite{04,05,07,23,27,30,34,59}. Diverse multi-frame tracking methods mentioned for sparse estimation are utilized as well to build the temporal coherence~\cite{11,16,29,44}. Moreover, Letouzey added an RGB-D camera into the multi-view system with a mesh representation~\cite{38}, aiming to enrich the geometry information with the depth data constraint.

\subsubsection{Binocular setting}
%双目
\begin{figure}[htbp]
\centering    
\subfigure[The scene projection between views and frames] { \label{fig:binocular-a}     
\includegraphics[width=0.4\columnwidth]{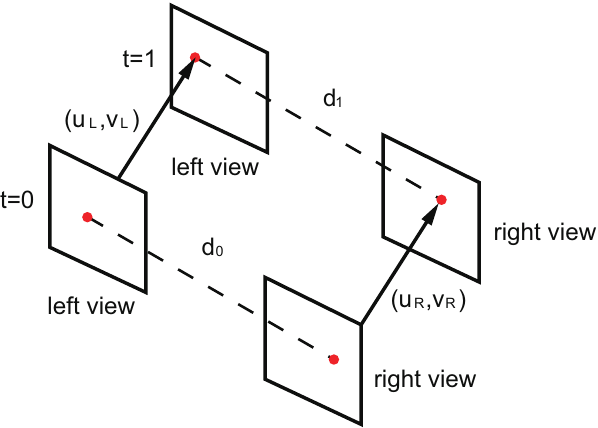}  
}     
\subfigure[The basic data terms for binocular scene flow] { \label{fig:binocular-b}     
\includegraphics[width=0.45\columnwidth]{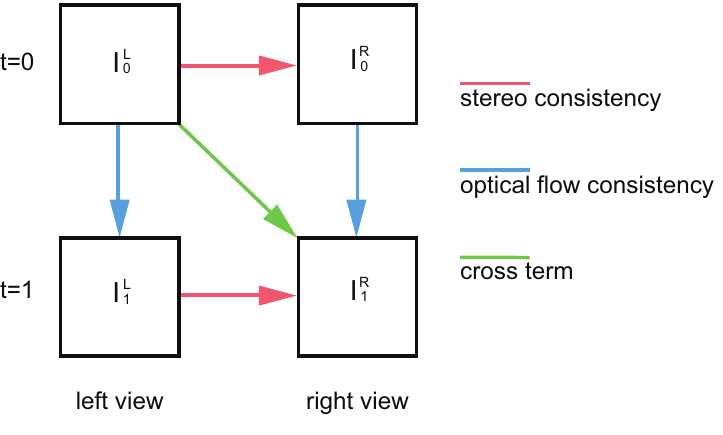}     
}     
\caption{Brief sketches about scene flow under binocular settings}     
\label{fig:binocular}     
\end{figure}

Binocular setting is regarded as a basic and simplified version of multi-view system, while the difference between the two is that binocular scene flow estimation is usually along with disparity estimation between two views but not the full-view 3D geometry knowledge. The relevance between views and frames is illustrated in Figure~\ref{fig:binocular-a}, and Figure~\ref{fig:binocular-b} depicts the basic data terms in a binocular setting which consists of stereo consistency terms in time $t$ and $t+1$ along with optical flow consistency terms in both views. The specific formulation is presented in Equation~\ref{eq:binoculardataterms}.

\begin{equation}
\footnotesize
\label{eq:binoculardataterm}
E_D=E_{fl}+\alpha E_{fr}+\beta E_{d0}+\gamma E_{dt}+\theta E_{cr}
\end{equation}
where
\begin{subequations}
\footnotesize
\label{eq:binoculardataterms}
\begin{align} 
&E_{fl}=\sum_{\Omega}\Psi(I(x+u,y+v)-I(x,y)) 
&E_{fr}=\sum_{\Omega}\Psi(I(x+u+d_{t+1},y+v)-I(x+d_t,y)) \\
&E_{d0}=\sum_{\Omega}\Psi(I(x+d_{t},y)-I(x+d_t,y))
&E_{dt}=\sum_{\Omega}\Psi(I(x+u+d_{t+1},y+v)-I(x+u,y+v)) \\
&E_{cr}=\sum_{\Omega}\Psi(I(x+u+d_{t+1},y+v)-I(x,y))
\end{align}
\end{subequations}

$E_{fl}$ and $E_{fr}$ are the optical flow consistency terms that assume the brightness of the same pixel stay constant between frames. Similarly, $E_{d0}$ and $E_{d1}$ are the stereo consistency terms that assume brightness constancy between views. Moreover, $E_{cr}$ is the cross term to constrain the constancy between both frames and views.

Most binocular-based methods fused stereo and optical flow estimation into a joint framework~\cite{12,18,19,31,37,45,49,62}. On the contrary, others decoupled motion from disparity estimation to estimate scene flow with stereo matching method replaceable at will~\cite{22,40,64,Mayer}, and Basha utilized a point cloud scene representation as a three-dimensional parametrization version of scene flow~\cite{28}. Moreover, local rigidity prior was presented along with segmentation prior and achieved promising results~\cite{39,51,61,69}. 
Specifically, Valgaerts introduced a variational framework for scene flow estimation under an uncalibrated stereo setup by embedding an epipolar constraint~\cite{33}, which makes it possible for scene flow estimation under two arbitrary cameras. In 2016, Richardt has made it a reality to compute dense scene flow from two handheld cameras with varying camera settings~\cite{73}. Scene flow was estimated under a variational framework with a DAISY descriptor~\cite{DAISY} for wide-baseline matching.

Table~\ref{tab:binoculardataterm} enumerates some typical methods under a binocular setting with diverse choices of data terms. Most methods chose optical flow consistency terms in both views and stereo consistency terms in both time $t$ and time $t+1$~\cite{12,28,39,51}, and few methods only take parts of terms mentioned above~\cite{18,22,37}. Cross term was utilized in ~\cite{37,39,49}. Moreover, Huguet~\cite{19} and Hung~\cite{49} utilized additional gradient constancy assumption besides intensity to enhance robustness against illumination changes, which turns the image intensity value $I(x,y)$ in energy function into image gradient $G(x,y)$. Additionally, extra RGB constancy terms $\frac{1}{4}(I_G-I_R)$ and $\frac{1}{4}(I_G-I_B)$ are taken in Hung's paper as well, which extends gray value intensity into three-channel information. However, it is proposed that image gradient is sensitive to noise and is view dependent~\cite{28,39}. Hence, the necessity of additional assumptions like gradient constancy remains further research to balance the pros and cons.

\begin{table}[htbp]
\footnotesize
\centering
\begin{tabular}{p{0.08\columnwidth}p{0.04\columnwidth}p{0.17\columnwidth}p{0.6\columnwidth}}
\toprule
\textbf{Literature} &\textbf{Year}& \textbf{Data term} & \textbf{Description} \\
\hline
Li\cite{12}&2005& $E_{fl},E_{fr},E_{d0},E_{dt}$& First method under a binocular setting in the early stage.\\
\hline
Isard\cite{18}&2006& $E_{fr},E_{dt},E_{cr}$& \vtop{\hbox{\strut First utilizing cross term.} \hbox{\strut Under a Markov random field(MRF) framework.}}\\
\hline
Huguet\cite{19}&2007& $E_{fl},E_{fr},E_{d0},E_{dt}$& Introducing a basic framework for scene flow under the binocular setting.\\
\hline
Wedel\cite{22}&2008& $E_{fl},E_{fr},E_{dt}$& Decoupling motion and stereo.\\
\hline
Basha\cite{28}&2010& $E_{fl},E_{fr},E_{d0},E_{dt}$& First utilizing point cloud representation.\\
\hline
Cech\cite{37}&2011& $E_{fl},E_{fr},E_{d0},E_{dt},E_{cr}$& Utilizing a seeded growing-propagation framework for fast implementation.\\
\hline
Hung\cite{49}&2013& $E_{fl},E_{fr},E_{d0},E_{dt},E_{cr}$& Assuming additional RGB intensity and gradient constancy.\\
\bottomrule
\end{tabular}
\caption{Typical methods under the binocular setting}
\label{tab:binoculardataterm}
\end{table}

\subsubsection{RGB-D data}
%深度信息
Depth was regarded as a function of space and time by Spies~\cite{02,08}. He added range flow motion constraint and introduced the range flow motion field. On basis of Spies' theory, Luckins added color channel constraint as an additional information to enhance the robustness~\cite{13}. Moreover, Schuchert added gradient constancy assumption and used pre-filtering to handle varying illumination~\cite{32}. 

\begin{figure}[htbp]
\centering
\subfigure[Depth map] { \label{fig:depth-a}     
\includegraphics[width=0.45\columnwidth]{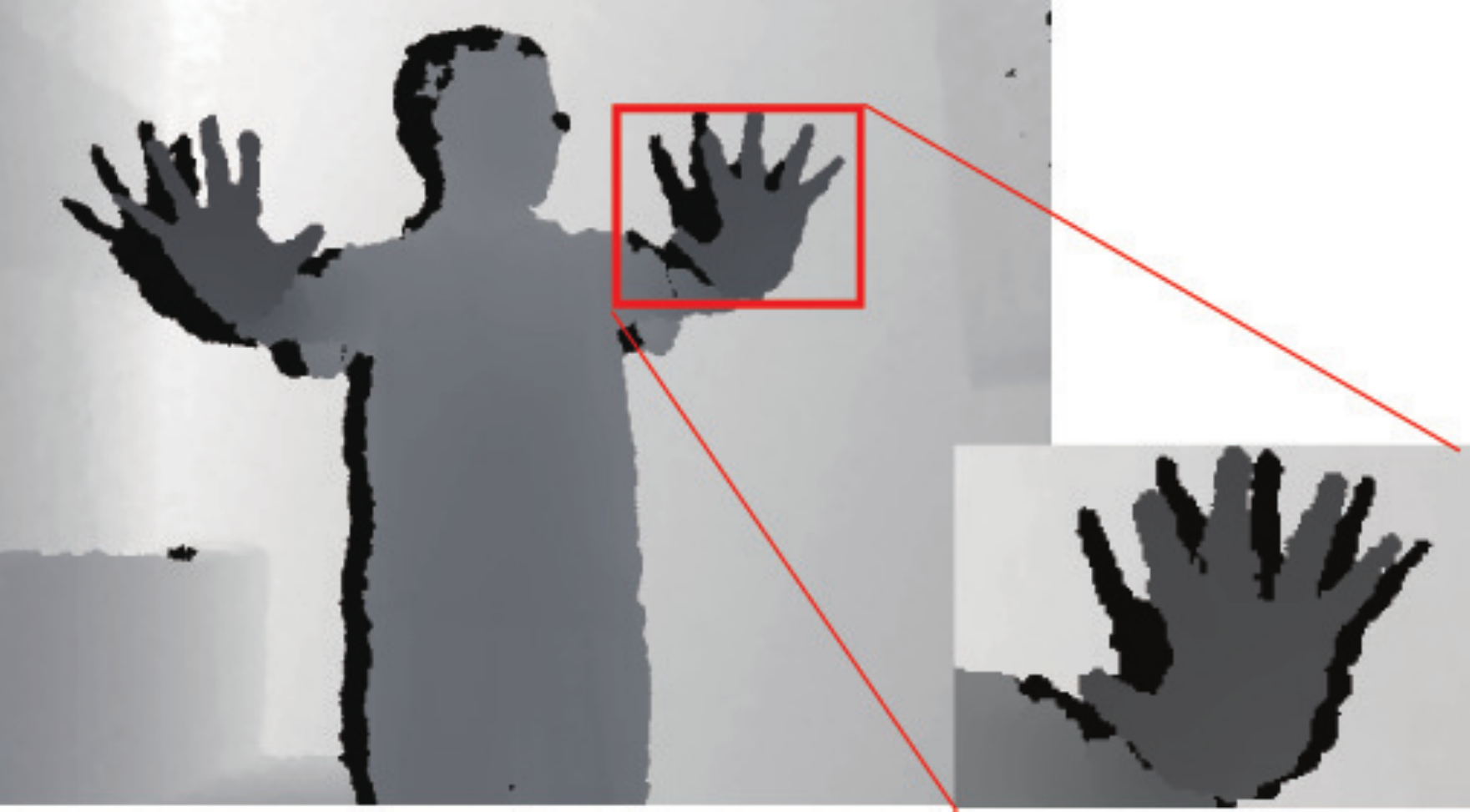}  
}     
\hspace{0.1\columnwidth}
\subfigure[Fusion of RGB image and depth map] { \label{fig:depth-b}     
\includegraphics[width=0.35\columnwidth]{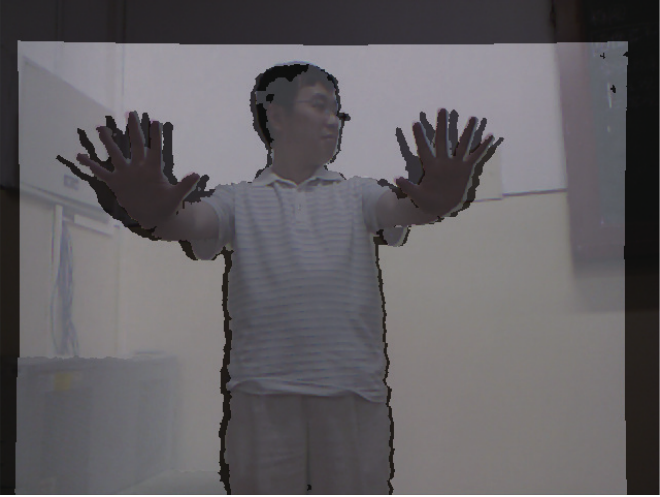}     
}   
\caption{RGB-D data acquired by Kinect sensor. It is clear that the boundary of object is occupied with many missing data colored in black due to the reflection.}
\label{fig:depthmap}
\end{figure}

With the development of RGB-D camera, depth can be acquired easily and accurately. RGB-D information can be seen as a cheap data source for geometry knowledge, and the depth information from RGB-D camera were seen as a cheap and efficient source for layering~\cite{63}, which provided layer ordering straightly without exhaustive search. However, the application of RGB-D scene flow estimation will be restrained on account of the limited sensing range and the unstable performance under illumination or reflection, which is struggling in an outdoor scene. A comparison between the state-of-art consumer RGB-D cameras is presented in Table~\ref{tab:depthsensor} to show the limitation of current RGB-D sensors in terms of range, frame rate and angle of depth measurement. Moreover, the quality of depth map is far from satisfactory due to the invalid data around object boundary region, noise and error pixels as Figure~\ref{fig:depthmap} presents.

\begin{table}[htbp]
\footnotesize
\centering
\begin{tabular}{lllll}
\toprule
\textbf{Sensor}& \textbf{Res}& \textbf{Dis}& \textbf{FPS}& \textbf{FOV(H,V)}\\
\midrule
SR4000 & 176$\times$144& 5$m$/10$m$& 50 & (43$^{\circ}$, 34$^{\circ}$)\\
\hline
Kinect v1& 640$\times$480 & 4.5$m$& 30 & (57$^{\circ}$, 43$^{\circ}$)\\
\hline
Kinect v2& 1920$\times$1080& 4.5$m$ & 30& (70$^{\circ}$, 60$^{\circ}$)\\
\hline
Camcube& 204$\times$204 & 7.5$m$ & 40 & (40$^{\circ}$, 40$^{\circ}$)\\
\bottomrule
\end{tabular}
\caption{Performance of the state-of-the-art RGB-D cameras. Notation for the header: \textbf{Res}: the maximum resolution, \textbf{Dis}: the maximum sensing range of the sensor, \textbf{FPS}: the abbreviation of frame per second, \textbf{FOV(H,V)}: the field of view in both horizontal and vertical side.}
\label{tab:depthsensor}
\end{table}

Gottfried was the first to use Kinect sensor for scene flow estimation~\cite{35}. He addressed all essential stages including calibration, alignment and estimation. Afterwards, following the common optical flow optimization, scene flow was solved under the variational framework~\cite{48,52}, and was then modified by combining additional local rigidity priors~\cite{50,60,55}. Similarly, the pixel assignment methods are becoming more and more popular~\cite{53,54,66,70}. In this way can the discontinuity be preserved smoothly. In addition, scene particle method~\cite{36,56} and feature matching method~\cite{58,71} are applied to scene flow estimation with pros and cons as supplements besides the common variational methods.

The basic data terms for RGB-D scene flow estimation consist of brightness constancy term(BC) and depth change consistency term(DCC):
\begin{equation}
\small
E_{BC}=\sum_{\Omega}\Psi(I(x+u,y+v)-I(x,y)) 
\hspace{0.05\columnwidth}
E_{DCC}=\sum_{\Omega}\Psi(Z(x+u,y+v)-Z(x,y)-W(x,y))
\end{equation}
which are utilized by most RGB-D scene flow estimation methods~\cite{02,35,48,50,53,55}. For robustness against varying illumination, Luckins added diverse additional color constraint, e.g., RGB, $l*a*b$, and hue\cite{13}, and Schuchert combining intensity constancy along with gradient constancy for per-pixel estimation~\cite{32}. Sun added layering and occlusion reasoning penalty for better performance~\cite{63}.

Table~\ref{tab:rgbddataterm} enumerates the typical RGB-D scene flow estimation methods with diverse choices of data terms. 

\begin{table}[htbp]
\footnotesize
\centering
\begin{tabular}{p{0.13\columnwidth}p{0.07\columnwidth}p{0.18\columnwidth}p{0.58\columnwidth}}
\toprule
\textbf{Literature} &\textbf{Year}& \textbf{Data term} & \textbf{Description} \\
\hline
Spies\cite{02}&2000& $E_{BC},E_{DCC}$ & Introducing the range flow.\\
\hline
Luckins\cite{13}&2005& $E_{DCC},E_{additional}$ & Additional color constraint\\
\hline
Schuchert\cite{32}&2010& $E_{BC},E_{GC},E_{DCC}$& Combining both intensity and gradient constancy constrains.\\
\hline
Gottfried\cite{35}&2011& $E_{BC},E_{DCC}$ & First utilizing consumer RGB-D camera Kinect v1.\\
\hline
Quiroga\cite{50}&2013& $E_{BC},E_{DCC}$ & Utilizing a duality-based optimization for efficient solution.\\
\hline
Sun\cite{63}&2015& $E_{BC},E_{DCC}$ & Along with penalty for layering and occlusion.\\
\bottomrule
\end{tabular}
\caption{Typical RGB-D scene flow estimation methods}
\label{tab:rgbddataterm}
\end{table}

\subsubsection{Light field data}
%光场
Light field data has enabled image refocusing and depth estimation with rich information~\cite{plenoptic}. That is to say, it can not only be treated as a depth data source, but provides much more information for constraint and regularization. Srinvasan was the first and till now the only one that utilized a light field camera Lytro Illum for scene flow estimation~\cite{68}. He proposed an oriented light-field window method as a matching manner and embedded it with the common RGB-D scene flow estimation framework, where depth data was acquired using the method proposed by Tao~\cite{lightfield}. In terms of data term, he only took brightness constancy of the oriented light field window into consideration, which is illustrated in Equation~\ref{eq:lightfielddataterm}. 
\begin{equation}
\small
\label{eq:lightfielddataterm}
E_D=\sum_{\Omega}\Psi(P_{Z_{1},x_0+u,y_0+v}(x,y,u,v,t+1)-P_{Z_{0},x_0,y_0}(x,y,u,v,t))
\end{equation}
where $P$ is the full oriented light field window operator. The penalty function is the L2 norm $\Psi(x)=x^2$.

Srinvasan's paper brings us with new ideas to extend the scene flow estimation area on account that light field data compensates the shortage of depth data source in terms of the sensing range and the robustness under an outdoor scene.

%\begin{figure}[htbp]
%\centerline{\includegraphics[width=6cm]{sample}}
%\caption{Flowchart of the scene flow estimation with a light field camera}
%\label{fig:lightfield}
%\end{figure}

\subsection{Calculation scheme}
\label{sub:method}
%求解方法
Scene flow estimation can be viewed as an energy function minimization in general, however, it is solved under different framework with diverse strategies and ideas. We categorized calculation schemes into these four main kinds: \emph{global variational method}, \emph{pixel assignment method}, \emph{feature matching method} and \emph{learning based method}. The boundaries between each kind are not always clear, but we focus on the overall solution idea to show the difference between each kind. Besides, we point out differences between each method in the same category in particular.

\subsubsection{Global variational method}
\label{subsubsec:globalvariational}
%全局局部结合方法
Global variational method has always been a classical method for either the optical flow or the scene flow estimation. As Equation~\ref{eq:energy} presents, while data terms provide local constraints to keep consistency, regularization terms give global propagation that yield a dense estimation, the motion field is constrained both locally and globally under the total variational(TV) framework.

\paragraph{Regularization term}
Under the total variational framework, a total variational(TV) regularizer is commonly chosen as the regularization term, or in other words, smoothness term. TV regularizer in terms of binocular or RGB-D can be formulated in Equation~\ref{eq:rgbdregterm}.
\begin{equation}
\small
E_R^{binocular}=\sum_{\Omega}\Psi(|\nabla u|^2+|\nabla v|^2+\lambda|\nabla d|^2+\mu |\nabla \delta d|^2)
\hspace{0.05\columnwidth}
E_R^{RGB-D}=\sum_{\Omega}\Psi(|\nabla u|^2+|\nabla v|^2+|\nabla W|^2) 
\label{eq:rgbdregterm}
\end{equation}

Besides, Basha estimated scene flow under a multi-view 3D point cloud\cite{28}. Thus the optical flow constraint was replaced by the three-dimensional scene flow constraint, and the smoothness of depth was added to the regularization term to penalize the shape. Zhang modified the regularization term with an anisotropic smoothness term to choose a more reliable pixel change between depth and appearance\cite{52}. A bilateral filtering was utilized for edge-preserving as well. And a tensor voting approach was utilized under the multi-view stereopsis~\cite{44}, where scene flow was refined in terms of direction by tensor voting in the temporal neighborhood and in terms of magnitude by a physical property between two frames.

Nevertheless, Vogel stated that TV regularizer is not good for scene flow estimation on account that it cannot handle discontinuities in the depth direction~\cite{39}. In his paper, scene flow was estimated by simultaneously regularizing the global rigid motion and the local non-rigid residual, where the regularization term consists of TV regularizer and a local rigidity prior penalized by Lorentzian function $\Psi(s)=log(1+\frac{s}{2\sigma^2})$.

\paragraph{Optimization}
Global variational methods utilizes partial differential equation to turn the energy minimization problem into a Euler-Lagrange equation solving problem, where Euler-Lagrange equation can be linearized by the Taylor expansion and then solved with iterations, and a coarse-to-fine optimization is commonly implemented during the iterations to handle large displacement.
Normally, the global variational methods are solved under two nested iterations. Taken the binocular-based scene flow as an example, the inside iteration utilizes the successive over-relaxation(SOR) iteration to compute small increments of scene flow as $(\delta u,\delta v,\delta d,\delta d')$, and the coarse-to-fine optimization serves as the outside iteration to update the warping result $u_0+\delta u,v_0+\delta v,d_0+\delta d,d_0^{'}+\delta d^{'}$ and leads to the final result $(u,v,d,d')$ for the full resolution. 

Moreover, Zach introduced a duality-based method to the optical flow estimation for optimization in 2007 and achieved real-time estimation~\cite{Zach2007}. It separates the energy function and holds them into the same framework for parallel optimization, which remarkably lowers the computational cost and complexity without accuracy loss. Following Zach's paper, Quiroga implemented an auxiliary flow in the energy function, which decompose the minimization into two simpler issues~\cite{60}. By alternating the updating of the scene flow and the auxiliary flow, the problem can be solved with great efficiency. A brief scheme is illustrated as follows:

\begin{enumerate}[\bf Step 1]
\small
\item {With auxiliary flow $\textbf{V'}$ introduced to the problem, it can be solved by alternating the updating both \textbf{V'} and $\textbf{V}$. Then the energy function defined in Equation~\ref{eq:energy} turns to: \[E(\textbf{V},\textbf{V'})=E_D(\textbf{V})+\alpha E_R(\textbf{V'})+\frac{1}{2\theta}|\textbf{V}-\textbf{V'}|^2\]}
where $\theta$ is a small constant.
\item {By fixing the scene flow $\textbf{V}$, the auxiliary flow $\textbf{V'}$ can be solved by minimizing: \[\sum_{\textbf{X}} \frac{1}{2\theta}|\textbf{V'}(\textbf{X})-\textbf{V}(\textbf{X})|^2+\alpha E_R(\textbf{V'}(\textbf{X}))\]}
\item {By fixing the auxiliary flow $\textbf{V'}$, the scene flow $\textbf{V}$ can be solved by minimizing: \[E_D(\textbf{V})+\sum_{\textbf{X}} \frac{1}{2\theta}|\textbf{V}(\textbf{X})-\textbf{V'}(\textbf{X})|^2\]}
\end{enumerate}

Similarly, Ferstl embedded primal-dual algorithm into a coarse-to-fine framework~\cite{53,54}. A total generalized variation regularization(TGV)~\cite{TGV} along with an anisotropic diffusion tensor was utilized to preserve edges. Moreover, Jaimez achieved a real-time RGB-D scene flow estimation~\cite{65}.

The brief summary of the typical global variational methods are illustrated in Table~\ref{tab:globalvariation}. 

\begin{table}[htbp]
\footnotesize
\centering
\begin{tabular}{llll}
\toprule
\textbf{Literature} &\textbf{Year}& \textbf{Regularization term} & \textbf{Optimization}\\
\hline
Huguet\cite{19}&2007& \vtop{\hbox{\strut smoothness of optical flow}\hbox{\strut smoothness of disparity and disparity change}} & SOR+coarse-to-fine\\
\hline
Gottfried\cite{48}&2013& \vtop{\hbox{\strut smoothness of optical flow}\hbox{\strut smoothness of depth}} & SOR+coarse-to-fine\\
\hline
Basha\cite{28}&2010& \vtop{\hbox{\strut smoothness of scene flow}\hbox{\strut smoothness of depth}} & SOR+coarse-to-fine\\
\hline
Jaesik\cite{44}&2012& \vtop{\hbox{\strut smoothness of scene flow magnitude}\hbox{\strut tensor voting smoothness of scene flow direction}} & - \\
\hline
Zhang\cite{52}&2013& \vtop{\hbox{\strut anisotropic smoothness of optical flow}\hbox{\strut anisotropic smoothness of depth}} & SOR+coarse-to-fine\\
\hline
Ferstl\cite{53}&2014& anisotropic TGV~\cite{TGV} regularizer & duality-based method + coarse-to-fine\\
\hline
Quiroga\cite{60}&2014& \vtop{\hbox{\strut smoothness of rotation field $\omega$}\hbox{\strut smoothness of translation field $\tau$}} & duality-based method\\
\bottomrule
\end{tabular}
\caption{The brief summary of the typical global variational methods}
\label{tab:globalvariation}
\end{table}

\subsubsection{Pixel assignment method}
\label{subsubsec:pixelassignment}
%pixel assignment
pixel assignment methods assumes local rigidity where pixels in a small region shares the same motion. It consists of three steps: pixel assignment, region motion estimation, and the compensation for each pixel. Each pixel is assigned to a specific region with prior knowledges. Then the center motion of each region is estimated, while a small motion residual of each pixel is tolerated and compensated by refinement afterwards. In this way the method combines both the global denseness and the local computational effectiveness, with discontinuity preserved well simultaneously.

Zhang first introduced this idea by fitting an affine motion model to each segment with global smoothness constraint in the early stage~\cite{06}. The scene was set under a multi-view system without rigid assumption. It is followed by Li to apply this kind of method under a binocular setting~\cite{12}.

In the last decade, this pixel-to-segment scene flow estimation draws people's attention for its superiority. Popham gave a pixel-to-patch assignment~\cite{30,59}.  The motion of each patch was estimated through a common variational method solved by Gauss-Seidel iteration, and the motion field of pixels in each patch were interpolated with a measurement covariance. Jaimez jointly estimated motion and segmentation~\cite{66}. The scene was assumed to be segmented into several labels and the pixel-to-segment issue was seen as a labelling problem, while lareweighedbelling is involved in the regularization. Scene flow for each segment is estimated with the global iteratively reweighed least squares(IRLS) minimization. Sun handled the issue with multiple hypothesis to jointly obtain occlusion reasoning, motion estimation and the scene segmentation~\cite{63}.  With a cheap acquisition of depth information, reliable layer ordering information can be obtained easily for the global-rigid, local-flexible motion field estimation.

Vogel solved the issue by simultaneously looking for a pixel-to-segment mapping and the segment motion~\cite{51}. The energy function can be formulated as:
\begin{equation}
\small
\label{eq:vogelenergy}
E(\textbf{P},\textbf{S})=E_D(\textbf{P},\textbf{S})+\lambda E_R(\textbf{P},\textbf{S}) + \mu E_S(\textbf{S})
\end{equation}
where $\textbf{S}: I\to S$ is the pixel-to-segment mapping  which assign each image pixel $\textbf{p}\in I$ to a segment $s \in S$. $\textbf{P}: S\to \Pi$ is the segment motion mapping  which assign each segment to a 3D rigidly moving plane $\pi \in \Pi$. $E_D$ is the data term and $E_R$ is the TV regularizer, and $E_S$ is an additional segmentation regularization term to refine segmentation during iterations.

\begin{figure}[htbp]
\centerline{\includegraphics[width=0.8\columnwidth]{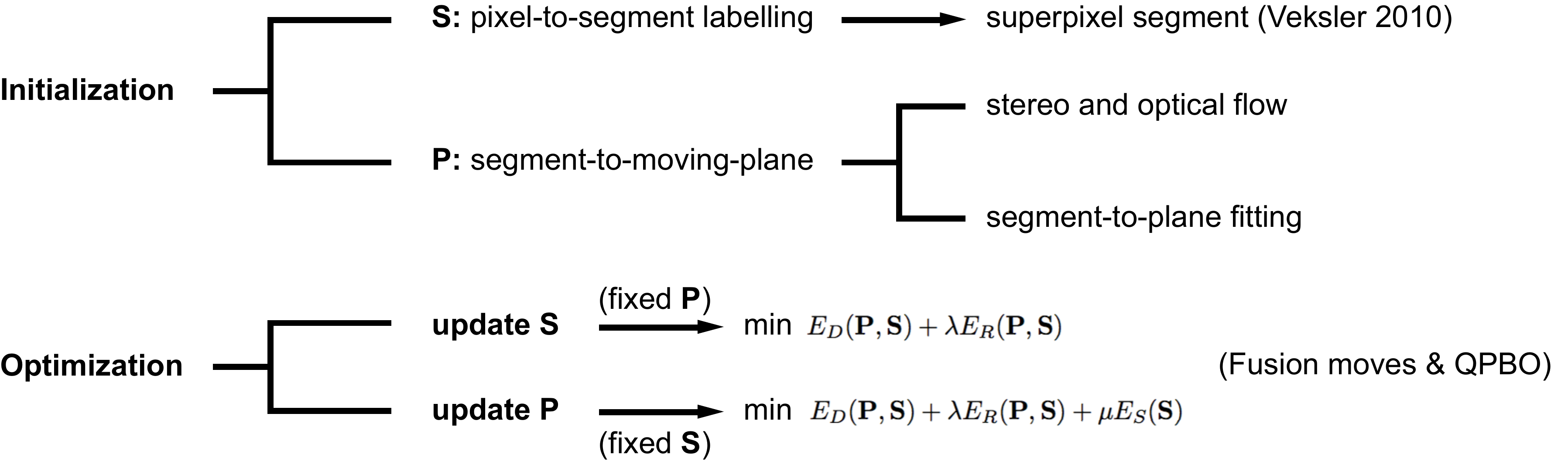}}
\caption{Flowchart of the pixel-to-segment assignment scene flow estimation~\cite{51}}
\label{fig:vogelchart}
\end{figure}

Using a superpixel segmentation for initialization~\cite{superpixel}, the energy is alternatively optimized using fusion moves~\cite{fusion} and quadratic pseudo-boolean optimization (QPBO)~\cite{QPBO}, and it achieves the state-of-the-art performance. The general idea is depicted in Figure~\ref{fig:vogelchart}. Afterwards, he introduced a temporal window to enforce coherence over long time intervals~\cite{61}, and proposed a detailed version with deep analysis and thorough comparisons later that forms the whole theoretical framework~\cite{69}.

Similarly, in 2016, Lv follows the idea by representing the dynamic 3D scene as a collection of rigidly moving planar segments~\cite{74}. The complex assignment problem is formulated with a factor graph formulation~\cite{factorgraph}, estimated as a non-linear least square problem and then optimized locally and globally.

The brief summary of the typical pixel assignment methods are illustrated in Table~\ref{tab:pixelassignment}. 

\begin{table}[htbp]
\footnotesize
\centering
\begin{tabular}{llll}
\toprule
\textbf{Literature} &\textbf{Year}& \textbf{Assignment} & \textbf{Optimization}\\
\hline
Vogel\cite{51}&2013& color pixel-to-segment (superpixel segmentation) & fusion moves + QPBO\\
\hline
Popham\cite{59}&2014& color pixel-to-patch (belief propagation) & Gauss-Newton minimization\\
\hline
Sun\cite{63}&2015& depth layering (K-means) & coordinate descent minimization\\
\hline
Jaimez\cite{66}&2015& color pixel-to-segment (K-means) &  IRLS + coarse-to-fine method\\
\hline
Lv\cite{74}&2016& color pixel-to-segment (superpixel segmentation) & Levenberg-Marquardt\\
\bottomrule
\end{tabular}
\caption{The brief summary of the typical pixel assignment methods}
\label{tab:pixelassignment}
\end{table}

\subsubsection{Feature matching method}
\label{subsubsec:featurematching}
Feature matching method mainly consists of three steps: feature extraction, feature matching, and propagation. While stereo estimation aims to find the spatial correspondence, scene flow seeks for temporal coherence. Thus, this kind of feature matching methods for scene flow estimation is suitable for binocular-based methods with stereo estimation solved simultaneously.  Cech~\cite{37} made use of Harris points as the feature and a multi-scale LK tracker~\cite{L-K} as a prematcher. The sparse estimation was then propagated to dense result using seeded growing method in terms of both stereo and motion. Sizintsev introduced a spatiotemporal quadric element(stequel) matching method~\cite{45}. Stereo correspondence and scene flow were estimated through the match cost solution with multiple constraints. Richardt utilized the DAISY descriptor~\cite{DAISY} for computing correspondence between views and frames~\cite{73}, and the dense scene flow was refined under the conventional variational framework.

Moreover, matching method is also implemented with an RGB-D camera. Letouzey matched SIFT features as a constraint and minimized the energy function as a preliminary estimation~\cite{38}. By re-projecting this preliminary estimation into the image plane, a preliminary map can be viewed as the initial value, where the SIFT features are the non-moving anchor points. Quiroga enforced consistency of scene flow with a sparse set of SURF features~\cite{50}. The features were extracted in the color image while depth information was taken as the matching constraint. Hornacek proposed a patch-wise estimation without assuming brightness constancy~\cite{58}. The matching cost takes both three-channel CIE l*a*b information and gradient information into consideration, initialized by SURF features. The pixel with tiny matching cost will be viewed with the same motion for propagation. On basis of Hornacek, Alhaija took edge as the sparse matching feature with a graph matching approach to handle large displacement and acquired promising results~\cite{71}.

The brief summary of the typical feature matching methods are illustrated in Table~\ref{tab:featurematching}. 

\begin{table}[htbp]
\footnotesize
\centering
\begin{tabular}{lllll}
\toprule
\textbf{Literature} &\textbf{Year} & \vtop{\hbox{\strut \textbf{Feature}} \hbox{\strut \textbf{Function}}} & \textbf{Regularization term} & \textbf{Optimization}\\
\hline
Cech\cite{37}& 2011 & \vtop{\hbox{\strut SIFT features}\hbox{\strut seed for propagation}} & \vtop{\hbox{\strut smoothness of optical flow}\hbox{\strut smoothness of disparity}} & - \\
\hline
Letouzey\cite{38}& 2011 & \vtop{\hbox{\strut Harris point}\hbox{\strut sparse matching constraint}} & smoothness of scene flow & Jacobi minimization \\
\hline
Quiroga\cite{50}& 2013 & \vtop{\hbox{\strut SURF features}\hbox{\strut sparse matching constraint}} & \vtop{\hbox{\strut smoothness of scene flow}\hbox{\strut smoothness of depth}} & duality-based method\\
\hline
Hornacek\cite{58}&2014& \vtop{\hbox{\strut SURF features}\hbox{\strut for patch generation}} & smoothness of scene flow & fusion moves + QPBO\\
\hline
Alhaija\cite{71}& 2015 &  \vtop{\hbox{\strut SIFT features}\hbox{\strut for edge descriptor generation}} & \vtop{\hbox{\strut smoothness of scene flow}\hbox{\strut projection consistency term}} & fusion moves + QPBO\\
\hline
Richardt\cite{73}& 2017 &  \vtop{\hbox{\strut DAISY~\cite{DAISY}}\hbox{\strut sparse matching initialization}} & \vtop{\hbox{\strut smoothness of optical flow}\hbox{\strut smoothness of disparity change}\hbox{\strut epipolar constraint}} & PatchMatch belief propagation\\
\bottomrule
\end{tabular}
\caption{The brief summary of the typical feature matching methods}
\label{tab:featurematching}
\end{table}

\subsubsection{Learning-based method}
\label{subsubsec:learning}
%Learning
Learning method has got enough attention over the past few years for solving computer vision tasks. To solve scene flow estimation, learning method can be utilized twofold. On the one hand, parameters as a part of the whole pipeline can be learned to enhance the robustness or the efficiency. Hadfield, for example, used machine learning technique to introduce an intelligent cost function as a penalization metric with limited improvement~\cite{57}. 

On the other hand, learning can be utilized for a per-pixel end-to-end estimation. Currently speaking, due to lack of large-scale dataset with ground truth and appropriate models, Mayer is the only one who utilized convolutional neural network(CNN)~\cite{CNN} to learn scene flow estimation~\cite{Mayer}, which reveals the embryo of learning scene flow. He also introduced a large scale dataset for training, which will be introduced in Section~\ref{subsec:Freiburg}. Optical flow estimation and disparity estimation were decoupled, where the disparity estimation network named "\emph{DispNet}" was presented on basis of the proposed FlowNet optical flow network~\cite{FlowNet}. Each network consisted of a contracting part for feature contraction and an expanding part utilizing up-convolutional layers and un-pooling for final estimation. A loss weight scheme was implemented to balance the weight between high resolution and low resolution. The downsampling factor in total is 64, and the network consider a maximum displacement of 160 pixels in the input images, which is much superior to a 4-level pyramid with a 50\% downsampling factor. 
%%%%%%%%
%%%%%%%%
%%%%%%%%
%%%%%%%%------------------------------dataset部分----------------------------
%%%%%%%%
%%%%%%%%
%%%%%%%%
%%%%%%%%
\section{Evaluation}
\label{sec:evaluation}
In this section, we enumerate and analyze the regular evaluation protocols and publicly-available datasets.
\subsection{Evaluation protocols}
\label{subsec:evaluationprotocol}

Scene flow is made up of 3D motion vectors, which contain range and direction information of displacement. Thus, the error is measured twofold in terms of magnitude and angle. Currently, the error is measured in three ways: the end-point error(EPE), the root mean square error(RMSE), and the average angular error(AAE). A brief introduction is presented as follows.

\subsubsection{Two-dimensional error measure}
\label{subsubsec:2derror}
%二维误差表示
Due to the fact that main datasets only provide ground truth under a two-dimensional representation in terms of optical flow and disparity, most methods re-project scene flow onto the image space or simply represent scene flow with a two-dimensional representation mentioned in Section~\ref{subsubsec:2drepresentation}, and the error is measured  in terms of optical flow and disparity. But first, we'll present the fundamental error measure named absolute error:

\paragraph{Absolute error}
Absolute error describes the absolute magnitude difference, which is the euclidean distance between the endpoints of ground truth vector and the estimated vector. It is presented in terms of optical flow and disparity in Equation~\ref{eq:absoluteerror}.
\begin{equation}
\small
E_{of}=\left|(\textbf{v}_e-\textbf{v}_g)\right|=\sqrt{(u_e-u_g)^2+(v_e-v_g)^2}
\hspace{0.1\columnwidth}
E_{disp}=\left|(d_e-d_g)\right| 
\label{eq:absoluteerror}
\end{equation}
where subscript $e$ annotates the estimated value, and subscript $g$ annotates the ground truth.

\paragraph{The average endpoint error(EPE)}
%EPE
EPE is introduced by Otte~\cite{EPE}. It's the mean of absolute error among all the pixels as Equation~\ref{eq:EPE} presents.
\begin{equation}
\small
E_{EPE_{of}}=\frac{1}{n}\sum_{i \in \Omega}E_{of}^i=\frac{1}{n}\sum_{i\in \Omega}\sqrt{(u_e^i-u_g^i)^2+(v_e^i-v_g^i)^2}
\hspace{0.05\columnwidth}
E_{EPE_{disp}}=\frac{1}{n}\sum_{i \in \Omega}E_{disp}^i=\frac{1}{n}\sum_{i\in \Omega}\left|(d_e^i-d_g^i)\right| 
\label{eq:EPE}
\end{equation}
where $n$ is the number of pixels, and $\Omega$ is the entire image plane.

\paragraph{The root mean square error(RMSE)}
%RMSE
\renewcommand{\thefootnote}{1}
While EPE indicates the overall accuracy level, RMSE indicates both the error distribution and overall accuracy level as $E_{RMSE}^2=\sigma_{AE}^2+E_{EPE}^2$ \footnotemark[1].
%RMSE和EPE,\SIGMA的关系推导
\footnotetext[1]{
\tiny
$\sigma_{AE}^2=\frac{1}{n}\sum_{i\in \Omega}[\sqrt{(u_e^i-u_g^i)^2+(v_e^i-v_g^i)^2}-E_{EPE}]^2=\frac{1}{n}\sum_{i\in \Omega}[(u_e^i-u_g^i)^2+(v_e^i-v_g^i)^2-2E_{EPE}\sqrt{(u_e^i-u_g^i)^2+(v_e^i-v_g^i)^2}+E_{EPE}^2]$

$=\frac{1}{n}\sum_{i\in \Omega}[(u_e^i-u_g^i)^2+(v_e^i-v_g^i)^2]-\frac{2}{n}E_{EPE}\sum_{i\in \Omega}\sqrt{(u_e^i-u_g^i)^2+(v_e^i-v_g^i)^2}+\frac{1}{n}\sum_{i\in \Omega}E_{EPE}^2$

$=E_{RMSE}^2-2E_{EPE}^2+E_{EPE}^2=E_{RMSE}^2-E_{EPE}^2$}

RMSE in terms of optical flow and disparity can be presented as Equation~\ref{eq:RMSE-of} and ~\ref{eq:RMSE-disp}.
\begin{subequations}
\small
\begin{align} 
&E_{RMSE_{of}}=\sqrt{\frac{1}{n}\sum_{i\in \Omega}{E_{of}^i}^2}=\sqrt{\frac{1}{n}\sum_{i\in \Omega}((u_e^i-u_g^i)^2+(v_e^i-v_g^i)^2)} \label{eq:RMSE-of}\\
&E_{RMSE_{disp}}=\sqrt{\frac{1}{n}\sum_{i\in \Omega}{E_{disp}^i}^2}=\sqrt{\frac{1}{n}\sum_{i\in \Omega}\left|(d_e^i-d_g^i)\right|^2} \label{eq:RMSE-disp}
\end{align}
\end{subequations}

In addition, to better measure the error in different scales, the normalization is needed. The normalized root mean square error(NRMSE) is scaled by the ground truth, which can be compared between different datasets. NRMSE in terms of optical flow and disparity can be presented as Equation~\ref{eq:NRMSE}.
\begin{equation}
\small
E_{NRMSE_{of}}=\frac{E_{RMSE_{of}}}{max\left\| \textbf{v}_g \right\|-min\left\| \textbf{v}_g \right\|} 
\hspace{0.05\columnwidth}
E_{NRMSE_{disp}}=\frac{E_{RMSE_{disp}}}{max\left| d_g \right|-min\left| d_g \right|} 
\label{eq:NRMSE}
\end{equation}

\paragraph{The average angular error(AAE)}
%AAE
The average angular error was introduced by Fleet in 1990~\cite{AAE}, which is introduced to measure the optical flow error deviation of angle:
\begin{equation}
\small
\label{eq:AAE-of}
E_{AAE_{of}}=\frac{1}{n}\sum_{i\in \Omega}arctan(\frac{u_e^i v_g^i-u_g^i v_e^i}{u_e^i u_g^i+v_e^i v_g^i})
\end{equation}

It can also be calculated with $arccos(\textbf{v}_g \cdot \textbf{v}_e)$ or $arcsin(\textbf{v}_g \times \textbf{v}_e)$.

\subsubsection{Three-dimensional error measure}
%三维误差表示
According to Equation~\ref{eq:3D-mapping}:
\begin{equation}
\small
\label{eq:3Derror}
(X,Y,Z)=(\frac{(x-c_x)b}{d}, \frac{(y-c_y)b}{d}, \frac{f*b}{d})
\hspace{0.05\columnwidth}
(X+U,Y+V,Z+W)=\omega_c((x+u-c_x), (y+v-c_y), f)
\end{equation}
where $\omega_c=\frac{b}{d+\delta d}$. The focal length of each camera is assumed the same, as $f_x=f_y=f$.

Hence, the transfer function between the three-dimensional point cloud scene flow representation and two-dimensional representation can be illustrated in Equation~\ref{eq:errorprojection}.
\begin{equation}
\small
\label{eq:errorprojection}
(U,V,W)=\omega_c(u-\frac{\delta d(x-c_x)}{d},v-\frac{\delta d(y-c_y)}{d},-\frac{f*\delta d}{d})
\end{equation}

We can see clearly that the projection between scene flow and optical flow along with disparity and disparity change information is complex, and the accuracy of disparity change $\delta d$ can significantly affect the result. That is to say, the protocol mentioned in Section~\ref{subsubsec:2derror} is not sufficient enough to evaluate scene flow. 

Wedel proposed a easy way for three-dimensional error measure\cite{22}. The RMSE and AAE are modified as Equation~\ref{eq:NRMSE-wedel} and ~\ref{eq:AAE-wedel}.
\begin{subequations}
\small
\begin{align} 
&E_{RMSE_{wedel}}=\sqrt{\frac{1}{n}\sum_{i\in \Omega}\left\|(u_e^i,v_e^i,d_e^i,\delta d_e^i)-(u_g^i,v_g^i,d_g^i,\delta d_g^i) \right\|} \label{eq:NRMSE-wedel}\\
&E_{AAE_{wedel}}=\frac{1}{n}\sum_{i \in \Omega}arccos(\frac{u_e^i u_g^i+v_e^i v_g^i+\delta d_e^i \delta d_g^i+1}{\sqrt{({u_e^i}^2+{v_e^i}^2+\delta {d_e^i}^2+1)({u_g^i}^2+{v_g^i}^2+\delta {d_g^i}^2+1)}}) \label{eq:AAE-wedel}
\end{align}
\end{subequations}

However, this may not precisely reveal the contribution to error measure between different unknowns. Basha provided a three-dimensional point cloud ground truth $\textbf{V}_g=(U_g,V_g,W_g)$, which made the three-dimensional error measure feasible~\cite{28}. The absolute error can be measured in a three-dimensional way:
\begin{equation}
\small
\label{eq:absoluteerrorsf}
E_{sf}=\left|(\textbf{V}_e-\textbf{V}_g)\right|=\sqrt{(U_e-U_g)^2+(V_e-V_g)^2+(W_e-W_g)^2}
\end{equation}

Hence, the EPE, RMSE, NRMSE, and AAE can be modified as:
\small
\[
E_{EPE_{sf}}=\frac{1}{N}\sum_{P \in \mathbb{S}}E_{sf}^P 
\hspace{0.03\columnwidth}
E_{RMSE_{sf}}=\sqrt{\frac{1}{N}\sum_{P\in \mathbb{S}}{E_{sf}^P}^2} 
\hspace{0.03\columnwidth}
E_{NRMSE_{sf}}=\frac{E_{RMSE_{sf}}}{max\left\| \textbf{V}_g \right\|-min\left\| \textbf{V}_g \right\|}
\]
\begin{equation}
\small
\label{eq:AAE_{sf}}
E_{AAE_{sf}}=\frac{1}{N}\sum_{P \in \mathbb{S}}arccos(\frac{U_e^P U_g^P+V_e^P V_g^P+W_e^P W_g^P+1}{\sqrt{({U_e^P}^2+{V_e^P}^2+{W_e^P}^2+1)({U_g^P}^2+{V_g^P}^2+{W_g^P}^2+1)}})
\end{equation}
where $\mathbb{S}$ is the surface in the three-dimensional space.

\subsubsection{Special metrics}
\label{subsubsec:specific}
Besides the protocols mentioned above, KITTI dataset~\cite{KITTI2012} and Sintel dataset~\cite{Sintel} analyzes EPE under different circumstances and introduced the special metric for evaluation.
\paragraph{KITTI metric}
KITTI metric is the specific criterion for performance evaluation on KITTI dataset~\cite{KITTI2012}. It employed an EPE threshold of  $\tau \in (2,\cdots, 5)$ pixel, and calculated the portion of pixels whose endpoint error in terms of optical flow and disparity is above the threshold among the entire image (threshold default is set as 3$px$). With occlusion ground truth, the official protocol is presented as:

\begin{table}[htbp]
\footnotesize
\centering
\begin{tabular}{c  c  c  c  c  c  c  c  c}
\hline
 {\bf Method} & {\bf Setting} & {\bf Out-Noc} & {\bf Out-All} & {\bf Avg-Noc} & {\bf Avg-All} & {\bf Density} & {\bf Runtime} & {\bf Environment}\\
 \hline
\end{tabular}
\end{table}

In 2015, Menze modified the dataset by adding background and foreground annotation~\cite{67}, and percentage of outliers in terms of background and foreground are distinguished in the evaluation metric. Moreover, a scene flow error was introduced. if either the disparity or the optical flow end-point error is above the threshold(where the default is 3 pixels), then the pixel is viewed as the scene flow outlier. The official protocol is presented as:

\begin{table}[htbp]
\footnotesize
\centering
\begin{tabular}{lllllllllllll}
\hline
{\bf Method} & {\bf D1-bg} & {\bf D1-fg} & {\bf D1-all} & {\bf D2-bg} & {\bf D2-fg} & {\bf D2-all} & {\bf Fl-bg} & {\bf Fl-fg} & {\bf Fl-all} & {\bf SF-bg} & {\bf SF-fg} & {\bf SF-all}\\ 
\hline
\end{tabular}
\end{table}

\paragraph{Sintel metric}
Sintel metric provides a thorough evaluation for Sintel benchmark~\cite{Sintel}. It employed EPE as error measure as well. Particularly, it measure the error distribution in terms of both occlusion and large displacement with different threshold,  which clearly reveal the performance under these two fundamental challenges. Moreover, the percentage of error pixels that remain visible in adjacent frames are taken as a criterion that reveal the temporal distribution of error. Thus, this metric provides sufficient information for evaluation and comparison. The official protocol is presented as:

\begin{table}[htbp]
\footnotesize
\centering
\begin{tabular}{c  c  c  c  c  c  c  c  c c}
\hline
 {\bf Method} & {\bf EPE all} & {\bf EPE matched} & {\bf EPE unmatched} & {\bf d0-10} & {\bf d10-60} & {\bf d60-140} & {\bf s0-10} & {\bf s10-40} & {\bf s40+}\\
 \hline
\end{tabular}
\end{table}

\subsection{Color decoding for visualization}
For better evaluating and analyzing scene flow performance, the color decoding manners are proposed for visualizing optical flow and error map, which make the it clear and direct to quantitatively evaluate the performance.

\begin{figure}[htbp]
\centering    
\subfigure[The HSV cone model] { \label{fig:HSV-a}     
\includegraphics[width=0.3\columnwidth]{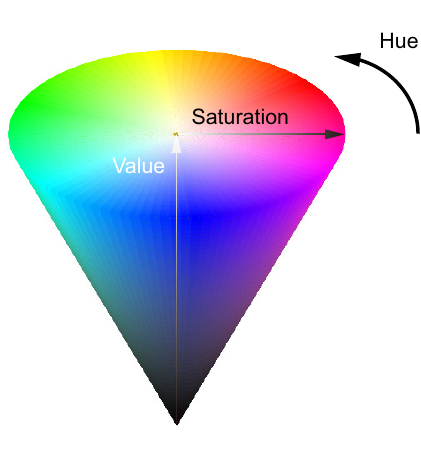}  
}     
\hspace{0.05\columnwidth}
\subfigure[An example of Monkaa optical flow ground truth colorization] { \label{fig:HSV-b}     
\includegraphics[width=0.6\columnwidth]{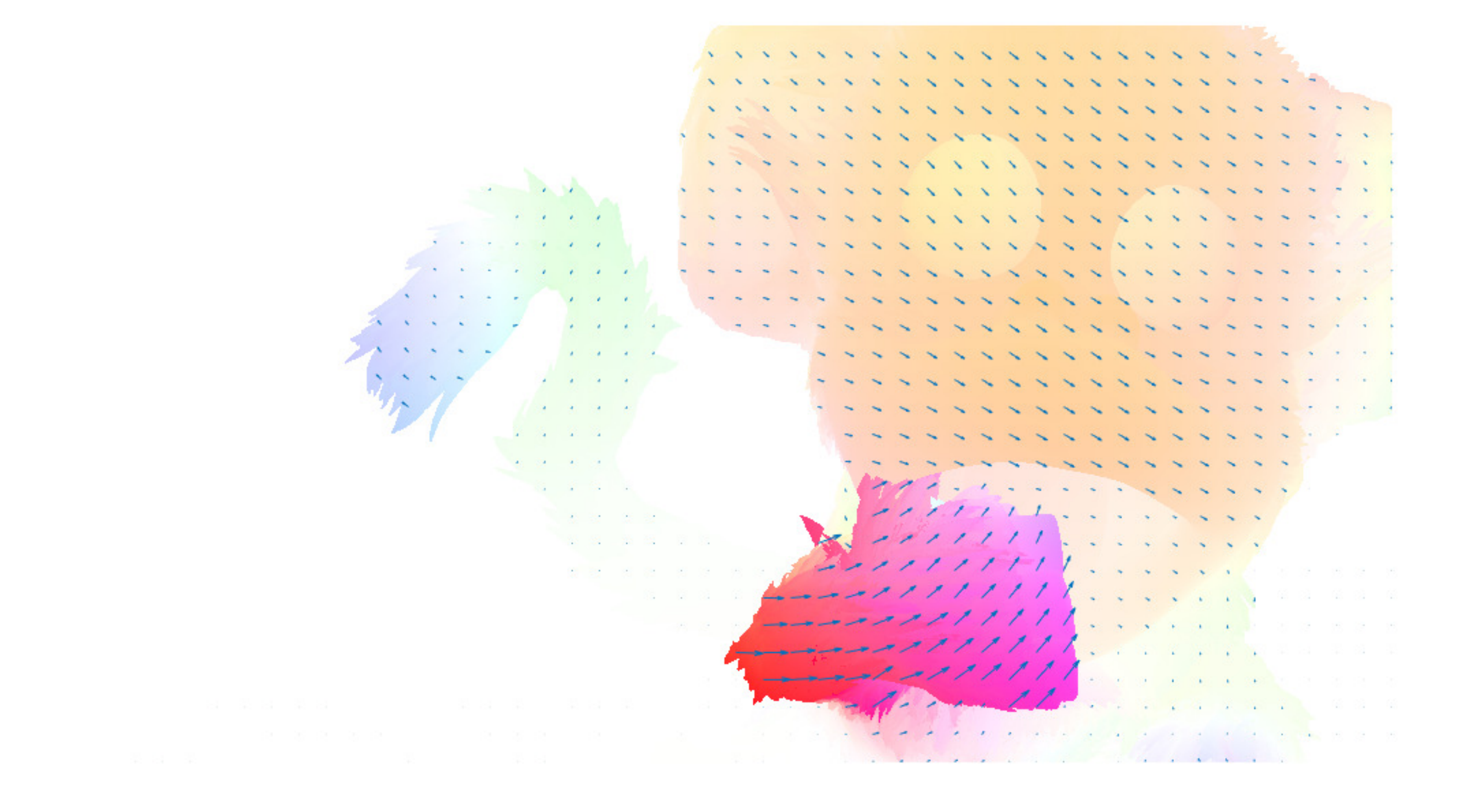}     
}     
\caption{Brief sketches about optical flow colorization}     
\label{fig:HSV}     
\end{figure}

\paragraph{Optical flow}
Optical flow is a motion field composed of two-dimensional vectors. Normally the optical flow are encoded with an HSV color space. "H" stands for hue, of which the range is $[0,360^{\circ}]$. We use the angular between the optical flow vector and x axis in the image plane as the hue value. "S" stands for saturation, which is represented by the ratio between the magnitude of each vector and the maximum magnitude among the whole motion field. "V" stands for value. Normally the value in non-occluded area is set as $0$, while the value in occluded area is set as $1$. In this way the motion vectors in the same direction share the same hue with different saturation to indicate the magnitude degree, the non-moving region is pure white, and the occluded region is black. The HSV model for optical flow colorization is illustrated in Figure~\ref{fig:HSV}.

\begin{figure}[htbp]
\centering    
\subfigure[Error map visualization. Each pixel in the error map is colorized by finding the specific interval where the EPE value is in, and the color is chosen with RGB value corresponding to the interval as illustrated below.] { \label{fig:errorcolorization-a}     
\includegraphics[width=0.95\columnwidth]{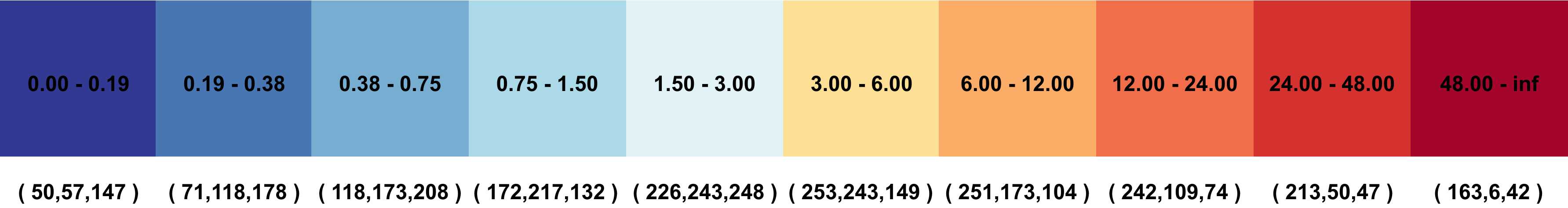}  
}     
\subfigure[An example~\cite{69} of scene flow error colorization in KITTI 2015 dataset] { \label{fig:errorcolorzation-b}     
\includegraphics[width=0.75\columnwidth]{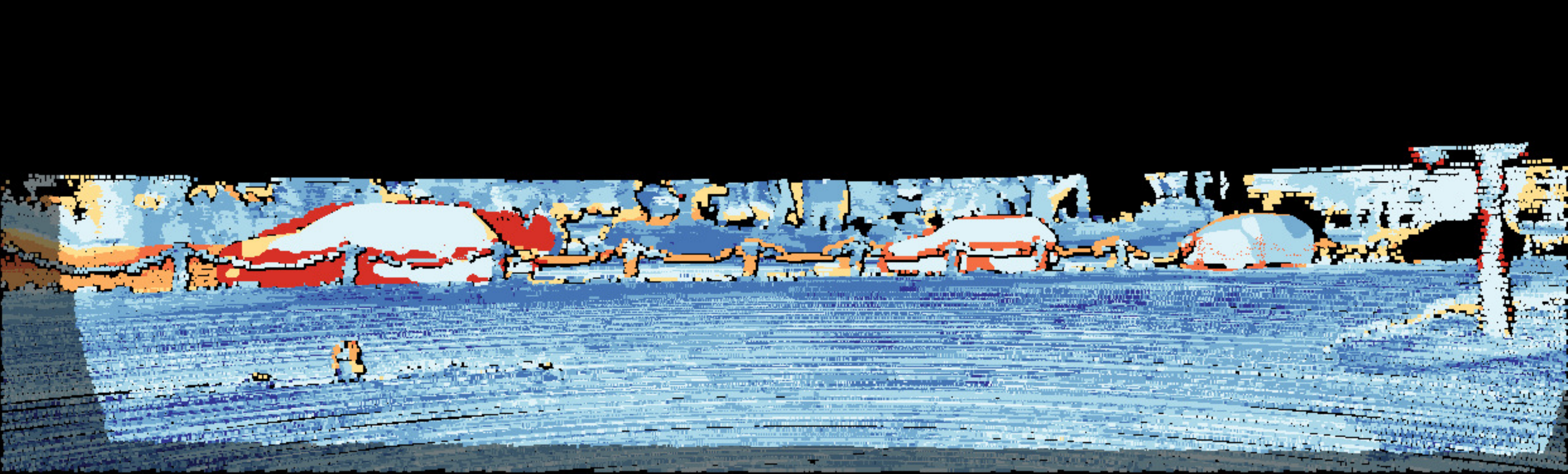}     
}     
\caption{Brief sketches about error colorization map}     
\label{fig:errorcolorization}     
\end{figure}

\paragraph{Error map}
Menze introduced an error visualization to directly distinguish the error distribution in the image plane. The correct pixel is depicted in blue while the error pixels are depicted in red. In this way can we directly find the rule for major error distribution. The corresponding relation between color and endpoint error for each pixel is presented in Figure~\ref{fig:errorcolorization}.

\subsection{dataset}
\label{sec:datasets}

Existing datasets mainly serve for scene flow evaluation under a binocular setting, which consist of optical flow ground truth and disparity ground truth. This can also be used for evaluating stereo and optical flow estimation. Due to the lack of RGB-D dataset, to evaluate RGB-D scene flow, the disparity map of each dataset should be converted to the depth map with Equation~\ref{eq:disparitytodepth} as an input. 

As is mentioned in Section~\ref{sub:representation}, scene flow can be represented as $\textbf{v}(u,v,\Delta d)$ or $\textbf{V}(\Delta X,\Delta Y,\Delta Z)$. While most datasets only provided optical flow and disparity ground truth, Basha~\cite{28} provided the three-dimensional ground truth $\textbf{V}(\Delta X,\Delta Y,\Delta Z)$, and Freiburg dataset~\cite{Mayer} provided additional disparity change ground truth that truly represent scene flow. Moreover, Middlebury~\cite{middlebury01,middlebury02}, Basha~\cite{28} and KITTI~\cite{KITTI2012,67} provide occlusion ground truth so that the error out of occluded region can be evaluated separately. 

\begin{figure*}
\centering
\subfigure[Middlebury 2003 - Cones]
{
	\begin{minipage}[t]{0.25\textwidth}
	\includegraphics[width=1\textwidth]{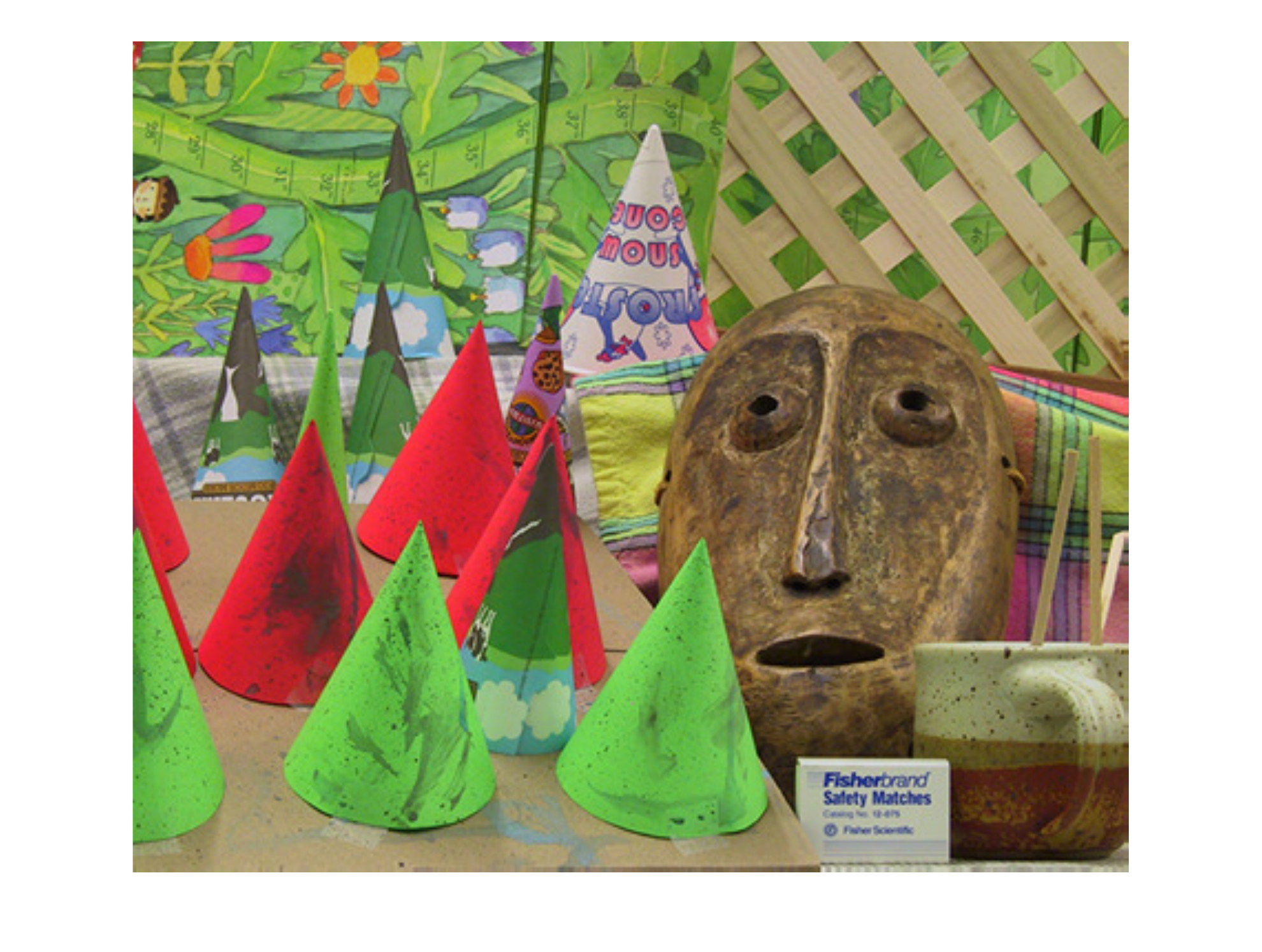} 
	\end{minipage}
	\begin{minipage}[t]{0.25\textwidth}
	\includegraphics[width=1\textwidth]{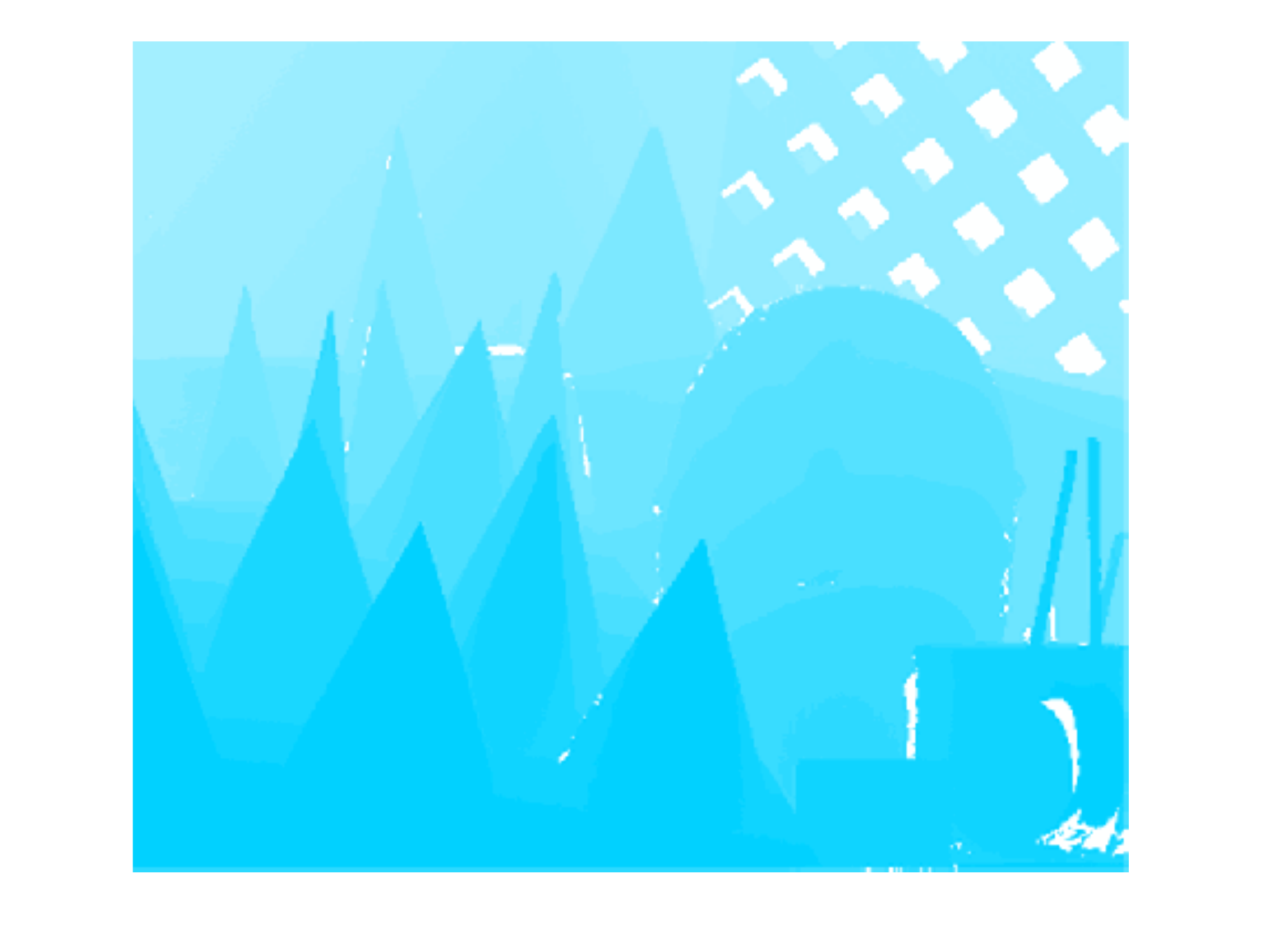} 
	\end{minipage}
	\begin{minipage}[t]{0.25\textwidth}
	\includegraphics[width=1\textwidth]{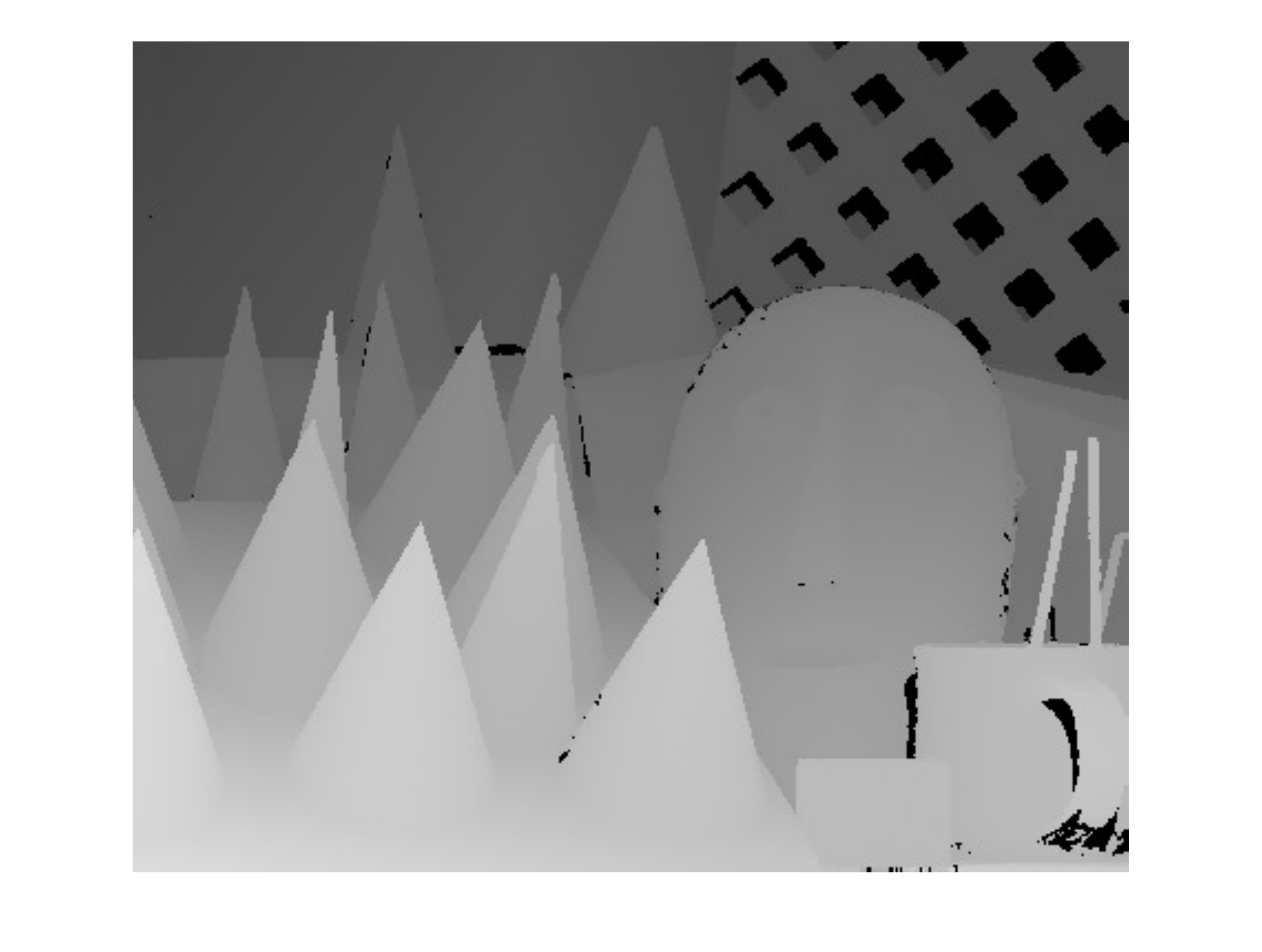}
	\end{minipage}
}
\subfigure[Huguet - rotating sphere]
{
	\begin{minipage}[t]{0.25\textwidth}
	\includegraphics[width=1\textwidth]{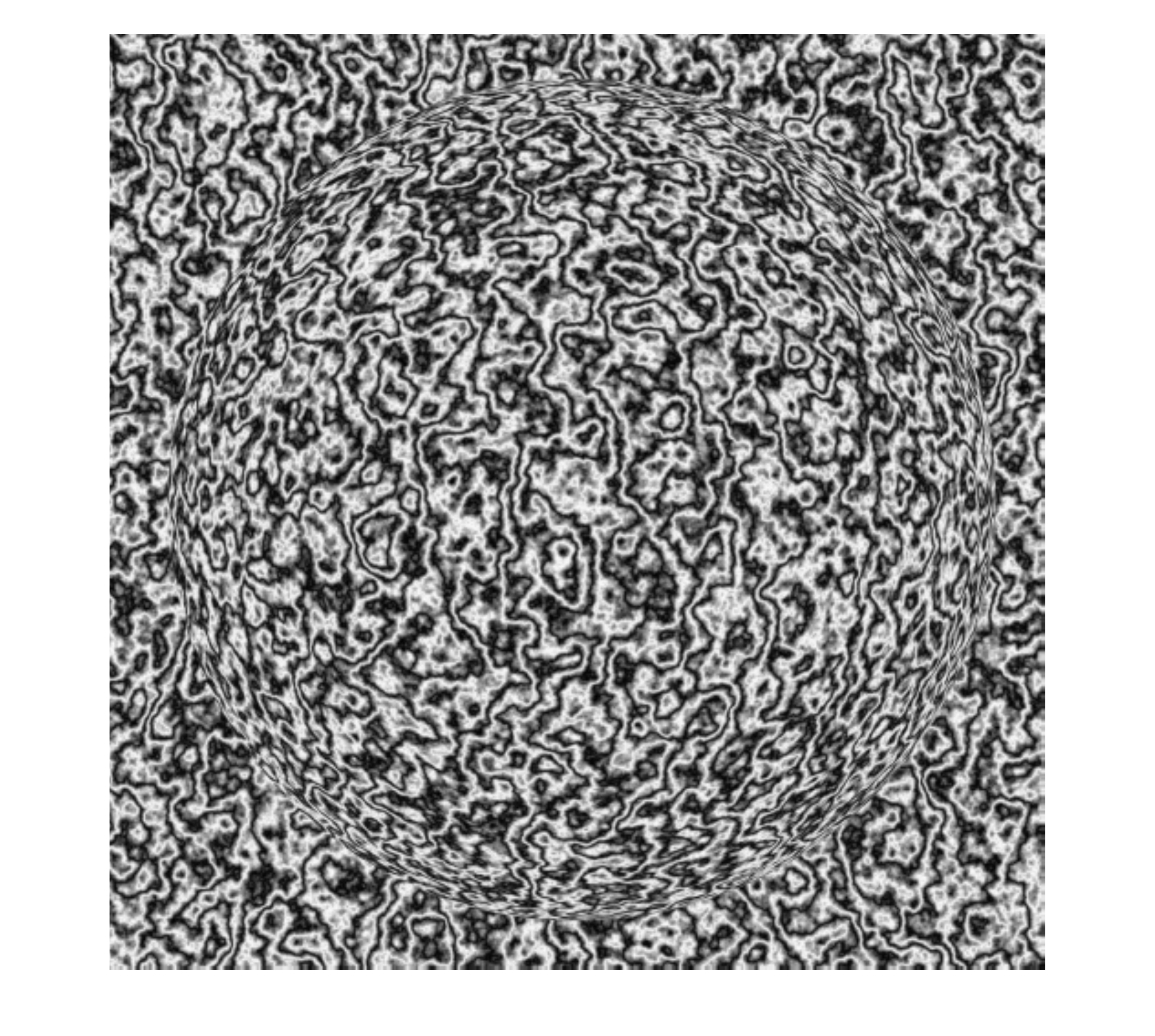} 
	\end{minipage}
	\begin{minipage}[t]{0.25\textwidth}
	\includegraphics[width=1\textwidth]{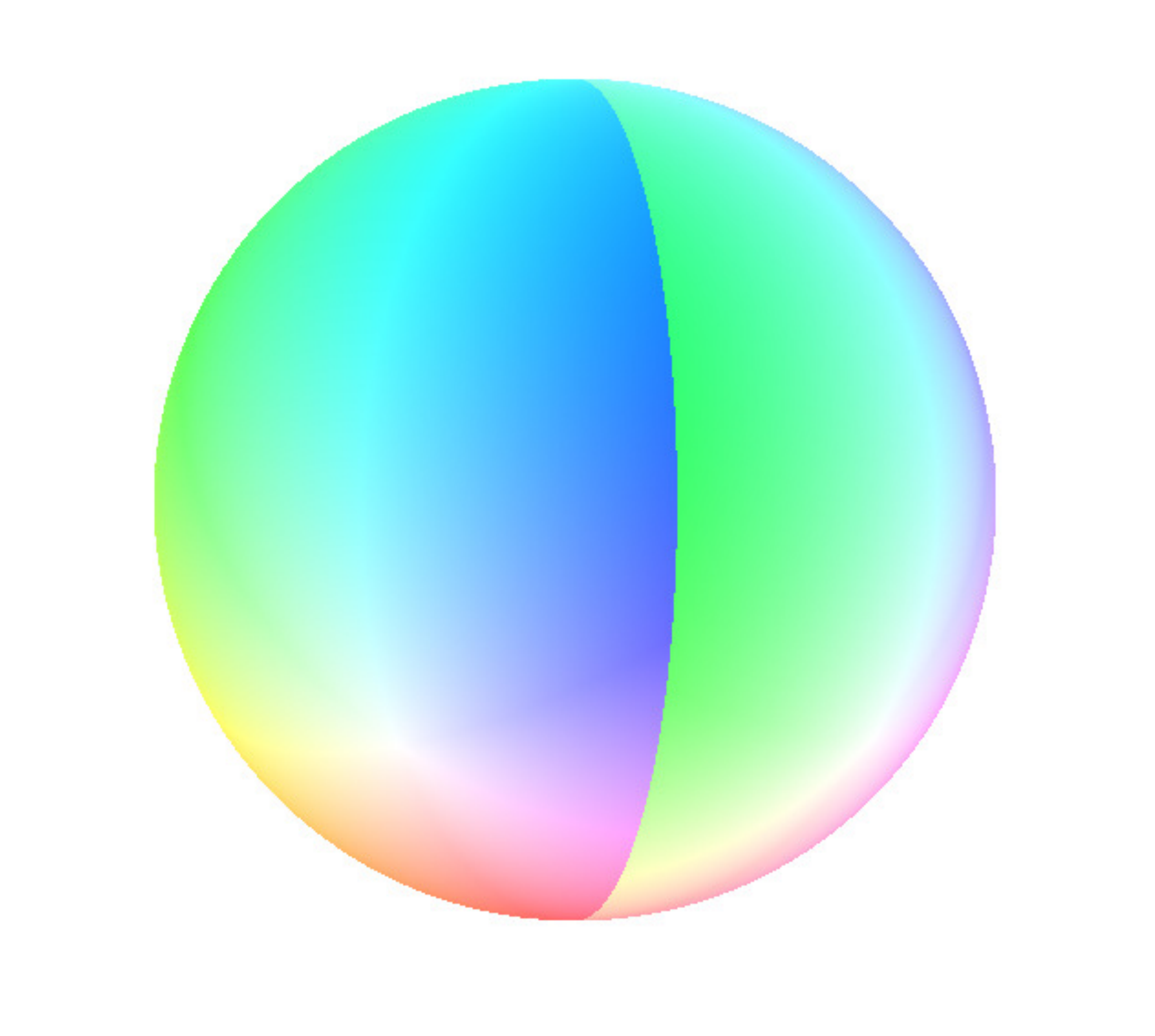} 
	\end{minipage}
	\begin{minipage}[t]{0.25\textwidth}
	\includegraphics[width=1\textwidth]{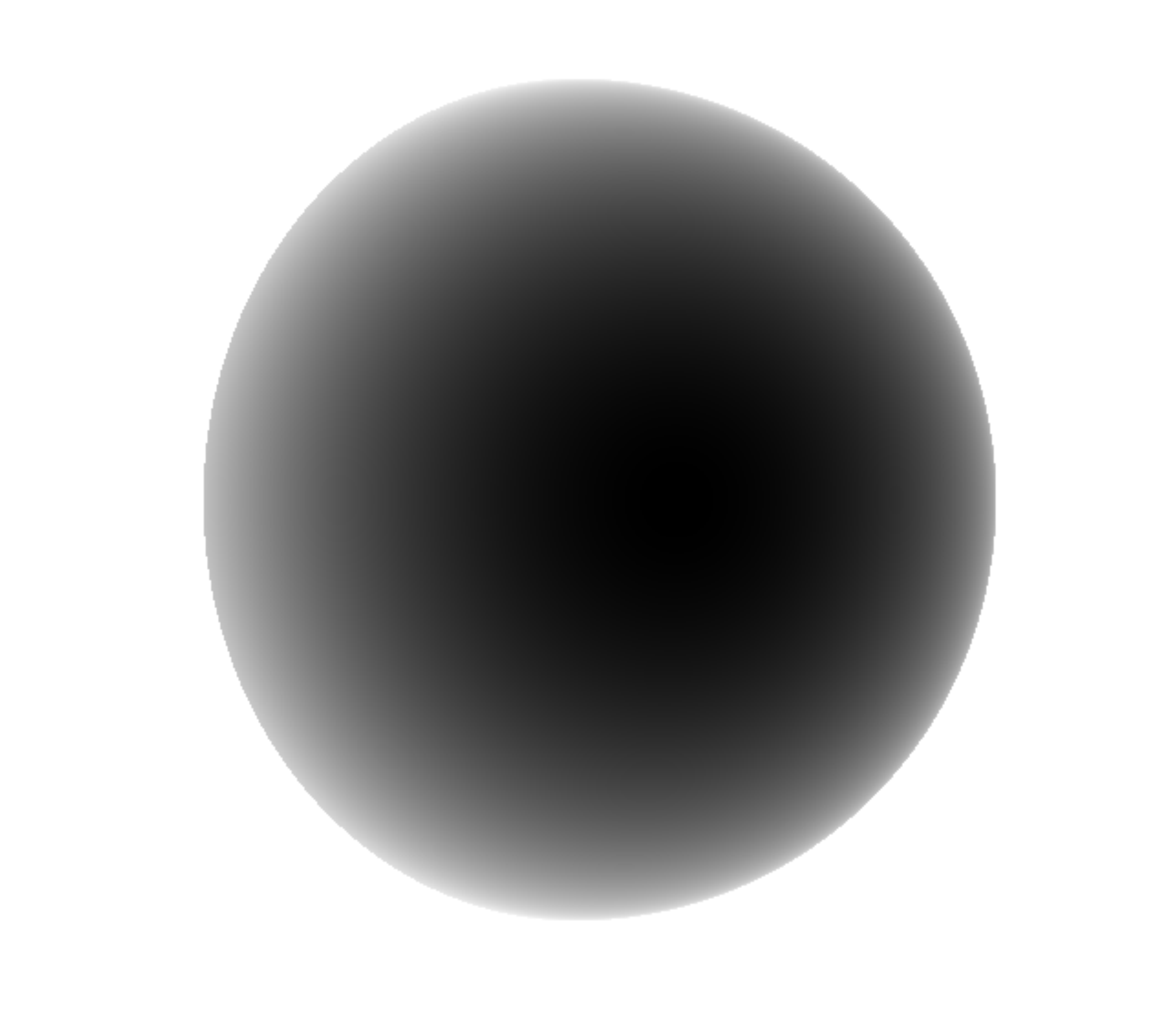}
	\end{minipage}
}
\subfigure[EISATS]
{
	\begin{minipage}[t]{0.25\textwidth}
	\includegraphics[width=1\textwidth]{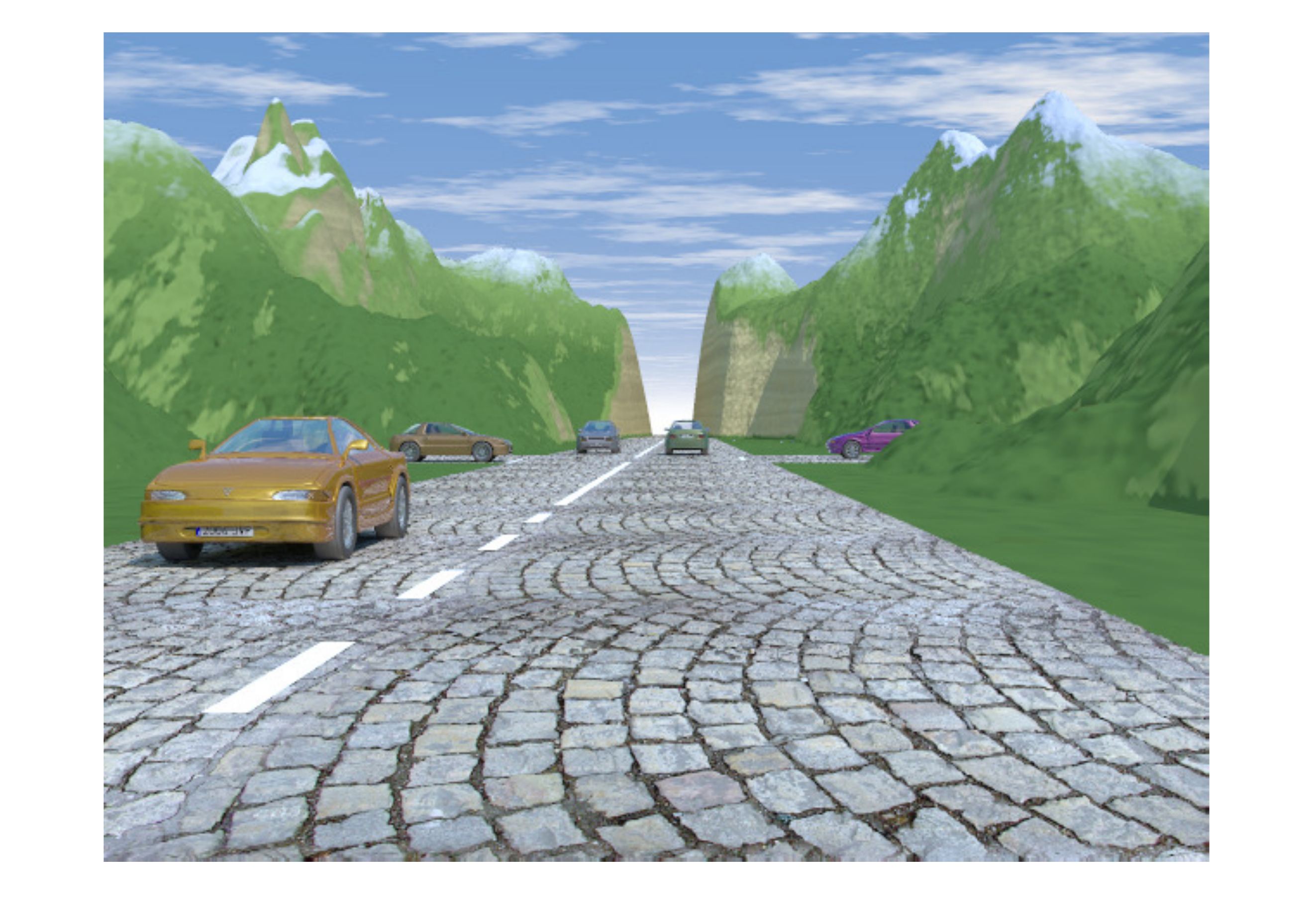} 
	\end{minipage}
	\begin{minipage}[t]{0.25\textwidth}
	\includegraphics[width=1\textwidth]{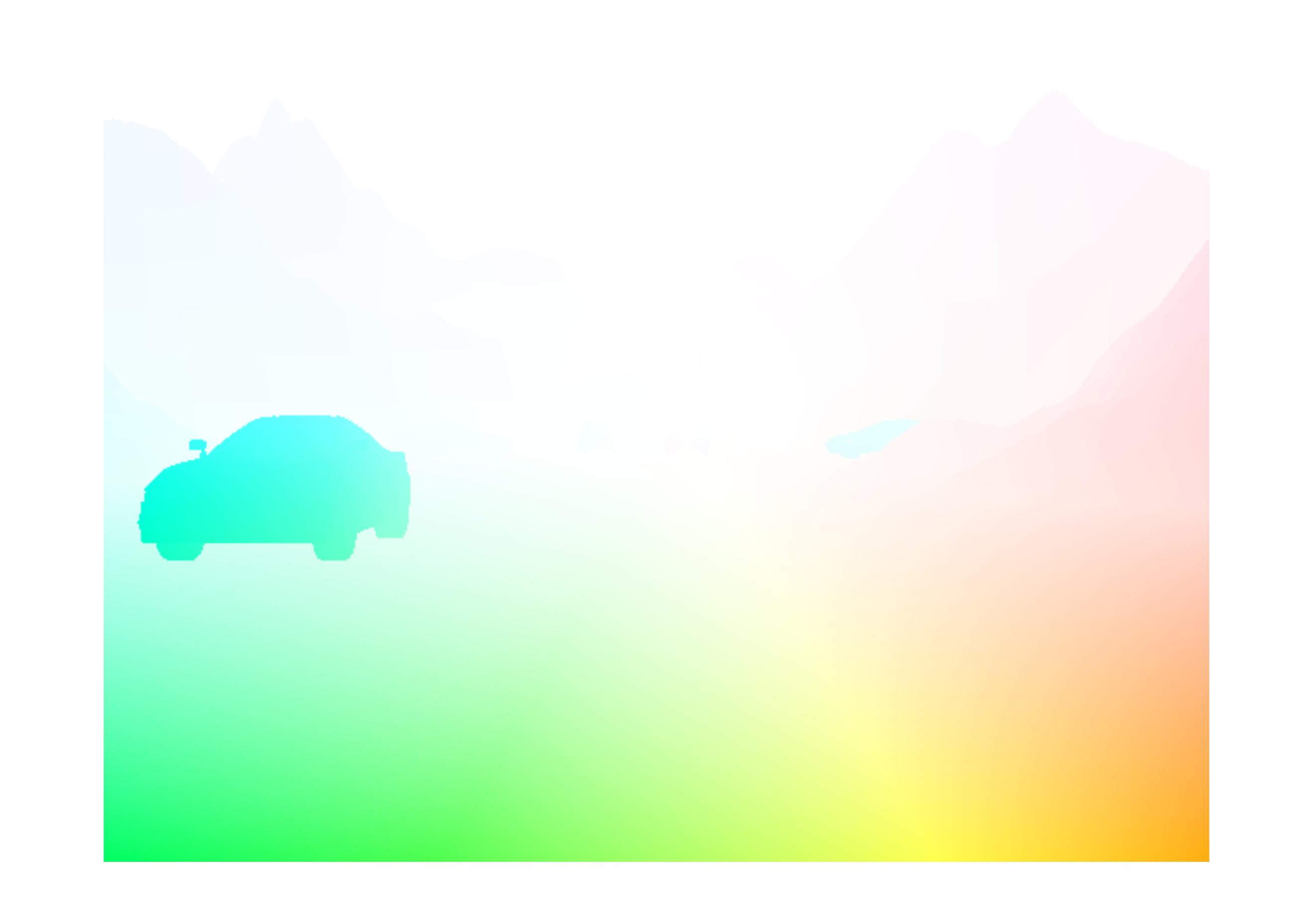} 
	\end{minipage}
	\begin{minipage}[t]{0.25\textwidth}
	\includegraphics[width=1\textwidth]{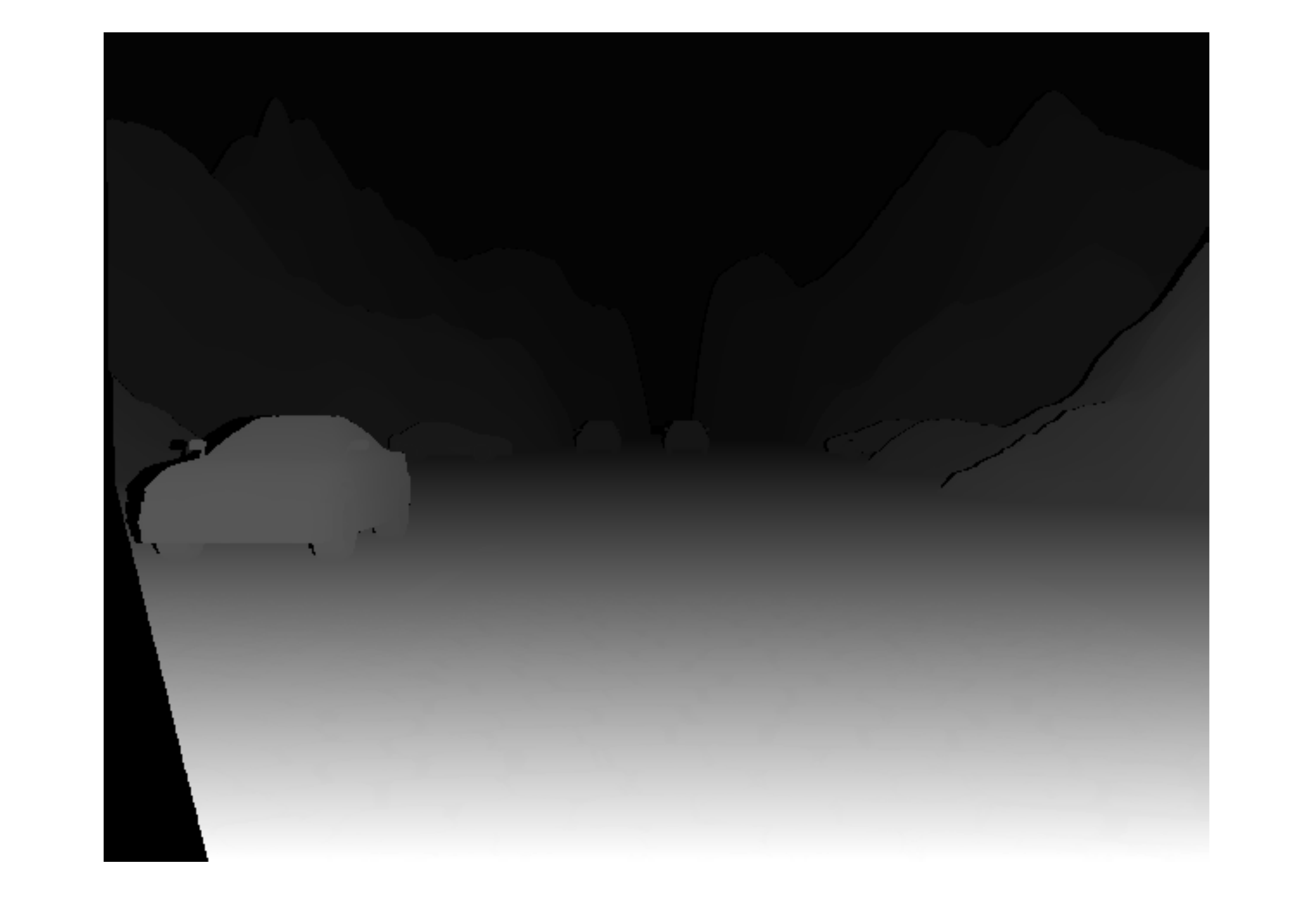}
	\end{minipage}
}
\subfigure[KITTI 2015]
{
	\begin{minipage}[t]{0.3\textwidth}
	\includegraphics[width=1\textwidth]{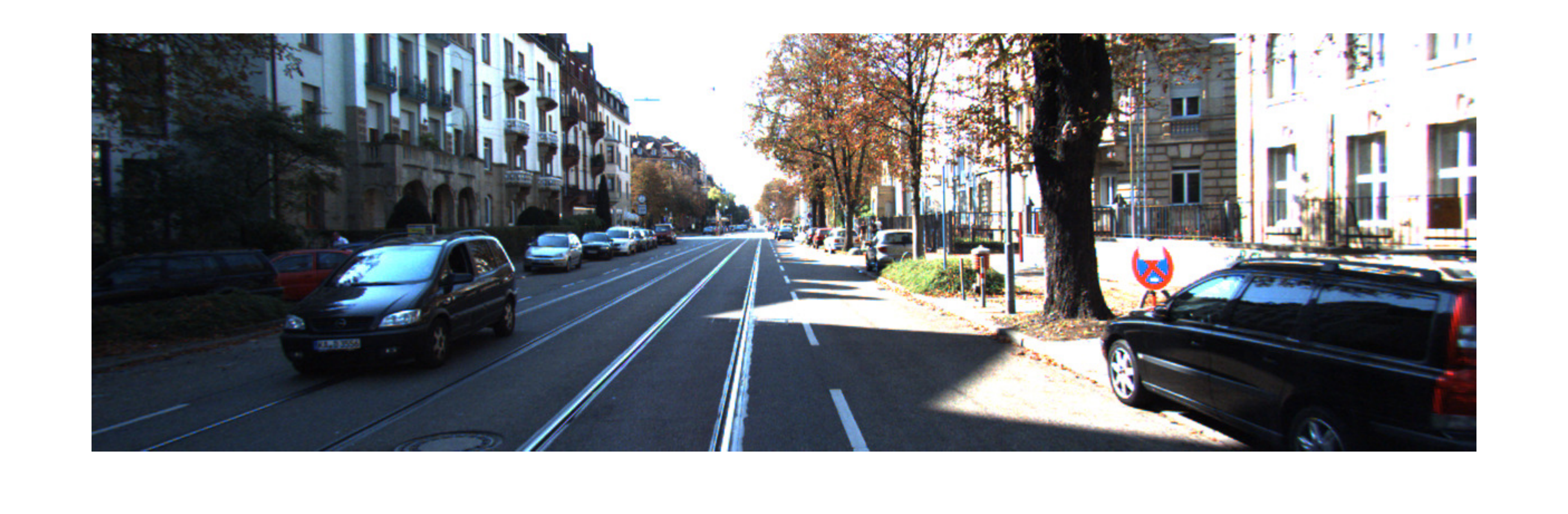} 
	\end{minipage}
	\begin{minipage}[t]{0.3\textwidth}
	\includegraphics[width=1\textwidth]{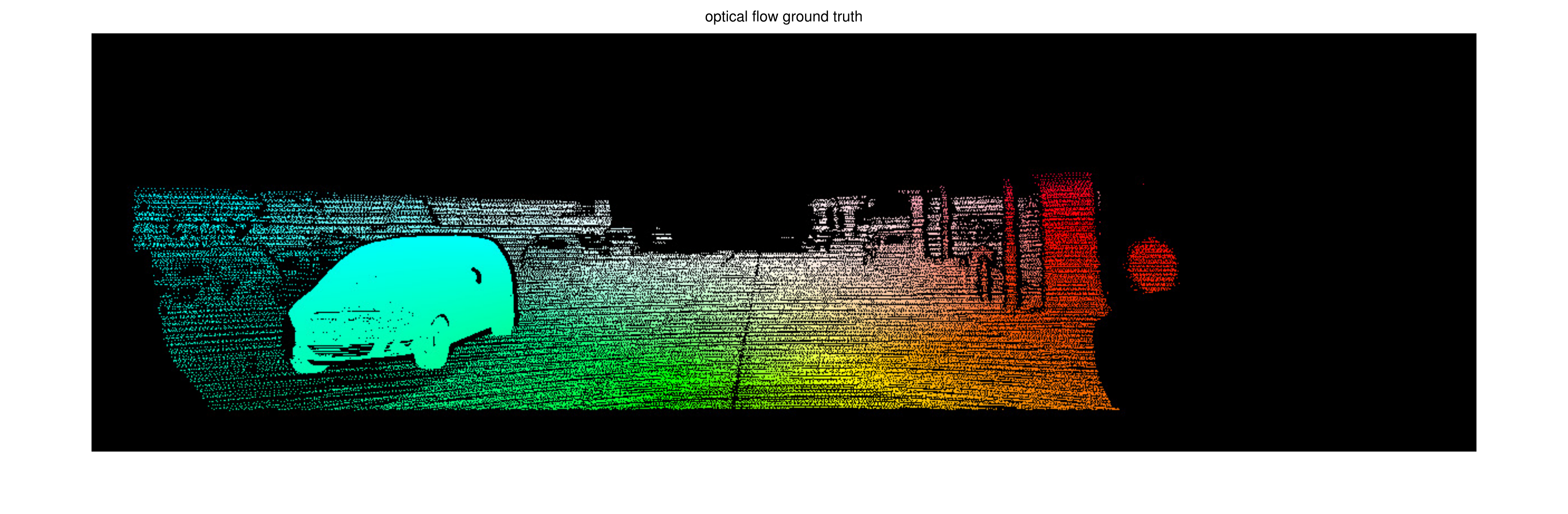} 
	\end{minipage}
	\begin{minipage}[t]{0.3\textwidth}
	\includegraphics[width=1\textwidth]{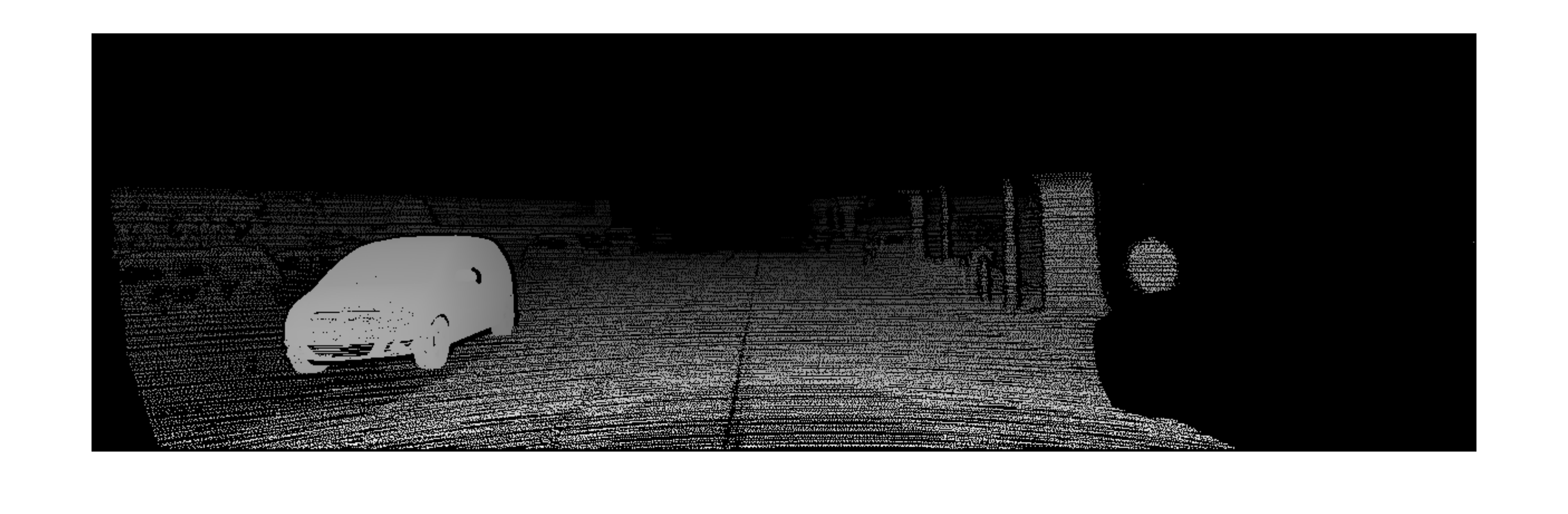}
	\end{minipage}
}
\subfigure[MPI Sintel]
{
	\begin{minipage}[t]{0.3\textwidth}
	\includegraphics[width=1\textwidth]{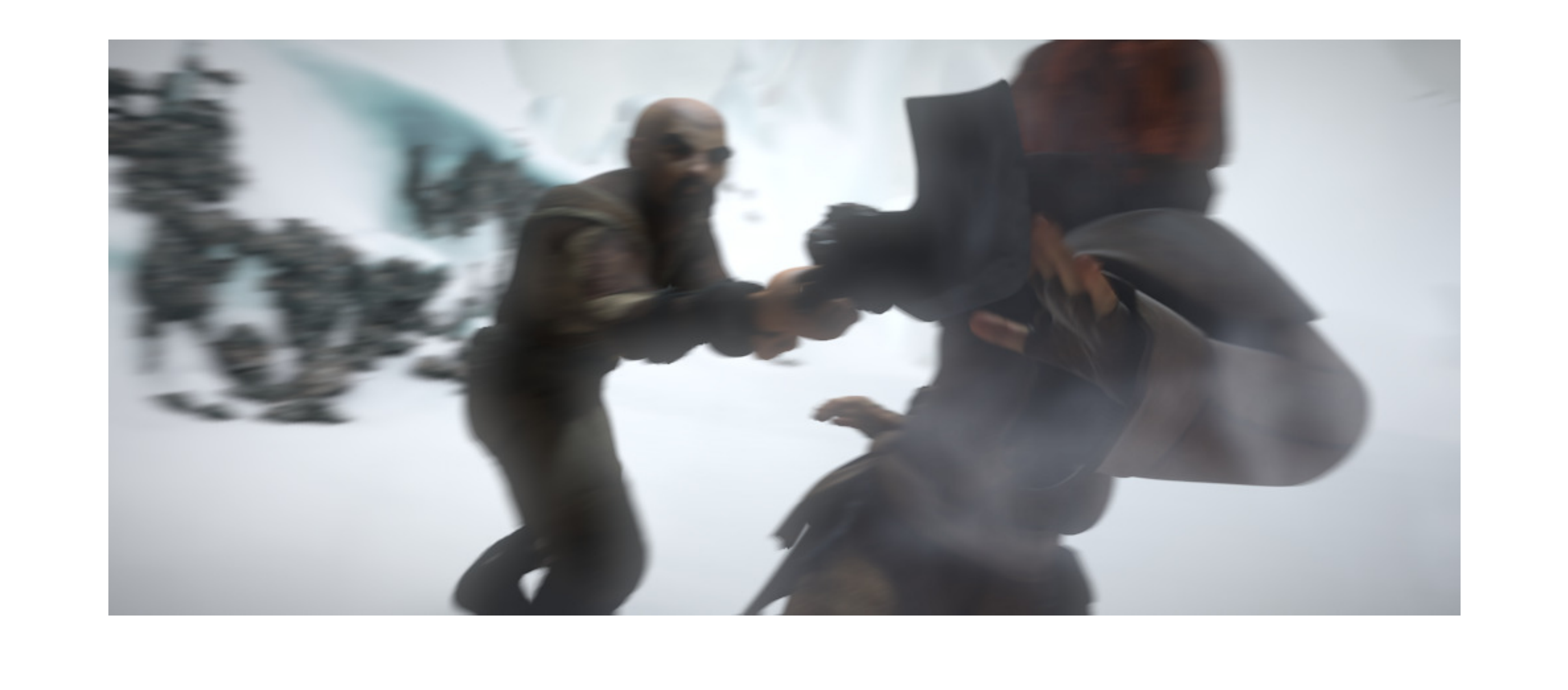} 
	\end{minipage}
	\begin{minipage}[t]{0.3\textwidth}
	\includegraphics[width=1\textwidth]{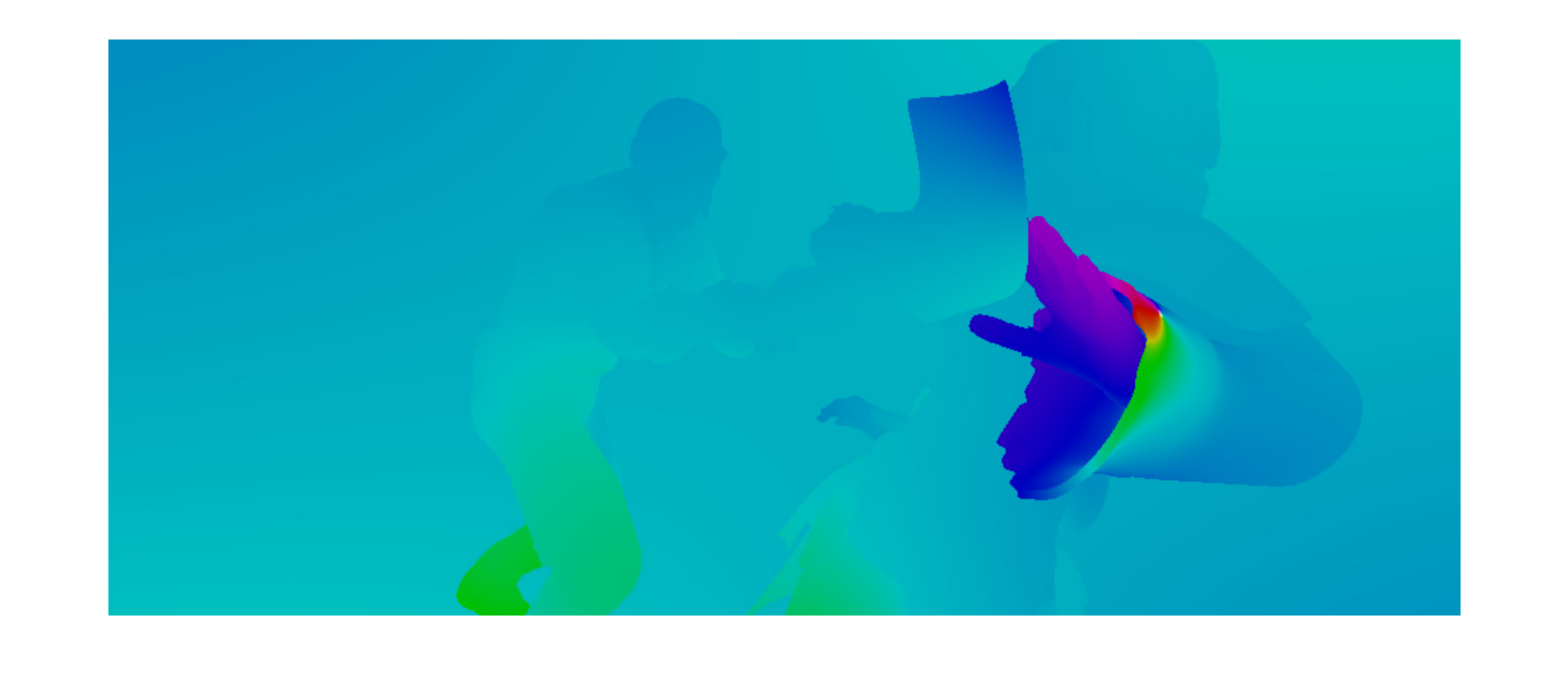} 
	\end{minipage}
	\begin{minipage}[t]{0.3\textwidth}
	\includegraphics[width=1\textwidth]{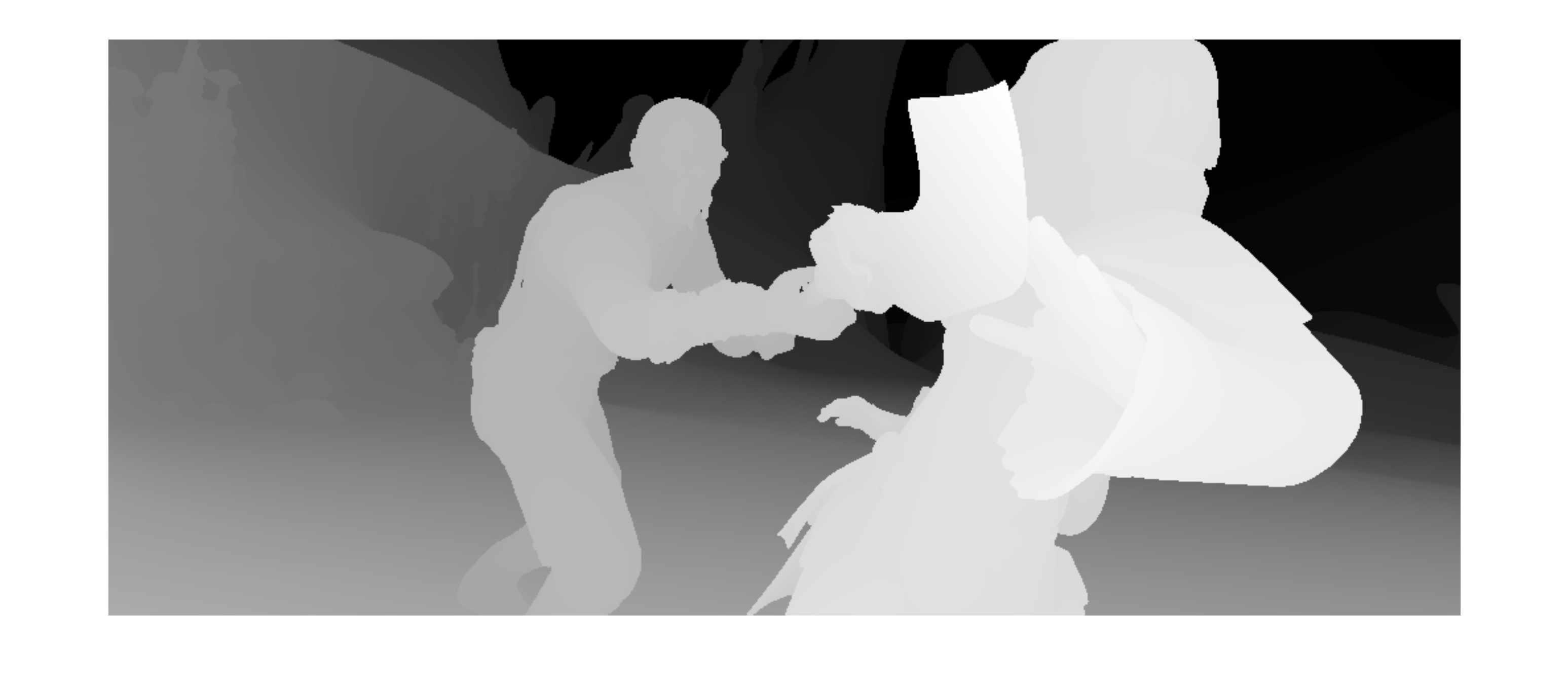}
	\end{minipage}
}
\caption{Sample images with stereo and optical flow ground truth. The first row presents the left-view color image, the second row presents the color-encoded optical flow, and the third row presents the left-view corresponding disparity map.}
\label{fig:dataset}
\end{figure*}

\begin{figure}[t]
\centering
\subfigure[Freiburg - FlyingThings3D]
{
	\begin{minipage}[t]{0.3\textwidth}
	\includegraphics[width=1\textwidth]{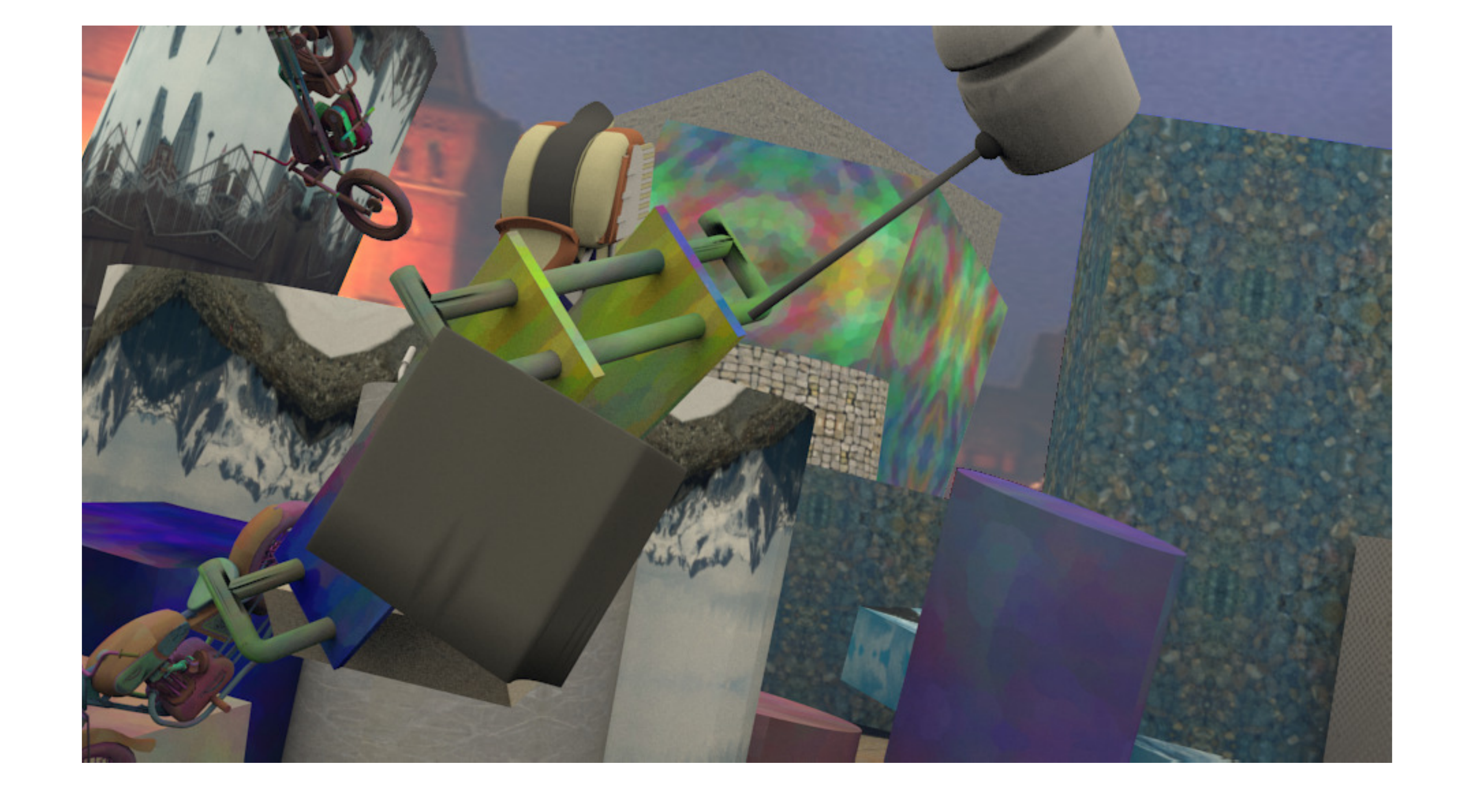} 
	\end{minipage}
	\begin{minipage}[t]{0.3\textwidth}
	\includegraphics[width=1\textwidth]{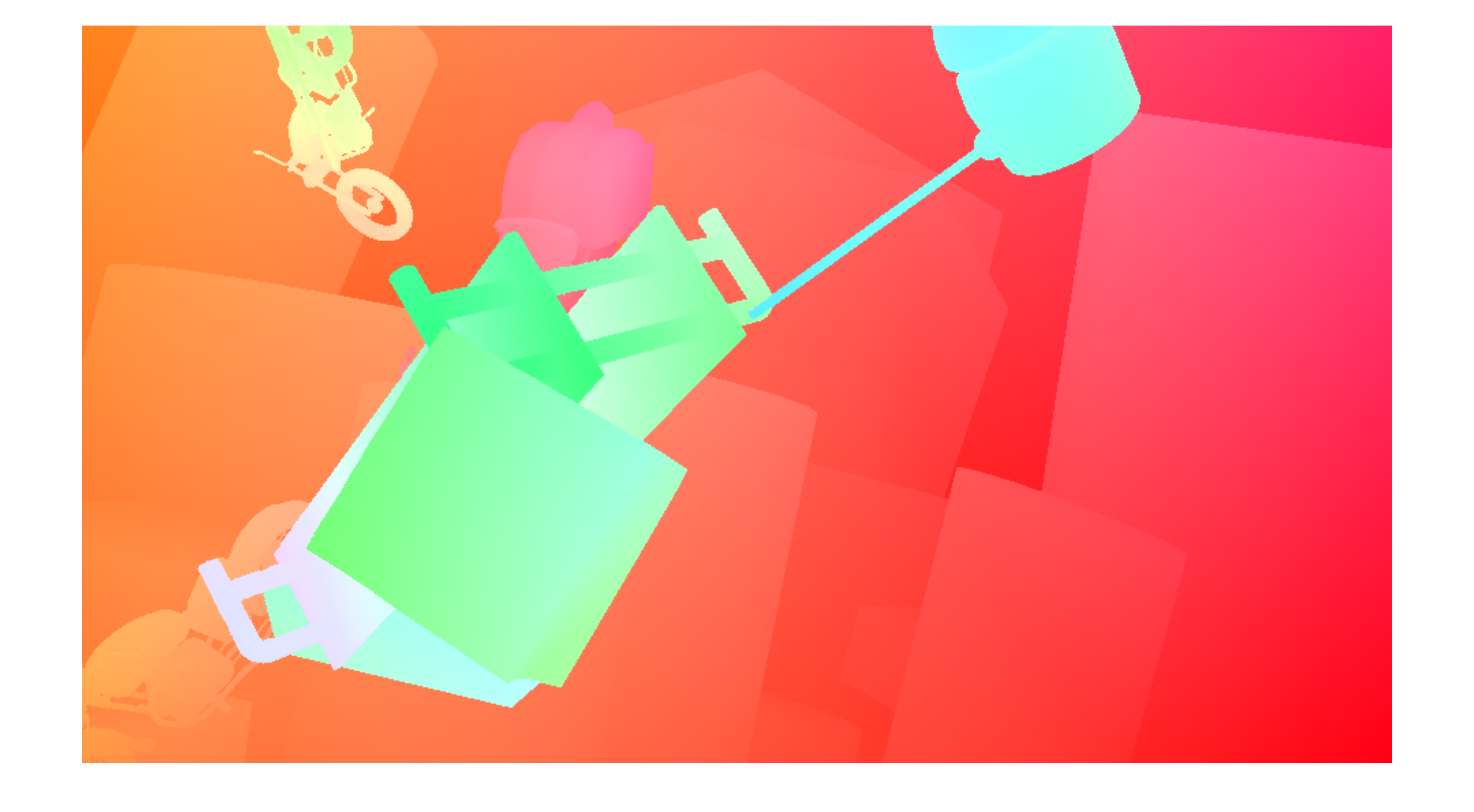} 
	\end{minipage}
	\begin{minipage}[t]{0.3\textwidth}
	\includegraphics[width=1\textwidth]{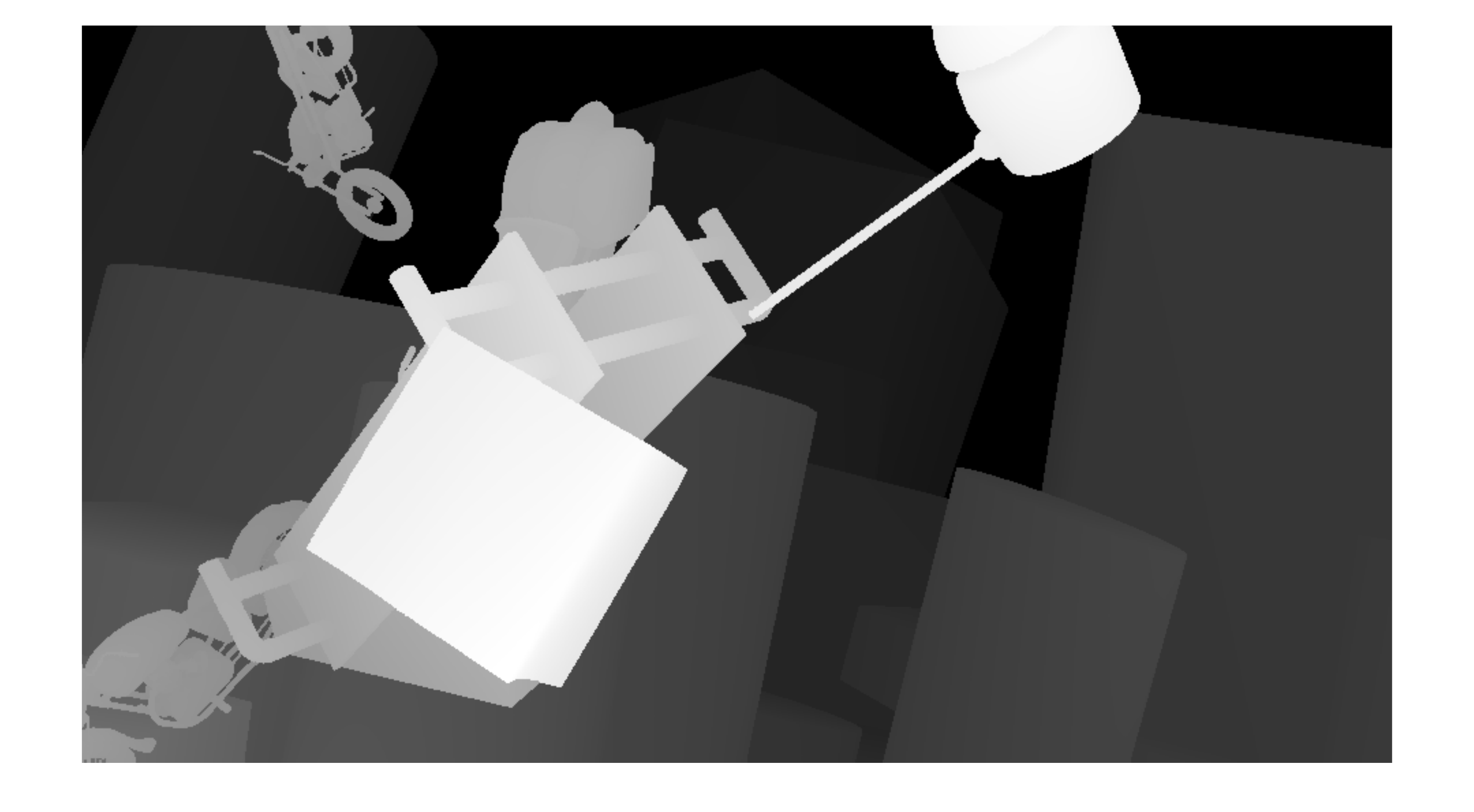}
	\end{minipage}
}
\vspace{-2ex}
\subfigure[Freiburg - Driving]
{
	\begin{minipage}[t]{0.3\textwidth}
	\includegraphics[width=1\textwidth]{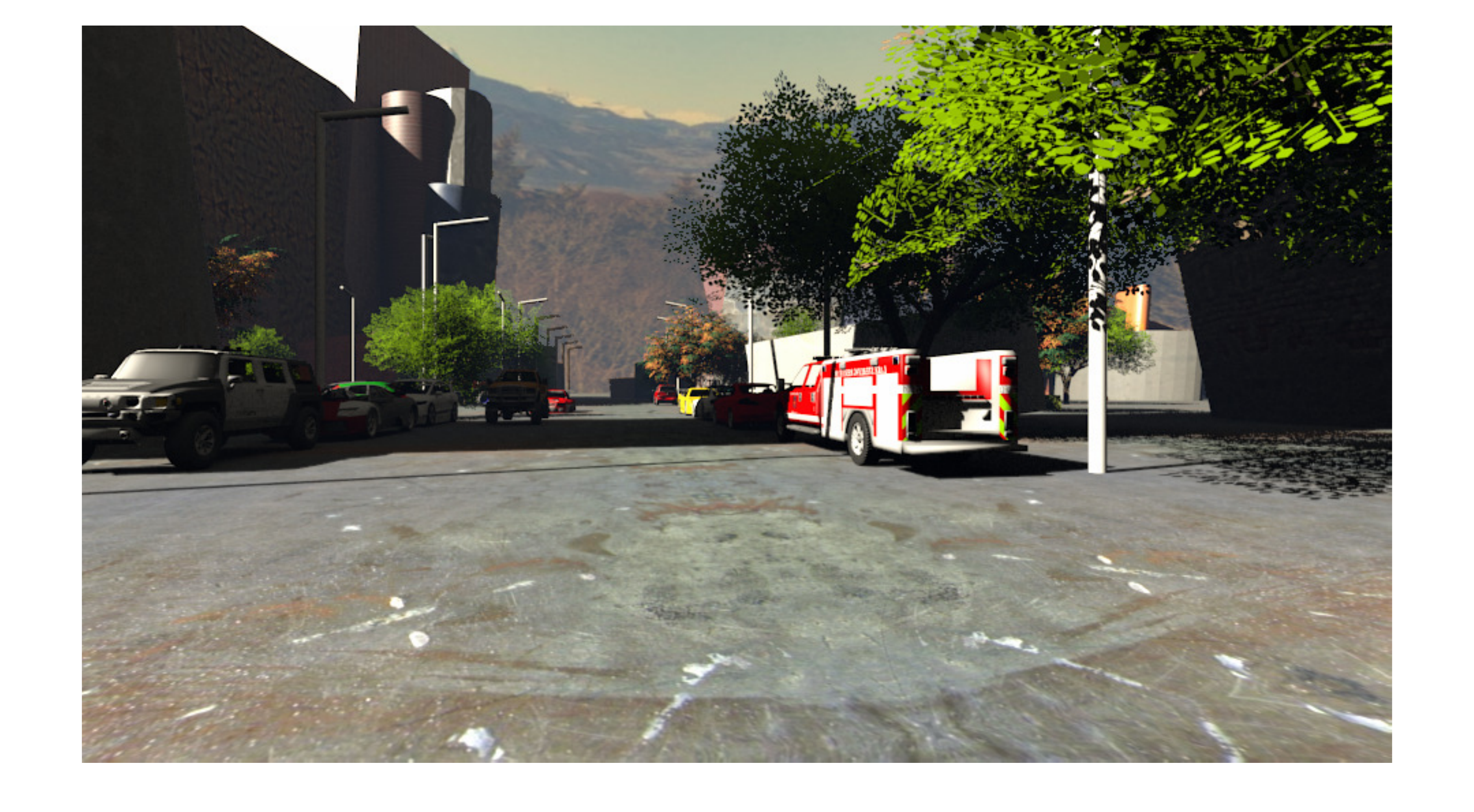} 
	\end{minipage}
	\begin{minipage}[t]{0.3\textwidth}
	\includegraphics[width=1\textwidth]{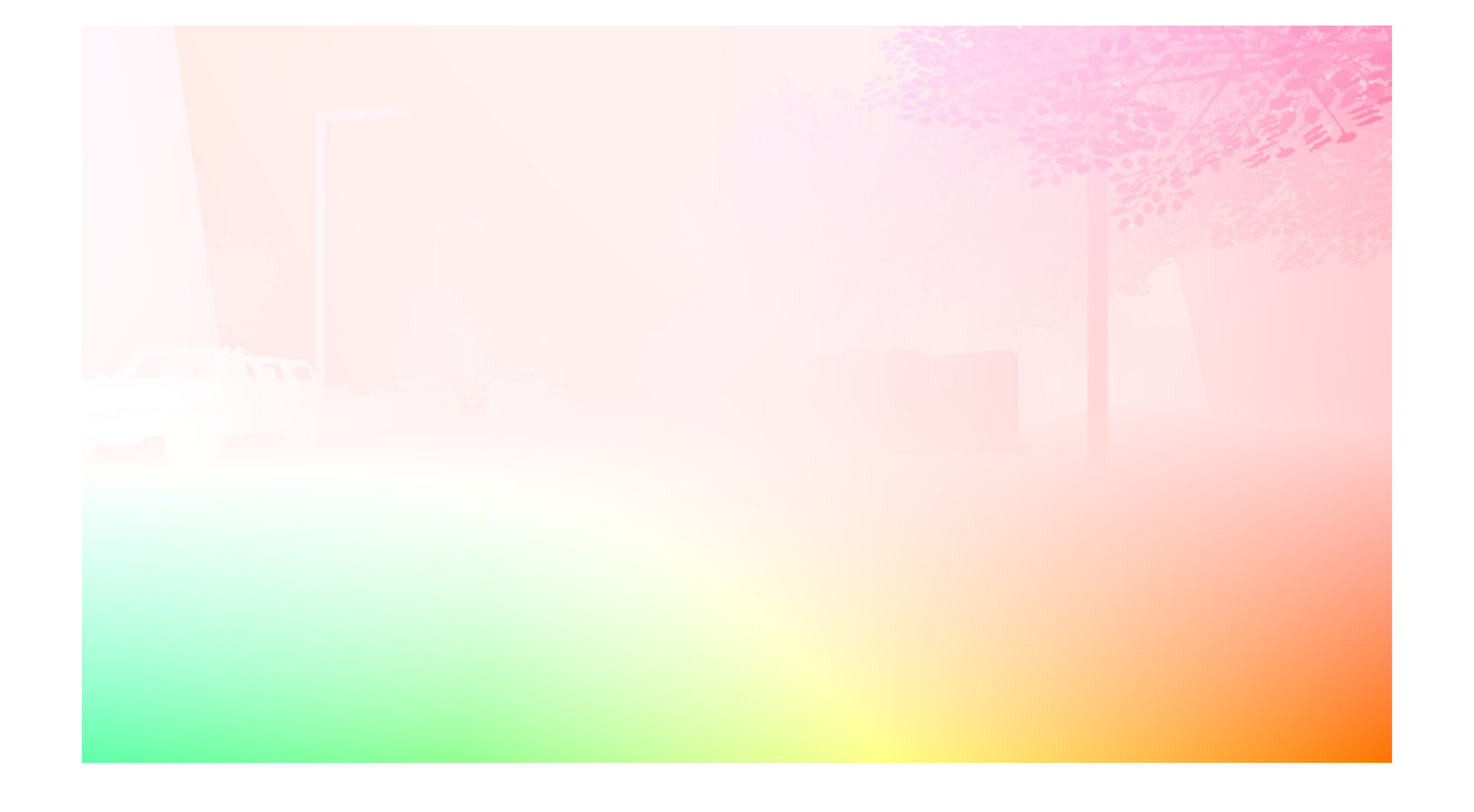} 
	\end{minipage}
	\begin{minipage}[t]{0.3\textwidth}
	\includegraphics[width=1\textwidth]{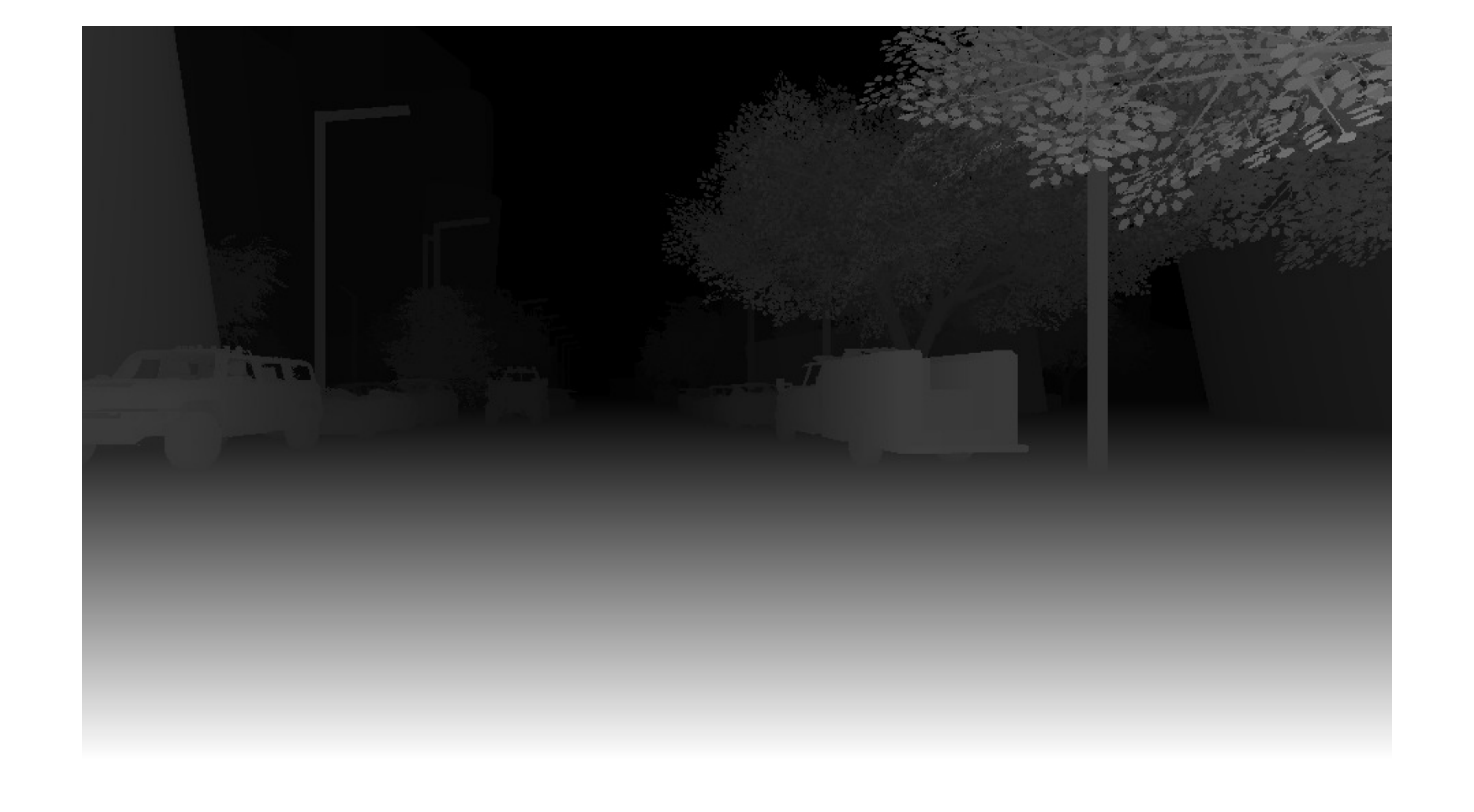}
	\end{minipage}
}
\vspace{-2ex}
\subfigure[Freiburg - Monkaa]
{
	\begin{minipage}[t]{0.3\textwidth}
	\includegraphics[width=1\textwidth]{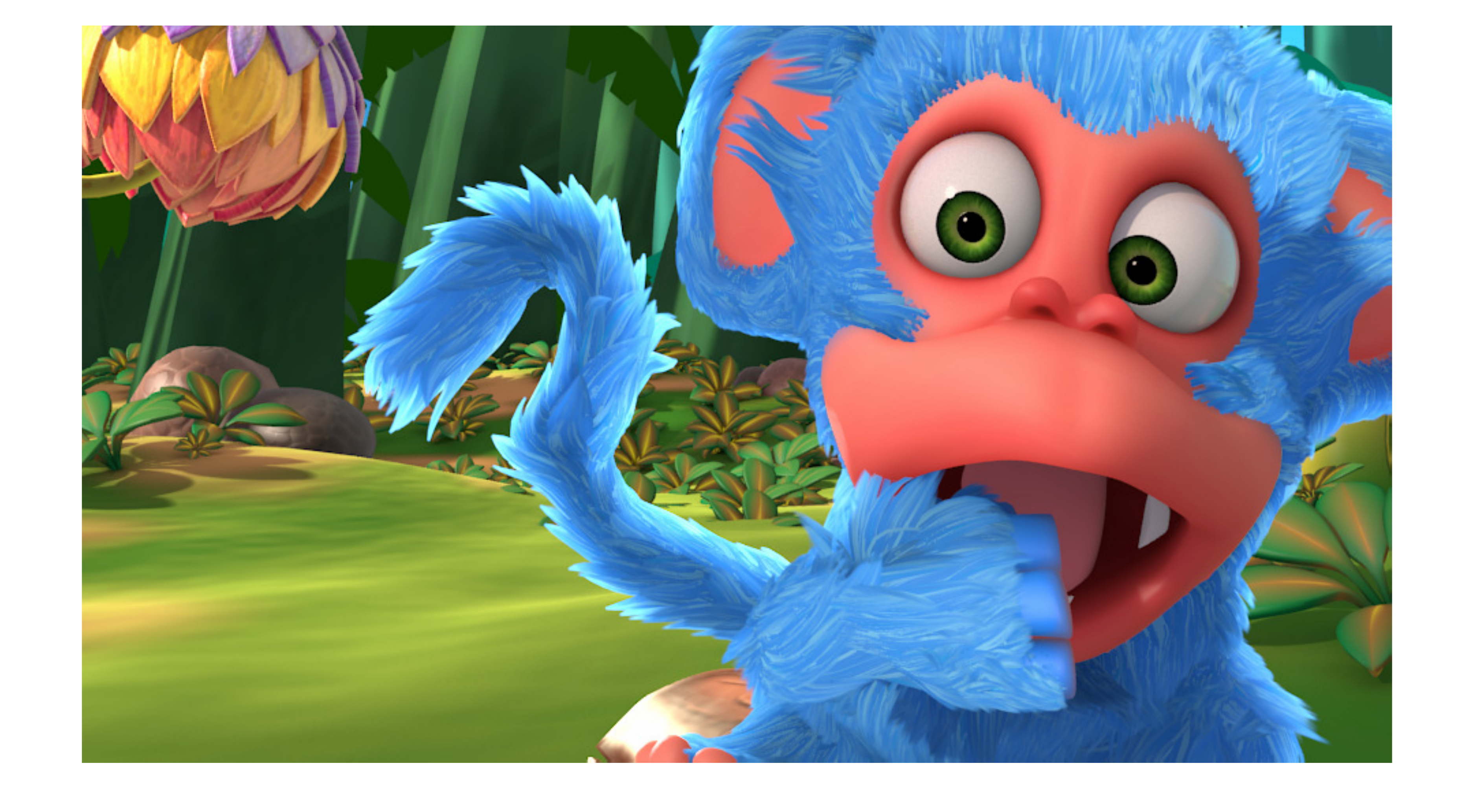} 
	\end{minipage}
	\begin{minipage}[t]{0.3\textwidth}
	\includegraphics[width=1\textwidth]{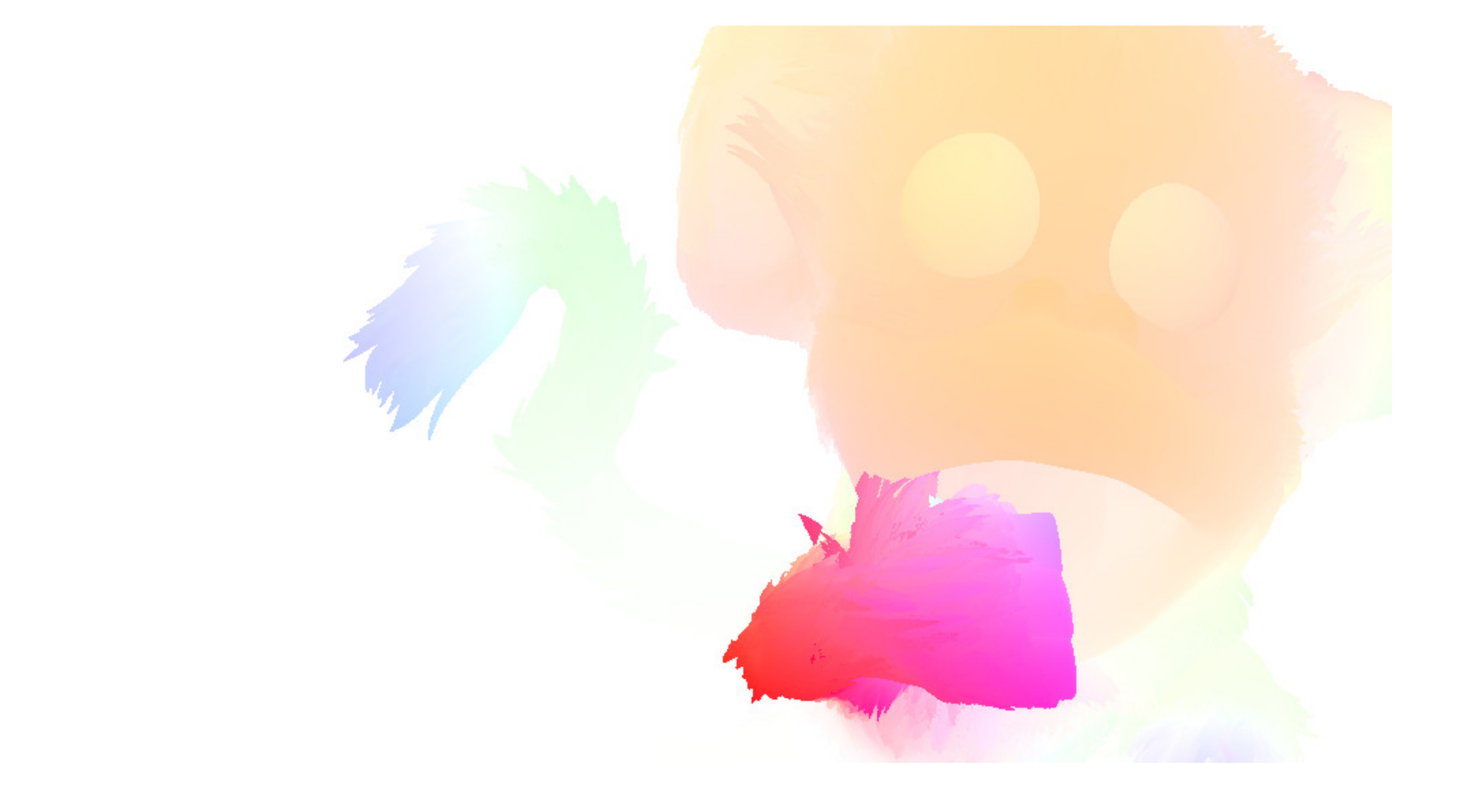} 
	\end{minipage}
	\begin{minipage}[t]{0.3\textwidth}
	\includegraphics[width=1\textwidth]{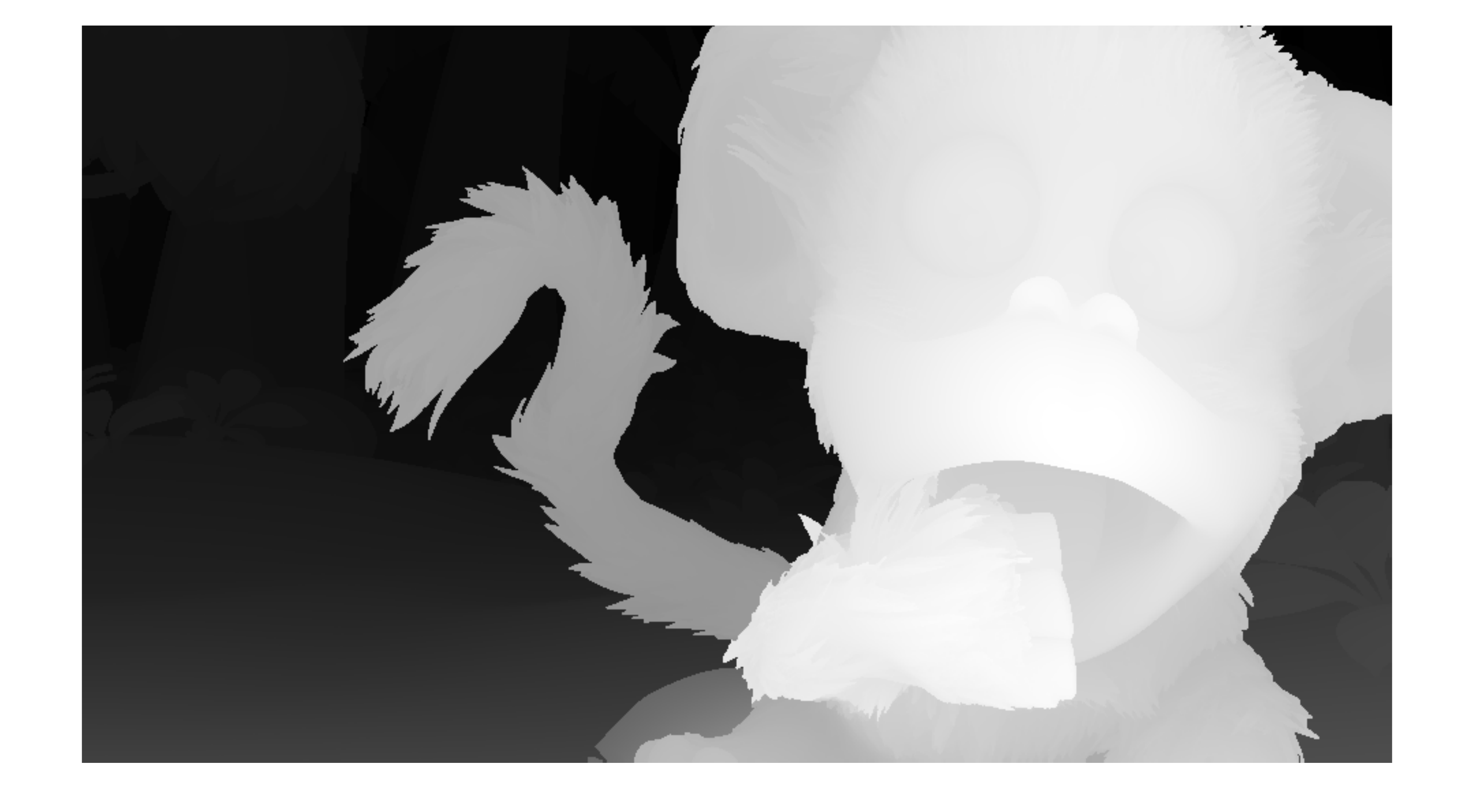}
	\end{minipage}
}
\caption{Sample images of Freiburg dataset. The first row: the left-view color image, the second row: the color-encoded optical flow ground truth, the third row: the left-view corresponding disparity ground truth.}
\label{fig:freiburg}
\end{figure}

In the following section, we briefly introduced each commonly used dataset. A thorough information is illustrated in Table~\ref{tab:dataset}. Sample images including color image, optical flow ground truth and disparity ground truth of each dataset is presented in Figure~\ref{fig:dataset} and \ref{fig:freiburg}.

\begin{table}
\footnotesize
\centering
\begin{tabular}{p{2cm}p{2cm}p{2cm}p{2cm}p{2cm}p{4cm}}
\toprule
\vtop{\hbox{\strut \textbf{Literature}}\hbox{\strut \textbf{year}}} & \textbf{Ground truth} &\textbf{Resolution}& \vtop{\hbox{\strut \textbf{Number[T]}}\hbox{\strut \textbf{(tr, te)}}}& \textbf{Annotations} & \textbf{URL} \\
\toprule
\vtop{\hbox{\strut Middlebury}\hbox{\strut ~\cite{middlebury01}, ~\cite{middlebury02}}\hbox{\strut 2001, 2003}}& \vtop{\hbox{\strut OF, Disp,} \hbox{\strut Occ, Te, DC}} & 450$\times$375 & 3 & \vtop{\hbox{\strut Cones}\hbox{\strut Teddy}\hbox{\strut Venus}} & \sloppy\url{http://vision.middlebury.edu/stereo/data/} \\
\hline
\vtop{\hbox{\strut Rotating sphere}\hbox{\strut ~\cite{19}, ~\cite{28}}\hbox{\strut 2007, 2010}}& \vtop{\hbox{\strut OF, Disp}\hbox{\strut} \hbox{\strut SF, Occ}} & 512$\times$512 & \vtop{\hbox{\strut 1}\hbox{\strut}\hbox{\strut 1}} & \vtop{\hbox{\strut Huguet}\hbox{\strut}\hbox{\strut Basha}} & \url{http://devernay.free.fr/vision/varsceneflow/}  \url{http://people.csail.mit.edu/talidekel/MVSF.html}\\
\hline
\vtop{\hbox{\strut EISATS}\hbox{\strut ~\cite{Povray}}\hbox{\strut 2008}}& \vtop{\hbox{\strut OF, Disp, DispC}} & 640$\times$480 & 498 & Set 2 & \url{http://ccv.wordpress.fos.auckland.ac.nz/eisats/}\\
\hline
\vtop{\hbox{\strut KITTI}\hbox{\strut ~\cite{KITTI2012}}\hbox{\strut 2012}}& \vtop{\hbox{\strut OF, Disp,} \hbox{\strut Occ, Lb}} & 1238$\times$374 & \vtop{\hbox{\strut 384}\hbox{\strut (194,195)}} & - & \url{http://www.cvlibs.net/datasets/kitti/eval_stereo_flow.php?benchmark=stereo} \\
\hline
\vtop{\hbox{\strut KITTI}\hbox{\strut ~\cite{67}}\hbox{\strut 2015}}&  OF, Disp, Occ & 1242$\times$375 & \vtop{\hbox{\strut 400}\hbox{\strut (200,200)}} & - & \url{http://www.cvlibs.net/datasets/kitti/eval_scene_flow.php}\\
\hline
\vtop{\hbox{\strut MPI Sintel}\hbox{\strut ~\cite{Sintel}}\hbox{\strut 2012}}& \vtop{\hbox{\strut OF, Dis, Dep} \hbox{\strut Occ, Seg, CM}} & 1024$\times$436 & \vtop{\hbox{\strut 1828}\hbox{\strut (1264,564)}} & - & \url{http://sintel.is.tue.mpg.de/downloads}\\
\hline
\vtop{\hbox{\strut Freiburg}\hbox{\strut ~\cite{Mayer}}\hbox{\strut 2016}}& \vtop{\hbox{\strut OF, Disp, DispC} \hbox{\strut Occ, Seg, MB}} & 960$\times$540 & \vtop{\hbox{\strut 39049}\hbox{\strut (34801,4248)}}& \vtop{\hbox{\strut FlyingThings3D}\hbox{\strut Driving}\hbox{\strut Monkaa}} & \url{http://lmb.informatik.uni-freiburg.de/resources/datasets/SceneFlowDatasets.en.html}\\
\bottomrule
\end{tabular}
\caption{A summary of the popular datasets. Notation for headers: \textbf{[T]}: number of scenes in total, \textbf{tr}: numbers of training images, and \textbf{te}: numbers of testing images. Notation for data format: OF: optical flow ground truth, Disp: disparity ground truth, DispC: disparity change ground truth, Dep: depth map ground truth, SF: point cloud scene flow ground truth, CM: camera motion, Seg: segmentation, Lb: semantic and instance labels and car labels,  Occ: occlusion region ground truth, MB: motion boundaries, Te: textureless region ground truth, DC: discontinuity region ground truth.}
\label{tab:dataset}
\end{table}

\subsubsection{Middlebury dataset}
%Middlebury
The Middlebury stereo dataset~\cite{middlebury01,middlebury02} is commonly used as a quantitatively evaluation benchmark for optical flow and stereo matching. Particularly, the subsets named Teddy, Cones and Venus provided both optical flow and disparity ground truth, and hence it is utilized for scene flow evaluation~\cite{21,19,36,28,46,50,52,53,54,56,58,60,63,68}. Each subset simulates a simple translational motion along the X axis. The 8 cameras are rectified and placed parallelly and equally along the X axis and thus the scene motion projected onto the image plane is the disparity between two cameras. Under a binocular setting, images from camera 2 and 6 are taken as stereo pairs from time \emph{t}, while images from camera 4 and 8 are taken as stereo pairs from time \emph{t+1}. The disparity ground truth is the disparity from camera 2 to 6, and the optical flow ground truth is the disparity from camera 2 to 4. Similarly, when it comes to RGB-D scene flow, the ground truth disparity map is converted into depth channel for evaluation.

\subsubsection{Rotating sphere}
%sphere
In 2007, Huguet utilized the Pov-Ray to render a publicly available rectified synthetic rotating sphere with optical flow and stereo ground truth~\cite{19}, which is commonly used as a benchmark~\cite{22,33,37,39,40,51,67}. Two hemispheres rotate in opposite directions which lead to strong discontinuity. Basha modified the rotating sphere by adding a rotating plane behind the rotating sphere with OpenGL~\cite{28}. It was used for evaluating three-dimensional parametrization scene flow on account of the ground truth it provided~\cite{44,46}. Moreover, it provided five rectified views which can also be evaluating multi-view scene flow, without full-view geometry ground truth though.

\subsubsection{EISATS dataset}
%povray
EISATS traffic scene datasetsdatasets is a synthetic stereo image sequence rendered by Pov-Ray with ground truth for both stereo and motion~\cite{Povray}. The sequence 1 consists of 100 frames while sequence 2 consists of 396 frames with ego-motion. The synthetic traffic scene consists of a few moving cars under an open environment. Only a few papers utilized this stereo pairs sequence for binocular scene flow estimation~\cite{22,31,40,49}.

\subsubsection{KITTI dataset}
%KITTI
Geiger took advantage of their calibrated autonomous driving platform and developed the novel challenging KITTI benchmark with 194 training scenes and 195 test scenes in 2012~\cite{KITTI2012}, and Menze annotated the dynamic scenes with 3D CAD models for all vehicles in motion and obtained a modified dataset with 200 training scenes and 200 test scenes in 2015~\cite{67}. A novel evaluation methodology is also introduced as the KITTI metric, which is illustrated in Section~\ref{subsubsec:specific}. These two binocular-based dataset have been utilized by multiple papers over the years~\cite{19,37,51,58,61,62,67,69}. The scene is much more realistic and challenging compared to the early Middlebury dataset, which is designed specifically for autonomous driving. However, due to the acquisition manner of data, there are missing value in both optical flow and disparity ground truth, as is illustrated in Figure~\ref{fig:dataset}. The density of ground truth is about 75\% to 90\%. Thus, KITTI dataset is not recommended for RGB-D scene flow evaluation on account that it needs dense disparity ground truth for simulating depth data.

\subsubsection{MPI Sintel dataset}
%sintel
MPI Sintel dataset is the largest dataset before 2015~\cite{Sintel}, which consists of 23 training sequences with 1064 frames and 12 test sequences with 564 frames in total. It derived from an open source animated film and the resolution is 1024$\times$436. The scenes are designed to be strictly realistic with fog and motion blur added. Moreover, beta version depth data were then added which can be a perfect dataset for RGB-D scene flow evaluation. Zanfir and Jaimez~\cite{66,70} utilized this dataset for scene flow evaluation in 2015 and gave quantitative analysis. It is highly recommended for its naturalistic setting and density, as well as its comprehensive evaluation protocol. Moreover, the video sequence ensures a multi-frame implement. With the development datasetof scene flow estimation, this dataset which consists of non-rigid motion and large displacement under a high resolution is reliable and challenging enough for evaluation.

\subsubsection{Freiburg dataset}
\label{subsec:Freiburg}
%freiburg
Freiburg dataset is the up-to-date and largest dataset for optical flow, stereo and scene flow evaluation~\cite{Mayer}. The disparity change ground truth make up for the blank of scene flow ground truth. It contains 34801 stereo training frames and 4248 test frames in 960$\times$540 resolution, which intended for deep learning. The dataset consists of three subsets named FlyingThings3D, Monkaa and Driving, which are all made by the 3D suite Blender~\cite{blender}. 

FlyingThins3D contains 21818 training frames with 2247 scenes and 4248 test frames. The amount of data is its first priority. This scene contains abundant texture features and complex occlusion between multiple objects, which is highly challenging.

Monkaa resembles the MPI Sintel dataset since it derived from an animated short film Monkaa as well. It is less naturalistic as Sintel, but the data scale is three times larger. Besides, the articulated motion and non-rigid motion with furry effect are the major issues to tackle.

The Driving scene resembles the KITTI dataset since the dynamic street scenes are captured from the view point of a driving car. The long frame range and fully dense ground truth ensure the reliable evaluation for scene flow under the particular task for autonomous driving.

Compared to other datasets, this dataset contains the largest and most comprehensive scenes and aims for scene flow evaluation particularly compared to other datasets. Besides the 2.5 tera-gigabytes of data in total, a sample pack is also available for less than 100 megabytes. Hence, this dataset is highly recommended for its comprehensiveness. 

\subsubsection{Other datasets}
%others
Many other datasets with ground truth are introduced as well. However, on account of limited reference, the comparisons under these datasets are deficient, and most of them lack public availability and innovation. We list them as follows just in case.

Similar to the rotating sphere dataset created by Huguet, in the early stage, Zhang introduced a synthetic deformable sphere using OpenInventor for quantitative analysis~\cite{03}. Spies modeled a structured light sensor to provide a synthetic textured sphere as well~\cite{08}. Valgaerts generated a general rotating sphere scene without rectification~\cite{33}, while Cech added a fast moving bar and a slanted background plane and then textured the whole scene with white noise using Blender~\cite{37}, which made it much more challenging for the scene flow estimation. 

Moreover, Ferstl created a translating and rotating cube in front of the static plane with white noise textured~\cite{53,54}, and Ghuffar created a noisy scene with two cubes moving on a plane in front of a static wall to testify the algorithm's robustness towards noise and occlusion~\cite{55}. Vogel generated nine synthetic box dataset with ground truth~\cite{39}, which consisted of pure rotation, translations in all axes and translation only in depth for independent analysis.

In addition, early in 2005, Luckins created a sloped plane and a sinusoidal plaid pattern named "splaid" for a rough evaluation~\cite{13}, the sampling is the size of 100$\times$100 with depth and RGB color. Gong generated a synthetic 3D scene that consists of a rotating earth model textured with Phong illumination and bump mapping and a translating galaxy background~\cite{24}. The scene is rendered with Gaussian noise and the camera is moving against the earth which makes it really tough for scene flow estimation. Ruttle utilized the Human Eva II dataset for motion tracking and pose estimation and gave the quantitative analysis~\cite{26}. Popham~\cite{30,59} evaluated his algorithm with a multi-view motion capture dataset named "Katy" and "Skirt"~\cite{Katy} with sparse motion trajectory ground truth. Sizintsev captured few sets of sequences with BumbleBee stereo camera and got ground truth with a structure light approach~\cite{middlebury02}, and then used this dataset for evaluation~\cite{45}. Alhaija captured seven pairs of images through Kinect to specifically evaluate scene flow under large displacement~\cite{71}. The matching ground truth was given by manually labeling each segment.

\subsubsection{Performance}
The performance of the methods that first utilizing these datasets for scene flow evaluation is presented in Table~\ref{tab:benchmarks} as benchmarks.

\begin{table}[htbp]
\footnotesize
\centering
\begin{tabular}{p{1.4cm}p{0.6cm}p{0.9cm}p{2.4cm}p{1.5cm}p{4.8cm}}
\toprule
\textbf{Literature} & \textbf{Year} & \textbf{Dataset} & \textbf{Error type} & \textbf{Error} & \textbf{Notes} \\
\toprule
Huguet~\cite{19} & 2007 & \vtop{\hbox{\strut sphere(Huguet)}\hbox{\strut Middlebury(Teddy)} \hbox{\strut Middlebury(Cones)}} & OF(RMSE,AAE) & \vtop{\hbox{\strut 0.69, 1.75}\hbox{\strut 2.85, 1.01}\hbox{\strut 3.07, 0.39}} & introducing the Huguet rotating sphere dataset and the first to use Middlebury dataset.\\
\hline
Wedel~\cite{22} & 2008 & EISATS & OF(RMSE, AAE) & 0.59, 4.13 &the first to use EISATS dataset.\\
\hline
Basha~\cite{28} & 2010 & sphere(Basha) & SF(NRMSE, AAE) & 9.71, 3.39 &introducing the Basha rotating sphere dataset.\\
\hline
Vogel~\cite{51} & 2011 & KITTI2012 & \vtop{\hbox{\strut OF(EPE ,  \%3px)}\hbox{\strut Disp(EPE ,  \%3px)}} & \vtop{\hbox{\strut 1.6, 7.07\%} \hbox{\strut 1.0, 4.87\%}} &the first to use KITTI2012 dataset.\\
\hline
Zanfir~\cite{70} & 2015 & Sintel & OF(EPE) & 4.6 &the first to use Sintel dataset.\\
\hline
Menze~\cite{67} & 2015 & KITTI2015 & \%3px(OF,Disp,SF) & 8.37, 5.79, 9.44 &introducing the KITTI2015 dataset.\\
\hline
Mayer~\cite{Mayer} & 2016 &  \vtop{\hbox{\strut Freiburg Driving}\hbox{\strut Freiburg Flying}\hbox{\strut Freiburg Monkaa}} & \vtop{\hbox{\strut EPE}\hbox{\strut(OF,Disp,DispC)}} & \vtop{\hbox{\strut 22.01,17.56,16.89}\hbox{\strut 13.45,2.37,0.91}\hbox{\strut 7.68,6.16,0.81}} &introducing Freiburg dataset.\\
\bottomrule
\end{tabular}
\caption{Summary of the methods that firstly utilized the datasets for scene evaluation. Notation for data format: OF: optical flow error. Disp: stereo matching error. SF: scene flow error. \%3px: the percentage of error pixels of which the EPE is larger than 3px.}
\label{tab:benchmarks}
\end{table}

%%%%%%%%
%%%%%%%%
%%%%%%%%
%%%%%%%%------------------------------Discussion部分----------------------------
%%%%%%%%
%%%%%%%%
%%%%%%%%
%%%%%%%%
\section{Discussion}
\label{sec:discussion}
This paper has categorized and analyzed most of scene flow estimation methods, few questions are arisen as follows, with our own opinions given as well. A future vision for scene flow estimation methods and scene flow evaluation protocols is depicted.
\subsection{Which kind of methods performs better?}
As the taxonomy Section~\ref{sec:taxonomy} presents, most methods share the similar energy minimization framework. Each kind has its pros and cons. Scene flow estimation under the multi-view stereopsis provides comprehensive geometry and motion information, which handles occlusion with less error. However, the large data size leads to the poor efficiency. Binocular-based scene flow estimation deals with two ill-posed problem in terms of stereo and motion, but it can be widely applied in diverse scenes. RGB-D scene flow estimation utilized the cheap depth information with better efficiency and better robustness under a low illumination condition, but it can only be applied in an indoor circumstance due to the limitation of sensors. And light field camera is just an emerging technique without the mature application. 

Calculation scheme is the key property of each algorithm. With four different categories introduced in Section~\ref{sub:method}, the tiny difference between methods that belongs to the same kind are presented in Table~\ref{tab:globalvariation}, ~\ref{tab:pixelassignment}, ~\ref{tab:featurematching}. Global variation methods mentioned in Section~\ref{subsubsec:globalvariational} is a common and general solution. Pixel assignment methods mentioned in Section~\ref{subsubsec:pixelassignment} aims to handle the inaccuracy around the boundaries for better performance, but additional ill-posed segmentation issue is involved in the framework. Feature matching methods aims for specific issue like large displacement and alleviating complexity for better efficiency, but the propagation from sparse features is worth attention. Learning based methods arouse extensive attention, but this emerging technique remains a long way to go.

\begin{figure}[htbp]
\centering    
\subfigure[Middlebury Teddy benchmarks] 
{ \label{fig:middleburybenchmark}     
\includegraphics[width=0.45\columnwidth]{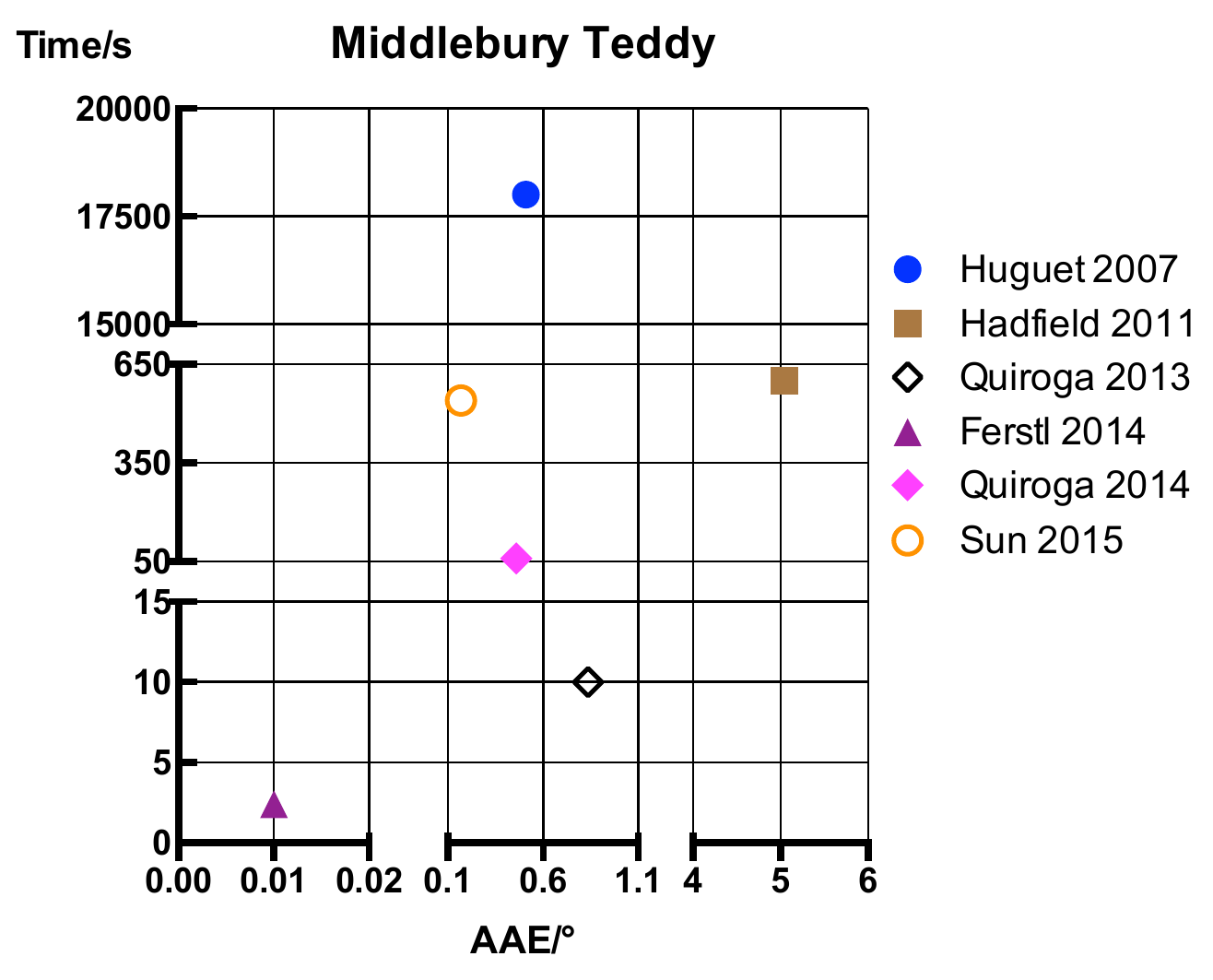}  
}     
\hspace{0.05\columnwidth}
\subfigure[KITTI 2015 benchmarks]
{ \label{fig:KITTI2015benchmark}     
\includegraphics[width=0.45\columnwidth]{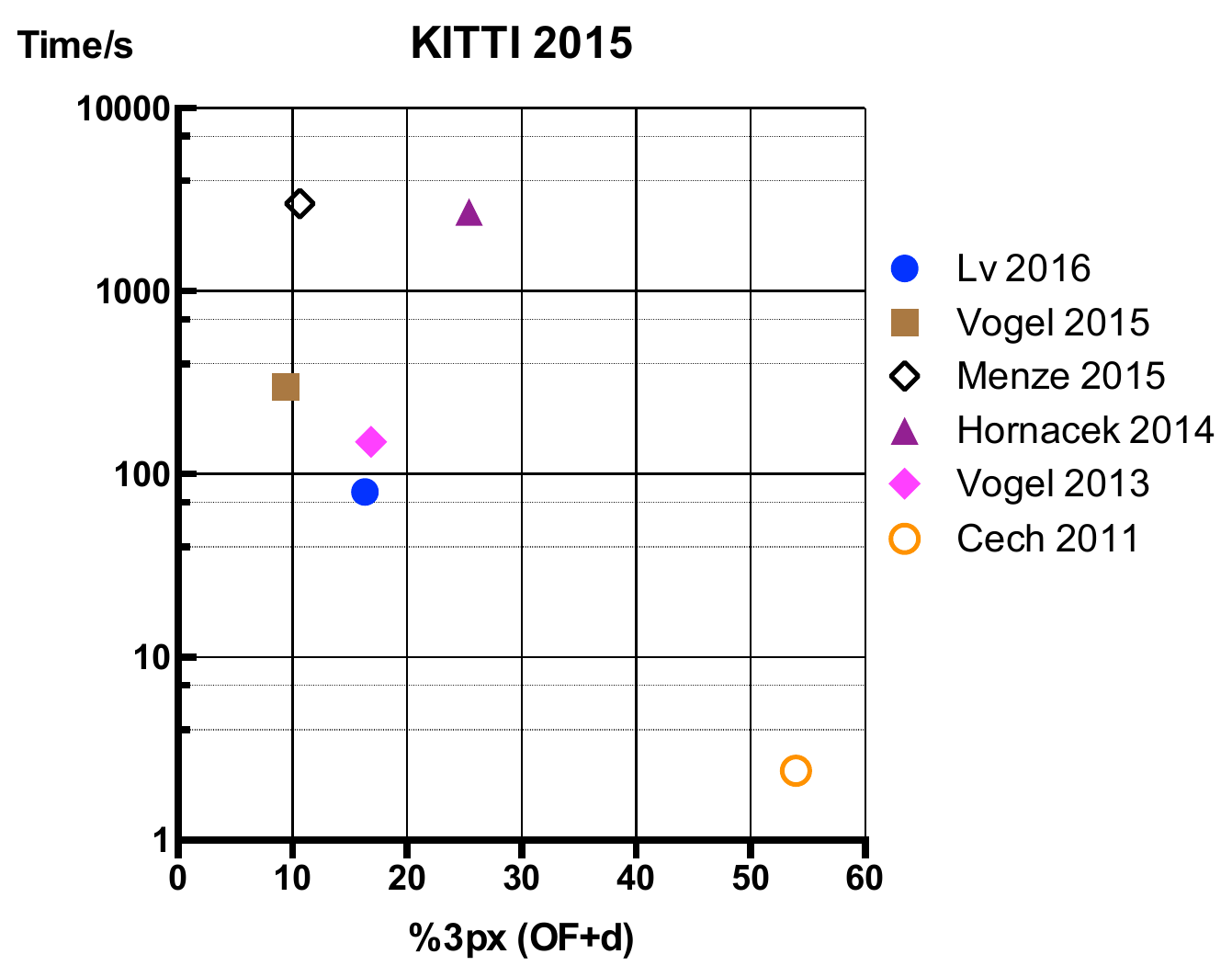}     
}     
\caption{The two-dimensional accuracy-efficiency diagram. A point close to origin point is ideal with less error and less computational time.}  
\label{fig:accuracy-efficiency}   
\end{figure}

A two-dimensional diagram is depicted in Figure~\ref{fig:accuracy-efficiency} to show the performances between efficiency and accuracy of the representative algorithms. Table~\ref{tab:accuracyperformance} summarizes the top-tier performance in some of the datasets, and Table~\ref{tab:efficiencyperformance} summarizes algorithms with great efficiency. 

\begin{table}[htbp]
\footnotesize
\centering
\begin{tabular}{p{2cm}p{1cm}p{2cm}p{2cm}p{5.8cm}}
\toprule
\vtop{\hbox{\strut \textbf{Literature}} \hbox{\strut \textbf{Year}}} &\vtop{\hbox{\strut \textbf{Category}}\hbox{\strut \textbf{So/Re/Sc}}}& \vtop{\hbox{\strut \textbf{Dataset}}\hbox{\strut \textbf{Error type}}}& \textbf{Error} & \textbf{Notes} \\
\toprule
\vtop{\hbox{\strut Hornacek~\cite{58}} \hbox{\strut 2014}} & \vtop{\hbox{\strut RGB-D/} \hbox{\strut patch/} \hbox{\strut feature matching}}  & \vtop{\hbox{\strut Teddy} \hbox{\strut Cones}\hbox{\strut OF(RMSE, AAE)}}  & \vtop{\hbox{\strut 0.35, 0.15} \hbox{\strut 0.54, 0.52}} & leading performance on Middlebury dataset\\
\hline
\vtop{\hbox{\strut Quiroga~\cite{60}} \hbox{\strut 2014}} & \vtop{\hbox{\strut RGB-D/} \hbox{\strut point cloud/} \hbox{\strut global variational}}  & \vtop{\hbox{\strut Teddy} \hbox{\strut Cones}\hbox{\strut OF(RMSE, AAE)}}  & \vtop{\hbox{\strut 0.49, 0.46} \hbox{\strut 0.45, 0.37}} & leading performance on Middlebury dataset\\
\hline
\vtop{\hbox{\strut Sun~\cite{63}} \hbox{\strut 2015}} & \vtop{\hbox{\strut RGB-D/} \hbox{\strut depth/} \hbox{\strut pixel assignment}}  & \vtop{\hbox{\strut Teddy} \hbox{\strut Cones}\hbox{\strut OF(RMSE, AAE)}}  & \vtop{\hbox{\strut 0.09, 0.17} \hbox{\strut 0.12, 0.13}} & the best performance on Middlebury dataset\\
\hline
\vtop{\hbox{\strut Vogel~\cite{69}} \hbox{\strut 2015}} & \vtop{\hbox{\strut binocular/} \hbox{\strut patch/} \hbox{\strut pixel assignment}}  & \vtop{\hbox{\strut KITTI2012} \hbox{\strut OF(EPE ,  \%3px)}\hbox{\strut Disp(EPE ,  \%3px)}}  & \vtop{\hbox{\strut}\hbox{\strut 1.0, 4.23\%} \hbox{\strut 0.7, 3.00\%}} & the best performance on KITTI dataset\\
\hline
\vtop{\hbox{\strut Jaimez~\cite{66}} \hbox{\strut 2015}} & \vtop{\hbox{\strut RGB-D/} \hbox{\strut point cloud/} \hbox{\strut pixel assignment}}  & \vtop{\hbox{\strut Sintel}\hbox{\strut OF(EPE)}}  & \vtop{\hbox{\strut}\hbox{\strut 1.203}} & the best performance on Sintel dataset\\
\bottomrule
\end{tabular}
\caption{Summary of the representative methods with leading accuracy performance}
\label{tab:accuracyperformance}
\end{table}

\begin{table}[htbp]
\footnotesize
\centering
\begin{tabular}{p{2cm}p{3cm}p{3cm}p{2cm}p{2cm}p{4cm}}
\toprule
\vtop{\hbox{\strut \textbf{Literature}} \hbox{\strut \textbf{Year}}} & \textbf{Resolution}& \textbf{Time/$s$} & \textbf{Implement} & \textbf{Description}\\
\toprule
\vtop{\hbox{\strut Gong~\cite{24}} \hbox{\strut 2009}} & $320 \times 240$ (QVGA)  & \vtop{\hbox{\strut 0.082(local)} \hbox{\strut 0.217(global)}} & GPU & \vtop{\hbox{\strut local: winner-take-all(WTA) scheme}\hbox{\strut global: dynamic programming}}\\
\hline
\vtop{\hbox{\strut Rabe~\cite{31}} \hbox{\strut 2010}} & $640 \times 480$ (VGA)  & \vtop{\hbox{\strut 0.040(De6D)} \hbox{\strut 0.100(Va6D)}} & GPU &  \vtop{\hbox{\strut De6D: fusing OF and stereo}\hbox{\strut Va6D: coupling OF and stereo}}\\
\hline
\vtop{\hbox{\strut Cech~\cite{37}} \hbox{\strut 2011}} & $640 \times 480$ (VGA)  & 1.5 & CPU(Linux) & \vtop{\hbox{\strut seeded growing algorithm}\hbox{\strut complexity $\mathcal{O}(n^2)$}}\\
\hline
\vtop{\hbox{\strut Wedel~\cite{40}} \hbox{\strut 2011}} & $320 \times 240$ (QVGA)  & 0.200/0.050 & CPU/GPU & \vtop{\hbox{\strut decoupling OF and stereo}\hbox{\strut variational framework}}\\
\hline
\vtop{\hbox{\strut Hung~\cite{49}} \hbox{\strut 2013}} & $640 \times 480$ (VGA)  & 28.000/1.400 & CPU/OpenMP & \vtop{\hbox{\strut voting for temporal constraint}\hbox{\strut anisotropic smoothness}}\\
\hline
\vtop{\hbox{\strut Jaimez~\cite{65}} \hbox{\strut 2015}} & $320 \times 240$ (QVGA)  & 7.150/0.042 & CPU/GPU & \vtop{\hbox{\strut primal-dual framework}\hbox{\strut regularization on 3D surface}}\\
\bottomrule
\end{tabular}
\caption{Summary of the representative methods with leading efficiency performance}
\label{tab:efficiencyperformance}
\end{table}

\subsection{Which dataset to choose?}
\begin{figure} [htbp]
\centering    
\subfigure[The average magnitude of the optical flow] 
{ \label{fig:average_mag}     
\includegraphics[width=0.38\columnwidth]{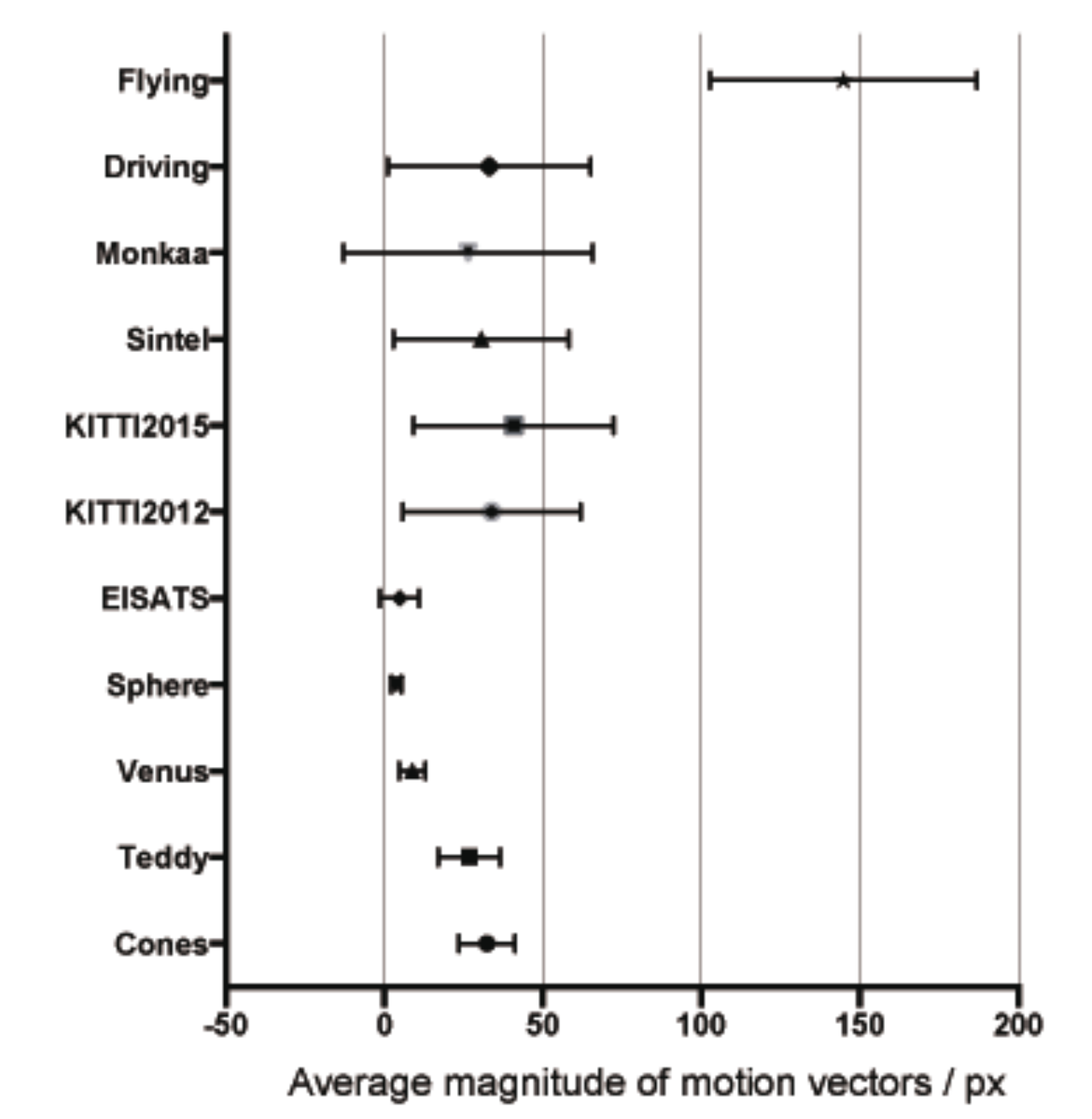}  
}     
\hspace{0.05\columnwidth}
\subfigure[The maximum magnitude of the optical flow]
{ \label{fig:max_mag}     
\includegraphics[width=0.38\columnwidth]{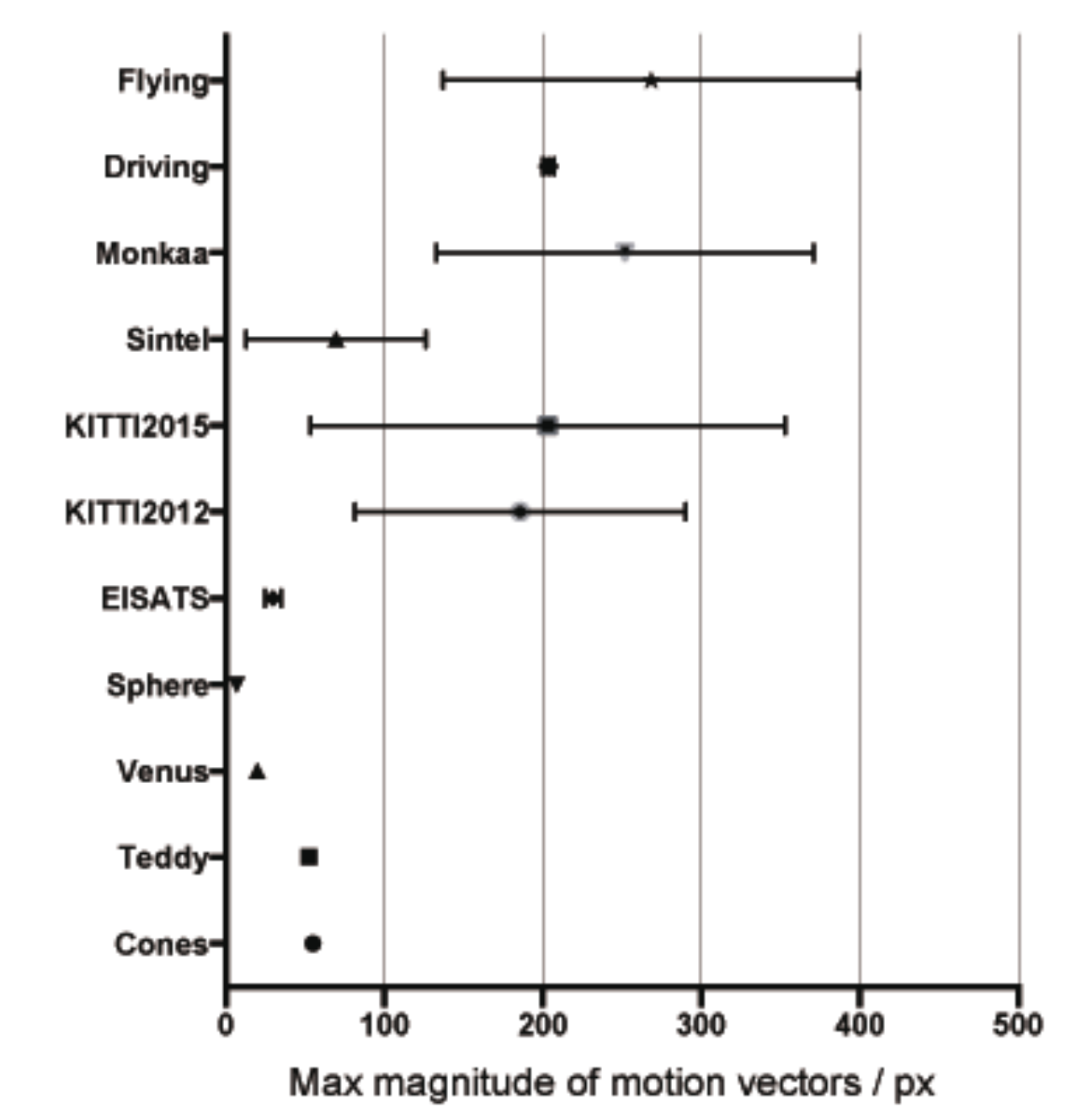}     
}     
\caption{The statistics of the ground truth optical flow magnitude in each dataset. The point annotates the exact value, and the segments on left and right of the point annotates the standard deviation value. (Note: Due to the enormous volume of data, we haven't downloaded the full package of Freiburg dataset. We take the sampler package for analysis which includes three frames in each subset.)}     
\label{fig:datasetstatistic}     
\end{figure}

We calculated the statistics of the ground truth optical flow in terms of average magnitude and the max magnitude to show the challenging level and large displacement situation of each dataset, as Figure~\ref{fig:datasetstatistic} presents. Where the bigger the average magnitude and standard deviation are, the more challenging the dataset is, and the bigger the max magnitude is, the more a robust large displacement handling is required.

We can see clearly that the up-to-date dataset is more challenging than datasets published before. Along with information provided in Table~\ref{tab:dataset}, we provide suggestion for datasets with different purpose as follows:
\\
\textbf{Comprehensiveness}
Taken multiple issues, e.g., categories of ground truth data, image resolution, challenging level, naturalism, scale, popularity, into consideration, we recommended MPI Sintel dataset and Freiburg dataset for their comprehensive property.
\\
\textbf{Challenging}
FlyingThings3D subset of Freiburg dataset shows the characteristic of large displacement, complex occlusion and diverse changes between frames, which is really challenging for scene flow estimation. Monkaa subset shows the similar characteristic which is recommended as well.
\\
\textbf{Public popularity}
Middlebury, KITTI(2012, 2015) and MPI Sintel dataset provide evaluation protocols and online ranking that can evaluate the performance of a method conveniently. Moreover, the evaluation can be compared with top tier methods in optical flow estimation field and stereo matching field to indicate the superiority of scene flow estimation.
\\
\textbf{Multi-view and RGB-D data source}
For multi-view stereopsis, Basha rotating sphere and KITTI2012/2015 provide multi-view extension for evaluation. To be noted, Basha provides only point cloud scene flow ground truth for quantitative analysis, while the ground truth of KITTI2012/2015 is sparse. In terms of RGB-D scene flow estimation, MPI Sintel dataset and Basha rotating sphere dataset provides depth ground truth so that disparity-to-depth conversion is no need. In addition, long range of distance may make the depth map transferred from disparity map unclear for visualization, which make image-based algorithm hard to work. The provided depth visualization map is a good option.
\\
\textbf{Large scale for learning}
Freiburg dataset is currently the only dataset with $10^5$ order of scale that is designed for training optical flow, disparity, and scene flow.

\subsection{Which protocol to choose?}
Protocols usually varies from each other based on different datasets. The previous datasets like Middlebury and rotating sphere usually use RMSE/NRMSE and AAE for evaluation, while newly-introduced datasets like KITTI, MPI Sintel and Freiburg utilize EPE for evaluation. We recommend researchers to use EPE as a overall protocol, while RMSE and AAE can be supplementary means that reveal error distribution and angular error. 
\\
\textbf{Should we evaluate the 3D error?}
As Equation~\ref{eq:errorprojection} presents, the disparity in time $t+1$ or the disparity change do have influence on scene flow estimation. Hence, we highly recommended that for 2D parametrization scene flow estimation methods, the accuracy of disparity $t+1$ or the disparity change should be evaluated and provided. Considering the fact that Equation~\ref{eq:NRMSE-wedel} and ~\ref{eq:AAE-wedel} is rarely utilized, and rotating sphere of Basha is the only dataset that provide three-dimension point cloud ground truth, we recommend researchers to provide EPE for optical flow, disparity in time $t$ and time $t+1$ as a common protocol for scene flow evaluation. 

\subsection{Future vision}
With 17 years of development, scene flow estimation has reached a promising status, while there are still many issues remains to be solved. In this section, we discuss the limitation of existing datasets, and present a brief vision on modification of algorithms.
\subsubsection{Limitation of current datasets}
The survey of existing evaluation methodologies reveals problems and limitations as elaborated below:
\\
\textbf{Size} On account that real-time scene flow has been achieved with GPU implementation, high resolution data can be introduced for more challenging work.
\\
\textbf{Ground truth} Point cloud is the real three-dimensional parametrization representation for scene flow, which reveals the difference between scene flow and optical flow significantly. Hence, the ground truth for scene flow under the point cloud representation is necessary. Meanwhile, occlusion, textureless region and discontinuity region ground truth are essential for evaluation due to the fact that errors mainly exist near these regions.
\\
\textbf{Data source} Current datasets mainly focus on scene flow under the binocular setting, while multi-view extension and specific datasets for RGB-D and light field based scene flow is required. Otherwise, the properties of missing data in RGB-D cameras and the abilities like refocusing of light field cameras will be neglected with current datasets.
\\
\textbf{Protocol} A protocol for evaluating performances under different datasets remains vacant. Moreover, the three-dimensional protocol isn't been applied due to the limitation of point cloud ground truth.

\subsubsection{The modification of methods in the future}
By checking the error map provided by KITTI benchmark, it's clear that inaccuracy mainly exists in the boundaries of objects. Since this is a common issue for all computer vision tasks, edge-preserving and reasonable filtering is the first priority.

GPU implementation has shown a great efficiency improvement, and the duality-based optimization has proved to enhance the efficiency of global variational methods without accuracy sacrifice. These kind of methods may be a routine in the future for better efficiency.

With the development of a robust and efficient estimation between two frames, some papers have studied motion estimation under a long sequence~\cite{11,31,23,49}. The multi-fames estimation with temporal prior knowledges deserves more attention. A robust temporal constraint can benefit the methods with a better initial value or a better feature~\cite{49} to match. The challenges like varying illumination and occlusion can be handled with the help of it.

The emerging learning based methods and light field technique has brought fresh blood to scene flow estimation. Learning method with CNN shows an upward tendency in the relevant issues of scene flow estimation like stereo matching and optical flow with promising accuracy and computational cost~\cite{MCCNN,CONTENT-cnn,FlowNet}. With the help of the up-to-date large scale training dataset~\cite{Mayer}, learning-based method has a profound potential to achieve an accurate and fast estimation. Light field camera provides more data than existing data source, which brings diverse possibilities for this field. Similar to the emergence of RGB-D cameras, this new source of data may lead to a new attractive branch.

On account of the fact that scene flow estimation relies highly on texture and intensity information, application will suffer in the night or an insufficient illumination circumstance. Moreover, the car headlights and lighting on the building that are frequent in the autonomous driving scene may interfere motion estimation significantly. Hence, the scene flow estimation with insufficient illumination is worth studying.

%geometry constraint ,  motion constrain:
%Hornacek\cite{58}, 2014,  $BC(RGB)+DCC$,  6d motion(rotation, translation)
%Zanfir\cite{70}, 2015,  $BC(RGB)+DCC$,  occlusion reasoning. 6d motion(rotation, translation)
%Alhaija\cite{71}, 2015,  $BC(RGB)+DCC$,  6d motion(rotation, translation)

%%%%%%%%
%%%%%%%%
%%%%%%%%
%%%%%%%%------------------------------Conclusion部分----------------------------
%%%%%%%%
%%%%%%%%
%%%%%%%%
%%%%%%%%
\section{Conclusion}
\label{sec:conclusion}
This paper presents a comprehensive and up-to-date survey on both scene flow estimation methods and the evaluation methodologies for the first time after 17 years since scene flow was introduced. We have discussed most of the estimation methods so researchers could have a clear view of this field and get inspired for their studies of interest. The representative methods are highlighted so the differences between these methods are clear, and the similarities between top-tier methods can be seen as a tendency for modification. The widely used benchmarks have been analyzed and compared, so are multiple evaluation protocols. This paper provides sufficient information for researchers to choose the appropriate datasets and protocols for evaluating performance of their algorithms.

There are still ample rooms for future research on accuracy, efficiency and multiple challenges. We wish our work could arise public interest in this field and bring it to a new stage.

\section*{Acknowledgements}
This work was supported by projects of National Natural Science Foundation of China [61401113]; and Natural Science Foundation of Heilongjiang Province of China [LC201426].

\end{document}